\documentclass[preprint,table,12pt,authoryear]{elsarticle}

\usepackage{amssymb}
\usepackage{amsmath} 
\usepackage{subfigure}
\usepackage{multirow}
\usepackage{tikz}
\usepackage{fancybox}
\usepackage{bm}
\usepackage{cleveref}
\usepackage{highlight}
\usepackage{ulem}
\usepackage{float}
\usepackage{graphicx}
\usepackage{highlight}
\usepackage{url} 

\usepackage{xcolor}

\newcommand{\acc}[1]{{\textcolor{black}{{#1}}}} 

\usepackage{pifont}
\newcommand*\colourcheck[1]{%
  \expandafter\newcommand\csname #1check\endcsname{\textcolor{#1}{\ding{52}}}%
}
\colourcheck{green}

\newcommand*\colourcross[1]{%
  \expandafter\newcommand\csname #1cross\endcsname{\textcolor{#1}{\ding{56}}}%
}
\colourcross{red}

\journal{}

\newcommand{\OurNeRF}{{BRDF-NeRF}} 
\newcommand{\OurNeRFShort}{{BRDF-NeRF}} 
\newcommand{\sota}{{\textit{state-of-the-art}}}

\newcommand{\Nerf}{{\textit{Neural Radiance Fields}}}
\newcommand{\NERF}{{{NeRF}}}
\newcommand{\Brdf}{{{Bidirectional Reflectance Distribution Function}}}
\newcommand{\BRDF}{{{BRDF}}}

\newcommand{\ac}{atmospheric correction}
\newcommand{\acshort}{AC}
\newcommand{\SatNeRF}{Sat-NeRF}
\newcommand{\SpSNeRF}{SpS-NeRF}
\newcommand{\lightvi}{light visibility}
\newcommand{\Lightvi}{Light visibility}

\definecolor{gLow}{HTML}{d7f2de}
\definecolor{gMid}{HTML}{6ffc92}
\definecolor{gHigh}{HTML}{00a629}

\begin{document}

\begin{frontmatter}



\title{\OurNeRF: Neural Radiance Fields with \\ Optical Satellite Images and BRDF Modelling}

\author[ipgp,lastig]{Lulin Zhang}
\author[lastig]{Ewelina Rupnik}
\author[ipgp]{Tri Dung Nguyen}
\author[ipgp]{Stéphane Jacquemoud}
\author[ipgp]{Yann Klinger} 

\address[ipgp]{{Université de Paris, Institut de Physique du Globe de Paris, CNRS}, 
	{Paris}, 
	{France}
}

\address[lastig]{{Université de Gustave Eiffel, IGN-ENSG, LaSTIG}, 
	{Saint-Mandé}, 
	{France}
}

\begin{abstract}
Neural radiance fields (NeRF) have gained prominence as a machine learning technique for representing 3D scenes and estimating the bidirectional reflectance distribution function (BRDF) from multiple images. However, most existing research has focused on close-range imagery, typically modeling scene surfaces with simplified Microfacet BRDF models, which are often inadequate for representing complex Earth surfaces. Furthermore, NeRF approaches generally require large sets of simultaneously captured images for high-quality surface depth reconstruction -- a condition rarely met in satellite imaging.
To overcome these challenges, we introduce BRDF-NeRF, which incorporates the physically-based semi-empirical Rahman-Pinty-Verstraete (RPV) BRDF model, known to better capture the reflectance properties of natural surfaces. Additionally, we propose guided volumetric sampling and depth supervision to enable radiance field modeling with a minimal number of views. Our method is evaluated on two satellite datasets: (1) Djibouti, captured at varying viewing angles within a single epoch with a fixed Sun position, and (2) Lanzhou, captured across multiple epochs with different Sun positions and viewing angles. Using only three to four satellite images for training, BRDF-NeRF successfully synthesizes novel views from unseen angles and generates high-quality digital surface models (DSMs).
\end{abstract}




\end{frontmatter}


\section{Introduction}
\label{intro}
 
Over the past two decades, significant progress has been made in image processing algorithms. In particular, 3D surface reconstruction has benefited from high-resolution spatial data and algorithms such as semi-global stereo matching (SGM), which can generate detailed surface maps of urban and natural environments \citep{hirschmuller:08:sgm,rosu2015measurement,mpd:06:sgm}. However, \sota~approaches still face challenges, particularly with radiometrically {heterogeneous} surfaces, complex reflectance functions, or diachronic acquisitions. 
Recent research has attempted to address these challenges by leveraging learning algorithms capable of modelling complex resemblance functions, given appropriate architectures and sufficient training datasets \citep{PSMNet,chebbi2023deepsim,wu2024evaluation}. At the same time, a new approach to surface reconstruction has emerged with Neural Radiance Fields (\NERF), which differs from other learning-based methods by operating in a self-supervised manner. It works on single pixels rather than patches, treats non-Lambertian surfaces and generates new synthetic views \citep{Mildenhall20eccv_nerf}.
{Besides, NeRF is capable of estimating the \Brdf~(\BRDF) of the surface at the same time.}
Understanding the \BRDF~of continental surfaces is crucial for a variety of applications, including land cover mapping, assessment of Earth's radiation budget, climate change studies, vegetation density analysis, and intercalibration of spaceborne sensors \citep{dumont2010high}. However, due to the anisotropic nature of the Earth's reflectance, estimation of the \BRDF~generally requires numerous angular measurements, which is challenging with a few satellite images.
 
\begin{figure}[htb!]
	\begin{center}
				\centering
				\includegraphics[width=0.88\linewidth]{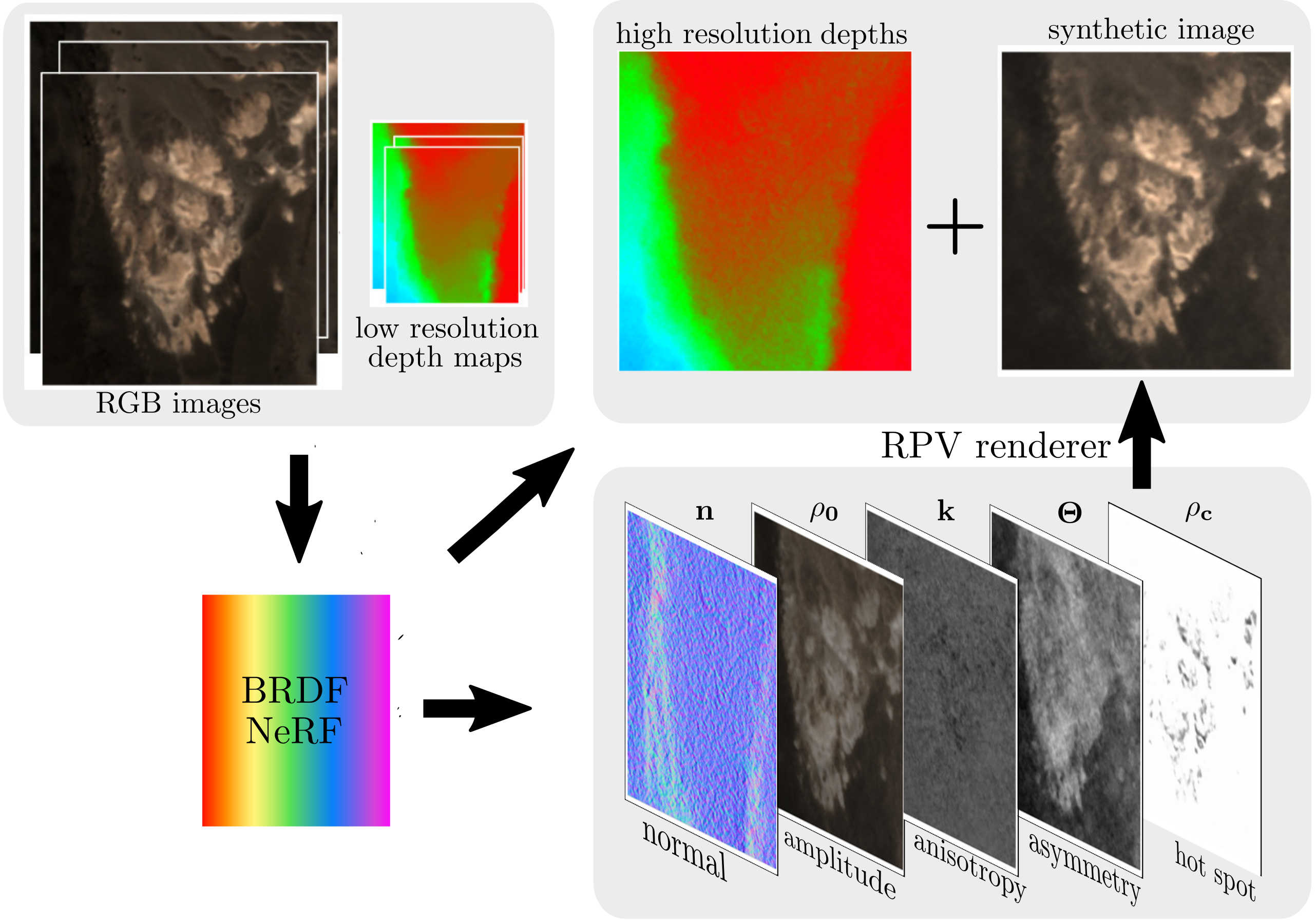} 
		\caption{ \textbf{\OurNeRF~Workflow}. A few satellite RGB images, and the corresponding low-resolution depth maps calculated using a classical stereo matching algorithm are fed into \OurNeRFShort~to predict the normals \textbf{n}, and the RPV parameters $\bm{\rho_0}$, $\textbf{k}$, $\bm{\Theta}$ and $\bm{\rho_c}$, describing the surface reflectance. $\bm{\rho_0}$ represents the amplitude component, $\textbf{k}$ controls the overall shape of the anisotropic behaviour, $\bm{\Theta}$ establishes the degree of forward or backward scattering, and $\bm{\rho_c}$ allows to model the hotspot effect. The five parameters are integrated into a RPV renderer to generate the synthetic image. In the meantime, high resolution depths are obtained by accumulating the weights in the volume estimated by \OurNeRF.}
  \label{teaser}
	\end{center}
\end{figure}

The aim of this paper is to model the surface's \BRDF~explicitly from sparse satellite views, and improve surface reconstruction, particularly over landscapes with anisotropic reflectance characteristics (e.g., bare soil, vegetation). {We select \NERF~as the algorithmic framework for its capacity to model angle dependent surface reflectance.}
Our \OurNeRFShort~workflow (\Cref{teaser}) is designed for satellite acquisitions with only three synchronous views and incorporates the semi-empirical Rahman-Pinty-Verstraete (RPV) \BRDF~model, widely used in the remote sensing community to represent the BRDF of natural surfaces \citep{rahman1993coupled}. 
To the best of our knowledge, this work is the first to integrate a \BRDF~model into neural radiance fields for land surfaces. It extends our previous work on neural radiance fields with sparse optical satellite images \citep{zhang2023spsnerf1}. The open-source code is available at \texttt{github.com/LulinZhang/BRDF-NeRF}. The terms \textit{DSM} and \textit{surface}, as well as \textit{SGM} and \textit{stereo matching} are used interchangeably throughout this article.

\section{Related works}

\subsection{\Nerf}
The vanilla \NERF~\citep{Mildenhall20eccv_nerf} leverages a large number of images captured with a pinhole camera to represent small-size scenes as particles that emit light, instead of reflecting light. Subsequent \NERF~variants proposed to relax some of those defining constraints, without compromising the quality of the outputs, i.e., the synthesised images and the 3D model. In the following paragraphs we briefly discuss the \sota~approaches relevant to our work.

\paragraph{NeRF from few views}
Since NeRF relies solely on pixel values for network training, a large number of input images is essential for generating photo-realistic novel views. Attempts to train NeRF with sparse input images often result in overfitting and inaccurate estimation of scene depth, leading to artifacts in the rendered novel views. This limitation restricts the applicability of NeRF and prolongs training time. To address this challenge, efforts have been made to adapt NeRF to sparse input images by introducing various regularization priors. A common approach is to incorporate depth supervision, including sparse depth \citep{deng2022depth, wang2022sparsenerf, Somraj_2023, guo2024depthguided} and dense depth \citep{wei2021nerfingmvs, roessle2022dense, zhang2023spsnerf1}. In addition, methods such as image features \citep{yu2021pixelnerf} or semantic regularization \citep{xu2022sinnerf} have also been explored.

\begin{figure}[!htb]
    \begin{center}
        \subfigure[Lambertian]{
            \begin{minipage}[t]{0.275\linewidth}
                \begin{tikzpicture}
                \node[anchor=south west,inner sep=0] (image) at (0,0) {
                \includegraphics[width=1\linewidth]{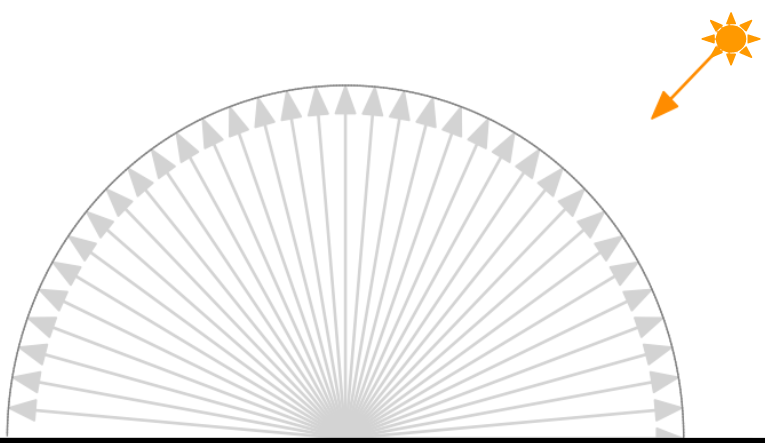}};
                \begin{scope}[x={(image.south east)},y={(image.north west)}]
                \end{scope}
                \end{tikzpicture}
            \end{minipage}%
        }
        \hspace{4mm}
        \subfigure[RPV: back scattering]{
            \begin{minipage}[t]{0.22\linewidth}
                \begin{tikzpicture}
                \node[anchor=south west,inner sep=0] (image) at (0,0) {
                \includegraphics[width=1\linewidth]{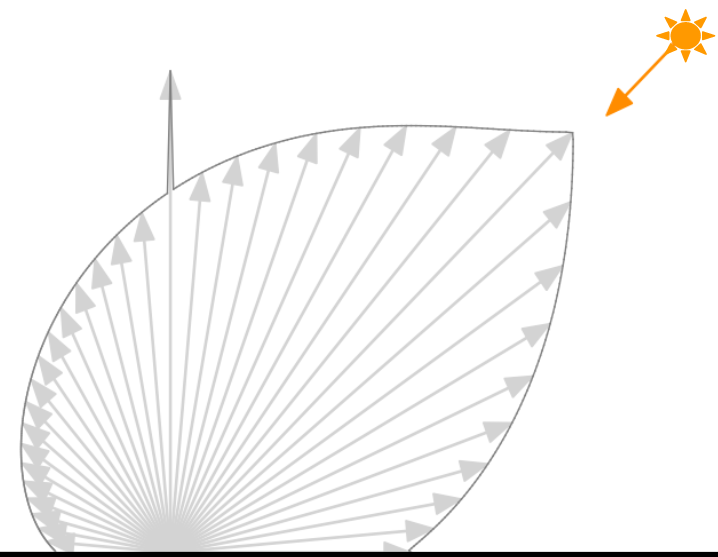}};
                \begin{scope}[x={(image.south east)},y={(image.north west)}]
                \end{scope}
                \end{tikzpicture}
            \end{minipage}%
        }
        \subfigure[RPV: forward scattering]{
            \begin{minipage}[t]{0.275\linewidth}
                \begin{tikzpicture}
                \node[anchor=south west,inner sep=0] (image) at (0,0) {
                \includegraphics[width=1\linewidth]{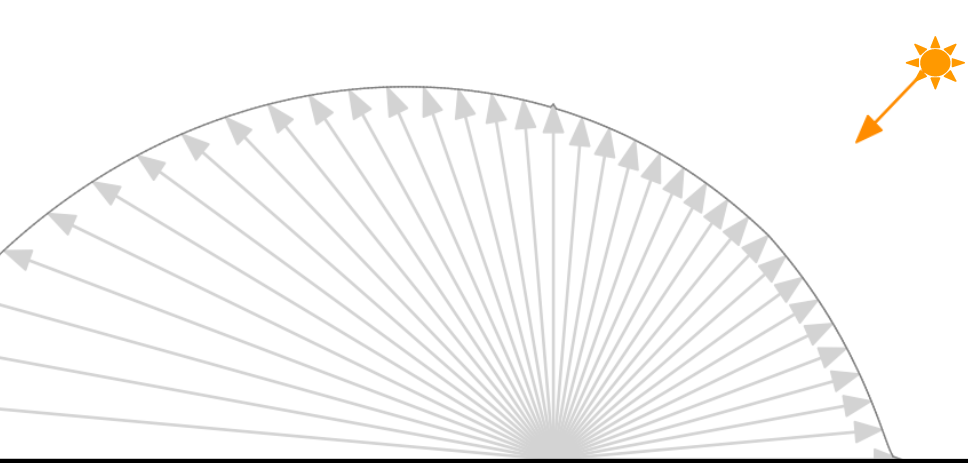}};
                \begin{scope}[x={(image.south east)},y={(image.north west)}]
                \end{scope}
                \end{tikzpicture}
            \end{minipage}%
        }
        
        \subfigure[Microfacet]{
            \begin{minipage}[t]{0.275\linewidth}
                \begin{tikzpicture}
                \node[anchor=south west,inner sep=0] (image) at (0,0) {
                \includegraphics[width=1\linewidth]{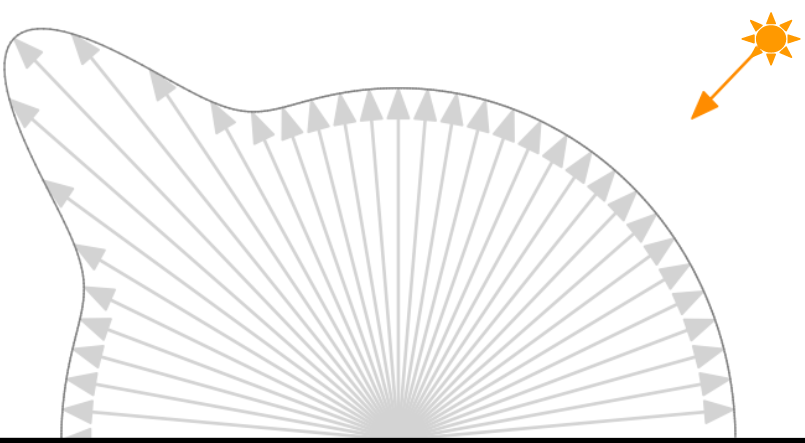}};
                \begin{scope}[x={(image.south east)},y={(image.north west)}]
                \end{scope}
                \end{tikzpicture}
            \end{minipage}%
        }        
        \hspace{4mm}
        \subfigure[RPV: bell shape scattering]{
            \begin{minipage}[t]{0.22\linewidth}
                \begin{tikzpicture}
                \node[anchor=south west,inner sep=0] (image) at (0,0) {
                \includegraphics[width=1\linewidth]{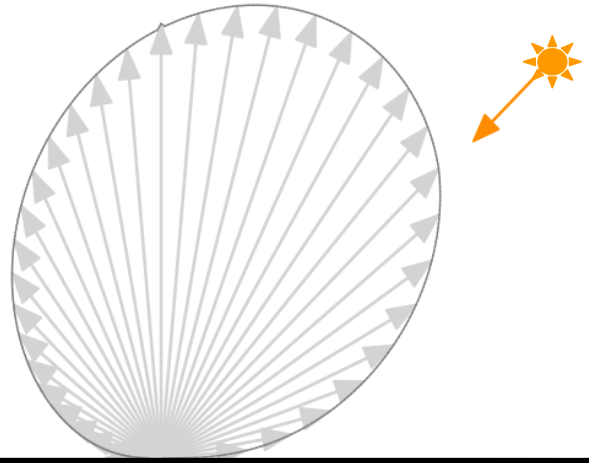}};
                \begin{scope}[x={(image.south east)},y={(image.north west)}]
                \end{scope}
                \end{tikzpicture}
            \end{minipage}%
        }        
        \subfigure[RPV: bowl shape scattering]{
            \begin{minipage}[t]{0.275\linewidth}
                \begin{tikzpicture}
                \node[anchor=south west,inner sep=0] (image) at (0,0) {
                \includegraphics[width=1\linewidth]{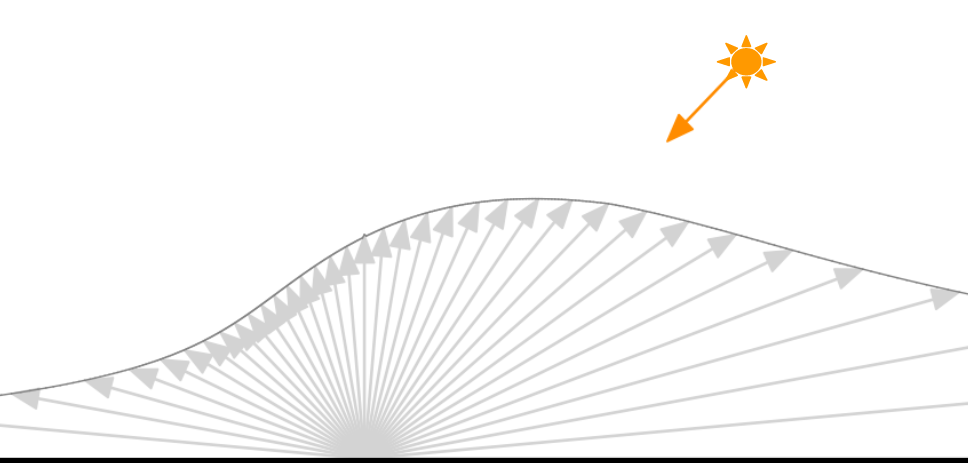}};
                \begin{scope}[x={(image.south east)},y={(image.north west)}]
                \end{scope}
                \end{tikzpicture}
            \end{minipage}%
        }        
        \caption{\textbf{Scattering Patterns.} Lambertian (a), Microfacet (d) and RPV (b, c, e, f) BRDF models. RPV can be used to simulate complex scattering indices.}
        \label{scatteringmode}
    \end{center}
\end{figure}

\paragraph{NeRF and BRDF}
While vanilla NeRF excels in view synthesis, it cannot relight or edit materials, due to its inability to decompose outgoing radiance into incoming radiance and surface material reflectance. Some researchers proposed to extend NeRF to incorporate information on the \textit{Bidirectional Reflectance Distribution Function} (BRDF), which characterises how materials reflect light under different viewing and lighting conditions. 
The majority of BRDF-compatible NeRF variants, such as those proposed by \citep{bi2020neural, srinivasan2020nerv, boss2021nerd, yang2022psnerf, verbin2022refnerf, mai2023neural}, adopt some version of the \textit{microfacet} BRDF model \citep{walter2007microfacet}. This model represents reflectance as the superposition of diffuse and specular components and typically includes a surface roughness parameter, influencing the appearance of the surface through a distribution of microfacet orientations. However, while \textit{microfacet} BRDF models offer effective parameterization, they poorly model BRDF of natural surfaces (e.g., soil, vegetation) which exhibit a more or less marked backscattering behavior (hotspot effect). Additionally, data-driven BRDFs pre-trained on BRDF databases have been explored in NeRF \citep{Zhang_2021}. However, the databases consist of artificial materials and have been built in controlled environments. The spectral and directional optical properties of natural materials are often very different. \Cref{scatteringmode} shows the scattering patterns of the Lambertian, Microfacet and RPV models. The latter is an anisotropic BRDF model widely used in remote sensing and which we will use in this work.

\paragraph{\NERF~in Earth Observations} 
Earth observation community has mainly focused on adapting the initial \NERF's design to meet the specificities of space imagery: changing shadows, dynamic scene due to asynchronous acquisitions, as well as sparse views.
Shadow-NeRF \citep{derksen2021shadow} pioneered the application of NeRFs to satellite images, where the authors have explicitly modelled the shadows of the scene by leveraging the Sun's direction. \SatNeRF \citep{mari2022sat} extends Shadow-NeRF by replacing the pinhole camera model with an empirical push broom model (i.e., \textit{Rational Polynomial Coefficients}) and modelling transient objects in the scene such as moving cars. EO-NeRF \citep{Mari_2023_CVPR} employs a novel geometry-based shadow rendering, resulting in more accurate digital surface models (DSMs).
\SpSNeRF~\citep{zhang2023spsnerf1} further adapted NeRF for scenarios with few satellite views by introducing spatial guidance within \NERF~sampling, conditioned on low-resolution input depths. 
Sat-Mesh~\citep{qu2023sat} used a latent vector to deal with inconsistent appearances in satellite imagery, while SUNDIAL \citep{behari2024sundial} proposed a secondary shadow ray casting technique to jointly learn satellite scene geometry, illumination components and Sun direction. SatensoRF \citep{zhang2024satensorf} decomposed colour into ambient, diffuse and specular light. Season-NeRF \citep{gableman2024incorporating} learned and rendered seasonal variations by incorporating time as an additional input variable.
GC-NeRF \citep{wan2024constraining} proposed a geometric loss to {create a compact weight distribution around the surface}. 
In addition,  RS-NeRF \citep{xie2023remote}, SAT-NGP \citep{billouard2024sat} and SatensoRF \citep{zhang2024satensorf} addressed the computational efficiency of \NERF~by accelerating the runtimes with hash encoding and voxel occupancy grids to sample points near the surface, as well as through tensor decomposition. 
Last but not least, Radar fields \citep{ehret2023radar} extended NeRF to spaceborne synthetic aperture radar (SAR) images. 

\subsection{BRDF in remote sensing}

\paragraph{Existing BRDF models} Numerous \BRDF~models have been developed to describe the spectral and directional reflectance of natural and artificial surfaces. They can be classified into physical, empirical and semi-empirical models. Physical models \citep{pinty1991extracting} are based on rigorously defined physical parameters and offer the most accurate descriptions of observed scenes. However, a large number of multiangular observations are required to retrieve these parameters by model inversion, making them impractical for optically complex surfaces. 
{Empirical models \citep{walthall1985simple, minnaert1941reciprocity, shibayama1985view}} are derived as simple statistical fits to observed data, and provide no additional insights into the surface type or structure. 
{Semi-empirical models \citep{rahman1993coupled, wanner1995derivation, hapke1981bidirectional, roujean1992bidirectional, lucht2000algorithm}} employ specific mathematical functions to best represent the physical interactions between the radiation field and the surface. They accept a reduced number of parameters, which facilitate their inversion. The semi-empirical RPV model \citep{rahman1993coupled} is among the most commonly used. It is capable of representing the reflectance of various natural surfaces with just four parameters (\Cref{scatteringmode}), and has been used to address atmospheric radiation transfer problems \citep{martonchik1998techniques, martonchik1998determination}; classify forest types \citep{koukal2014evaluation}; simulate plant leaves reflectance \citep{biliouris2009rpv}; estimate BRF values under unmeasured illumination and viewing angles \citep{lattanzio2007consistency}; estimate surface albedo \citep{martonchik1998techniques, martonchik1998determination, privette2002first}; and identify surface properties \citep{widlowski2001characterization, gao2003detecting}.

\paragraph{Deriving BRDF}
Most of the parameters controlling \BRDF~cannot be measured in the field, but are obtained by invertion of surface reflectance models on observations. To guarantee reliable estimates, the surface must be observed over a wide range of illumination/viewing angles. Laboratory and field measurements using goniophotometers have traditionally been used to measure reflectance \citep{lv2016multi, sandmeier2000brdf, combes2007new}. 
Over the last few decades, several spaceborne instruments have been designed to carry out multiangular observations, such as MISR, POLDER, MODIS, CHRIS/Proba, and VIIRS. These instruments have limited spatial resolutions, ranging from a few tens to a few hundreds of meters. \citep{labarre2019retrieving} have inverted the Hapke model on a set of 21 multiangular Pleiades images, acquired at a spatial resolution of 2m. However, such acquisitions are rarely available and inversion with three or four images is ill-posed.

In this paper, we explore the potential for estimating the \BRDF~of natural surfaces using as few as three high-resolution multispectral optical images. Our approach offers new possibilities for studying solar radiation reflected from the Earth's surface, taking advantage of the multiplicity, temporal coverage, and high spatial resolution of optical satellite imagery.
\section{Radiance Fields with RPV Reflectance}
We briefly introduce the vanilla \NERF~architecture and discuss two key ingredients of our \OurNeRFShort: geometry modelling (depths and normals) as well as the radiometric rendering (RPV BRDF model). The \OurNeRFShort~workflow is described in \Cref{teaser}.

\paragraph{Preliminaries}
\NERF~\citep{Mildenhall20eccv_nerf} represents a continuous volumetric field of a static scene that emits light, optimized with a fully connected deep network. Given a 3D point $\textbf{x} = (x, y, z)$ accompanied with a viewing angle $\textbf{w}_r = (d_x, d_y, d_z)$, \NERF~predicts a volume density $\sigma$ and a colour $\textbf{c} = (r, g, b)$. 
\NERF~renders images by sampling $N$ query points along each camera ray and accumulating the colours with weights defined by density, and imposes the rendered images to be close to the training images. Each camera ray $\textbf{r}$ is defined by an origin point $\textbf{o}$ and a viewing direction vector $\textbf{w}_r$ such that $\textbf{r}(t) = \textbf{o} + t\textbf{w}_r$. Each query point $\textbf{x}_i$ in $\textbf{r}$ is defined as $\textbf{x}_i = \textbf{o} + t_i\textbf{w}_r$, where $t_i$ lies between the near and far bounds of the scene, $t_n$ and $t_f$. The rendered pixel value $\textbf{C}(\textbf{r})$ of ray $\textbf{r}$ is calculated as follows:
\begin{gather}\label{NeRFCr}
    \textbf{C(r)} = \sum_{i=1}^{N} {T_i {\alpha}_i \textbf{c}_i}~, 
\end{gather}
whith ${\alpha}_i = 1 - e^{-{\sigma}_i {\delta}_i}$, $T_i = \prod_{j=1}^{i-1} {(1 - {\alpha}_j)}$ and ${\delta}_i = t_{i+1} - t_i$. $\alpha_i$ represents the opacity of the current query point $\textbf{x}_i$ and $T_i$ is the transmittance. The contribution of colour $\textbf{c}_i$ to the accumulated colour $\textbf{C}(\textbf{r})$ increases with opacity and transmittance.

\subsection{Geometric modelling}
%
We incorporate geometric information to extend the applicability of \OurNeRFShort~to sparse view acquisitions and to predict surface normals that are essential for accurate estimates of \BRDF, as detailed in \Cref{subsec:rpv}.
\paragraph{Depth supervision}
Instead of querying ray points crossing the entire volume of the scene, as is the case in the vanilla \NERF, we narrow it down to a buffer space defined around the location of an approximately known surface. This tactic reduces ambiguity and enables reliable volume densities to be estimated with fewer images. We further encourage the depths predicted by \NERF~to remain close to the input surface by using the following loss term introduced in \citep{zhang2023spsnerf1}:
\begin{equation}
    \mathcal{L}_{depth}(\textbf{r}) = \sum_{\textbf{r} \in R_{sub}} (corr(\textbf{r})(D(\textbf{r}) - \overline{D}(\textbf{r}))^2~,
    \label{depthloss}
\end{equation}
where ${D}(\textbf{r})$ are the predicted depths calculated as ${D}(\textbf{r}) = \sum_{i=1}^{N} {T_i {\alpha}_i t_i}$, while the $\overline{D}(\textbf{r})$ are the input depths obtained from stereo matching on low-resolution images. We have observed that the performance of stereo matching on low resolution images is marginally affected by a change in surface BRDF and can therefore provide sufficiently good depth initialisations for our radiance fields. The parameter corr(\textbf{r}), which corresponds to the similarity score obtained by stereo matching, acts as a weight or confidence. It adjusts the level of supervision, having a strong impact where confidence is high and a minimal impact where input depths are uncertain.
$R_{sub}$ is a subset of rays that satisfy at least one of the following two conditions: (1) $S(\mathbf{r}) > \Sigma(\mathbf{r})$; (2) $\left|(D(\textbf{r}) - \overline{D}(\textbf{r}))\right| > \Sigma(\mathbf{r})$, where
$S(\textbf{r})^2 = \sum_{i=1}^{N} {T_i {\alpha}_i (t_i - D(\textbf{r}))^2}$ represents the uncertainty of the predicted depth, and $\Sigma(\textbf{r}) = 1 - corr(\textbf{r})$ represents the uncertainty of the input depth. In other words, depth supervision is only applied to rays for which the predicted depths are more uncertain than the input depths.
%

 

\paragraph{Surface normal} \BRDF~is a function that depends on both the incident and viewing angles, which are defined relative to the surface normal. Therefore, the surface normal is crucial to accurately recovering the \BRDF. In \NERF, it can be derived as \textit{analytical} or \textit{learned}. The analytical normal is calculated as the negative of the normalized gradient of the density field $\sigma$ with respect to the spatial location \textbf{x} as $\textbf{n(x)} = -\frac{\nabla_{x} (\sigma)}{ \Vert \nabla_{x} (\sigma) \Vert _2 }$ \citep{srinivasan2020nerv}. The learned normal is predicted from a spatial MLP and can be supervised implicitly \citep{bi2020neural} or with the analytical normal \citep{verbin2022refnerf, li2022neural}. 
%
Approaches relying on learnt normals led to smooth surfaces and a loss of detail in our case studies.
%
Consequently, we chose to incorporate the analytical normal into our architecture because, despite its computational cost, it provides more accurate and better resolved normal vectors \citep{srinivasan2020nerv}. 


\subsection{Radiometric rendering}\label{subsec:rpv}
The geometric approach presented above guarantees decent 3D reconstructions of Lambertian scenes. Next, we adapt this approach to handle non-Lambertian natural surfaces by estimating a \BRDF~and incorporating it into the rendering \Cref{NeRFCr}.
\paragraph{RPV equation}
\label{RPVrenderingequation}
We estimate the reflectance of natural surfaces using the Rahman-Pinty-Verstraete (RPV) model \citep{rahman1993coupled}, a semi-empirical model well suited to satellite images (see \Cref{renderingequation}). We chose this model for its simplicity, its physics-based parameters and its ability to represent asymmetric \BRDF, including the hotspot effect. The latter corresponds to a sharp increase in reflectance, which becomes maximum when the illumination and viewing directions are coincident.

\begin{figure}[t!]
	\begin{center}
				\centering
				\includegraphics[width=0.52\linewidth]{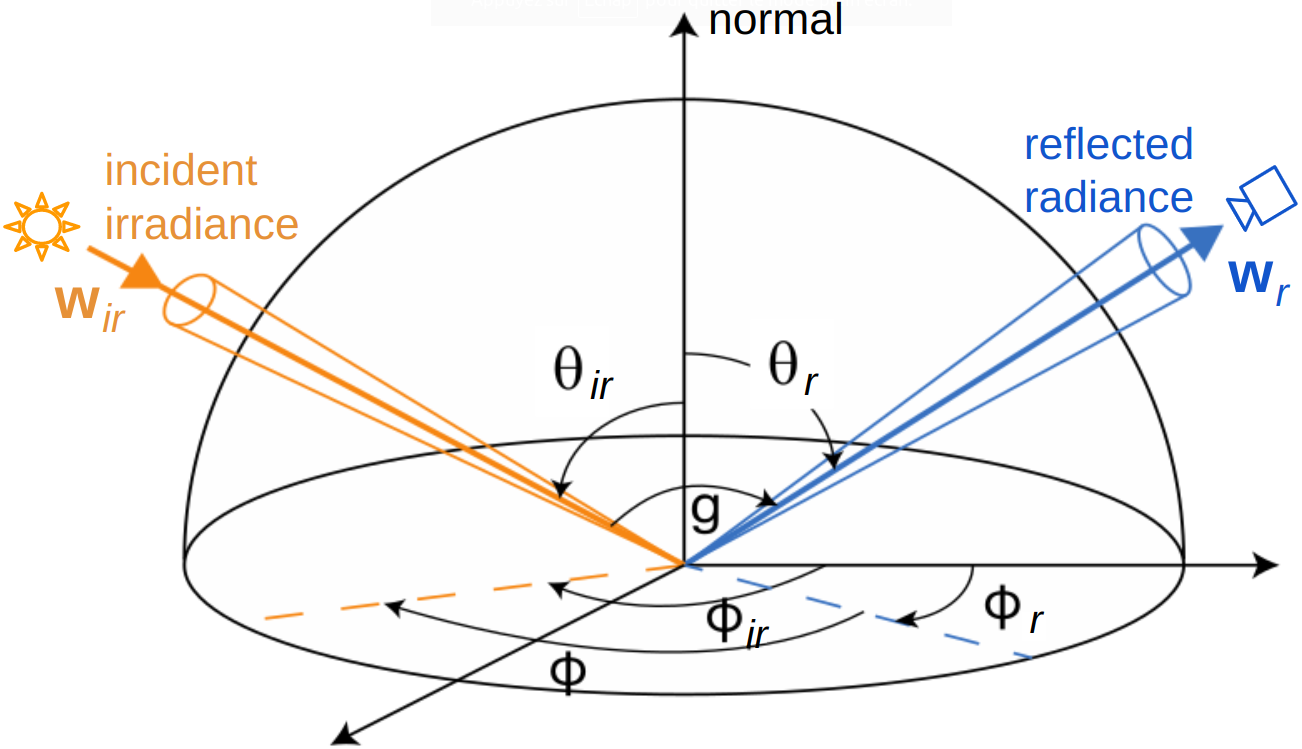}
		\caption{\textbf{BRDF Defining Directions}. Decomposition of the illumination $\textbf{w}_{ir}$ and viewing $\textbf{w}_{r}$ directions into {${\theta}_{ir}$ (solar zenith angle) and ${\theta}_{r}$ (viewing zenith angle), $\Phi_{ir}$ (solar azimuth angle) and $\Phi_{r}$ (viewing azimuth angle), g (phase angle) and $\Phi$ (relative azimuth angle).} }
  \label{BRDFconfiguration}
	\end{center}
\end{figure}

In this model, the colour $\textbf{c}$ of a surface point, defined by the normal vector $\textbf{n}$, the illumination direction $\textbf{w}_{ir}$ and the viewing direction $\textbf{w}_{r}$ (\Cref{BRDFconfiguration}), is calculated as the product of the incoming light $L_{ir}$, the cosine of the incident angle $|\textbf{w}_{ir} \cdot \textbf{n}|$, and the bidirectional reflectance factor simulated by $RPV$:

\begin{equation}
    \textbf{c} (\textbf{n}, \textbf{w}_{ir}, \textbf{w}_r) = L_{ir} \cdot |\textbf{w}_{ir} \cdot \textbf{n}| \cdot RPV (\textbf{n}, \textbf{w}_{ir}, \textbf{w}_r) \label{renderingequation},
\end{equation}
$L_{ir}$ is set to a unit vector, and $|\textbf{w}_{ir} \cdot \textbf{n}|$ is approximated by $|\textbf{w}_{ir} \cdot [0, 0, 1]|$ because the analytical normal $\textbf{n}$ is not sufficiently smooth. The $RPV$ term can be broken down into an amplitude parameter $\bm{\rho_0}$ and three angle-dependent functions: modified Minnaert function $M$, Henyey-Greensteon function $F_{HG}$, and backscatter function $H$:
\begin{equation}
    RPV (\textbf{n}, \textbf{w}_{ir}, \textbf{w}_r) = \bm{\rho_0} \cdot M({\theta}_{ir}, {\theta}_r, \textbf{k}) \cdot F_{HG}(g, \bm{\Theta}) \cdot H(\bm{\rho_c}, G) \label{rpvequation}, 
\end{equation}
with $M({\theta}_{ir}, {\theta}_r, \bm{k}) = (cos\theta_{ir} cos\theta_r(cos\theta_{ir}+cos\theta_r))^{\bm{k}-1}$, $F_{HG}(g, \bm{\Theta}) = (1-\bm{\Theta}^2) \cdot (1+2\bm{\Theta} cosg+\bm{\Theta}^2)^{-3/2}$, $H(\bm{\rho_c}, G) = 1 + (1-\bm{\rho_c})/(1+G)$, and the geometric factor $G = (tan^2 \theta_{ir} + tan^2 \theta_r - 2 tan \theta_{ir} tan \theta_r cos \Phi \nonumber)^{1/2}$.
%
%
The illumination $\textbf{w}_{ir}$ and viewing $\textbf{w}_{r}$ directions are decomposed into zenith angles ${\theta}_{ir}$ and ${\theta}_{r}$, azimuth angles ${\Phi}_{ir}$ and ${\Phi}_{r}$, relative azimuth angle $\Phi$ and phase angle $g$, all defined in a spherical coordinate system determined by the surface normal $\textbf{n}$ (\Cref{BRDFconfiguration}).




\paragraph{Detailed Description of RPV Model Input Parameters}
The RPV parameter $\bm{\rho_0}$ in \Cref{rpvequation} plays the role of pseudo-albedo. The modified Minnaert function controls the anisotropic behaviour of the surface using the parameter~\textbf{k}. If $\textbf{k} \approx 1$ the surface is quasi-Lambertian; if $\textbf{k} \textless 1$ a bowl-shaped pattern dominates (reflectance values increase with the viewing zenith angle); and if $\textbf{k} \textgreater 1$ a bell-shaped pattern dominates (reflectance values decrease with the viewing zenith angle) \citep{widlowski2004canopy}.
The parameter $\bm{\Theta}$ of the Henyey-Greenstein function controls the amount of radiation scattered in the forward (0 $\leq \bm{\Theta} \leq$ 1) or backward (-1 $\leq \bm{\Theta} \leq $ 0) directions. The backscatter function $H$ is written as a function of a geometric factor $G$ and a parameter $\bm{\rho_c}$, which represents the sharp increase in reflectance in the hotspot direction. 
When $\theta_{ir} = \theta_r$ and $\Phi_{ir} = \Phi_r$, the geometric factor disappears and $H$ reaches its maximum value, contributing to increase total reflectance. When estimating the RPV model, the ranges of variation of the parameters $\bm{\rho_0}$, \textbf{k}, $\bm{\Theta}$ and $\bm{\rho_c}$ are fixed to [0, 1], [0, 2], [-1, 1] and [0, 1] {\citep{koukal2014evaluation}}.

\begin{figure}[!htb]
    \begin{center}
        \subfigure[Point location]{
            \begin{minipage}[t]{0.205\linewidth}
                \centering
                \includegraphics[width=1.0\linewidth]{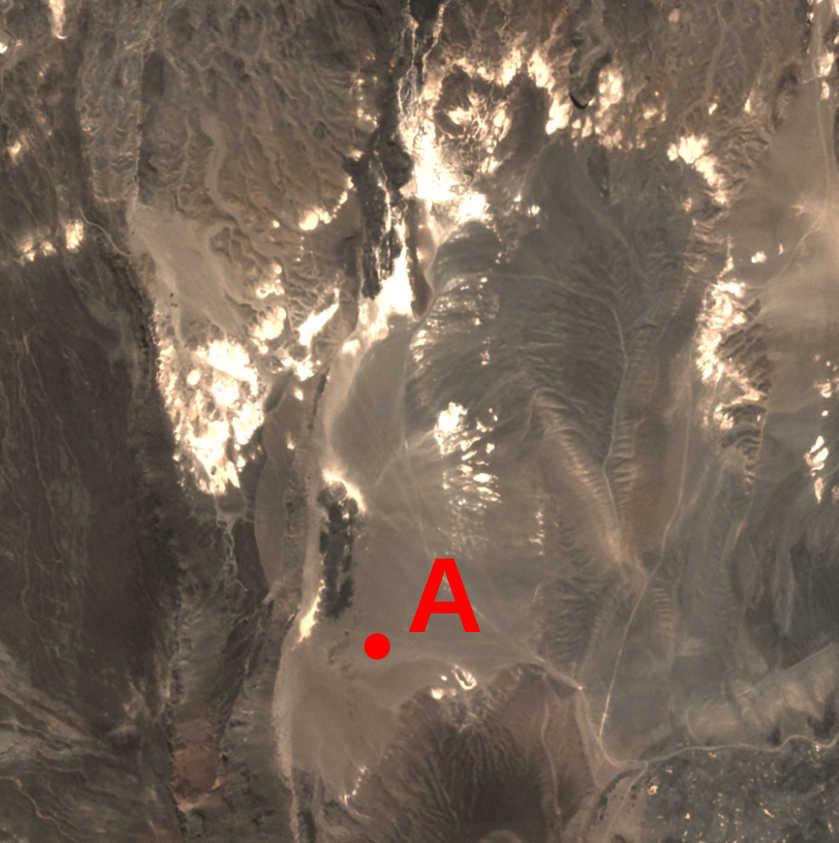}
            \end{minipage}%
        }
        \subfigure[Backward]{
            \begin{minipage}[t]{0.235\linewidth}
                \begin{tikzpicture}
                \node[anchor=south west,inner sep=0] (image) at (0,0) {
                \includegraphics[width=1\linewidth]{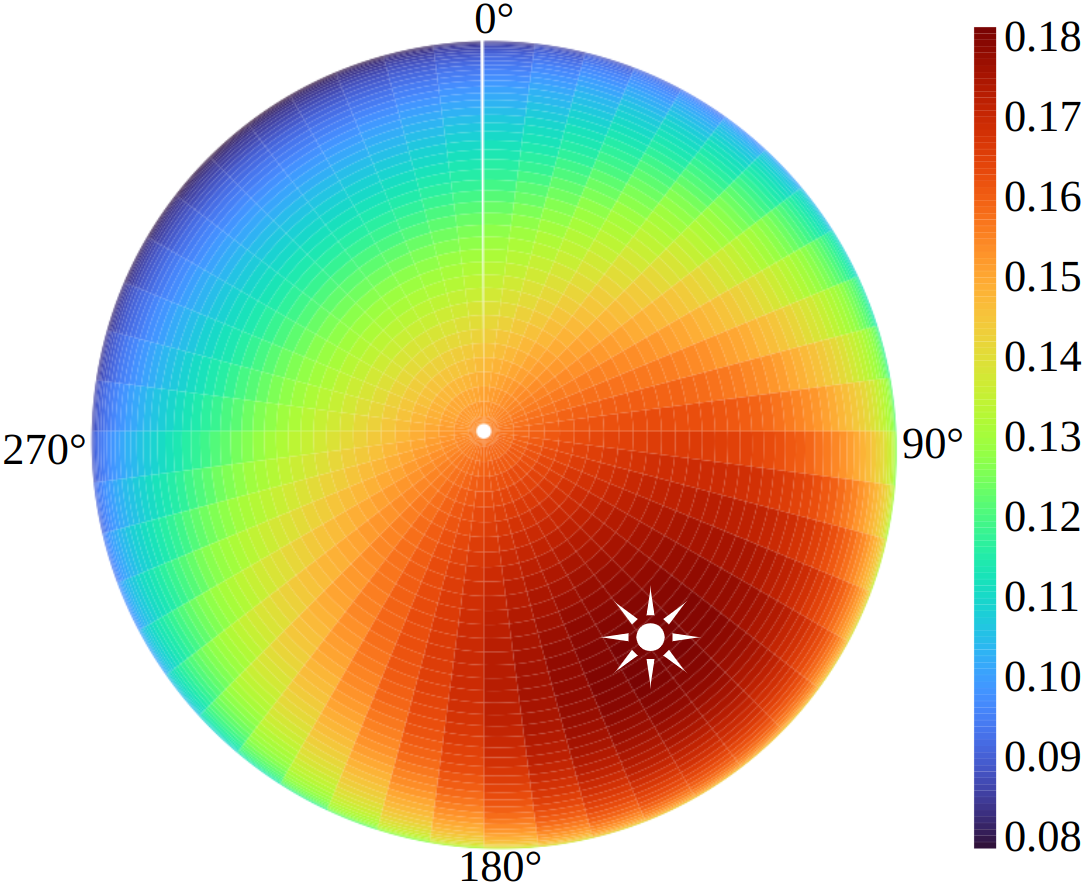}};
                \begin{scope}[x={(image.south east)},y={(image.north west)}]
                \end{scope}
                \end{tikzpicture}
            \end{minipage}%
            }
        \subfigure[Forward]{           
            \begin{minipage}[t]{0.235\linewidth}
                \begin{tikzpicture}
                \node[anchor=south west,inner sep=0] (image) at (0,0) {
                \includegraphics[width=1\linewidth]{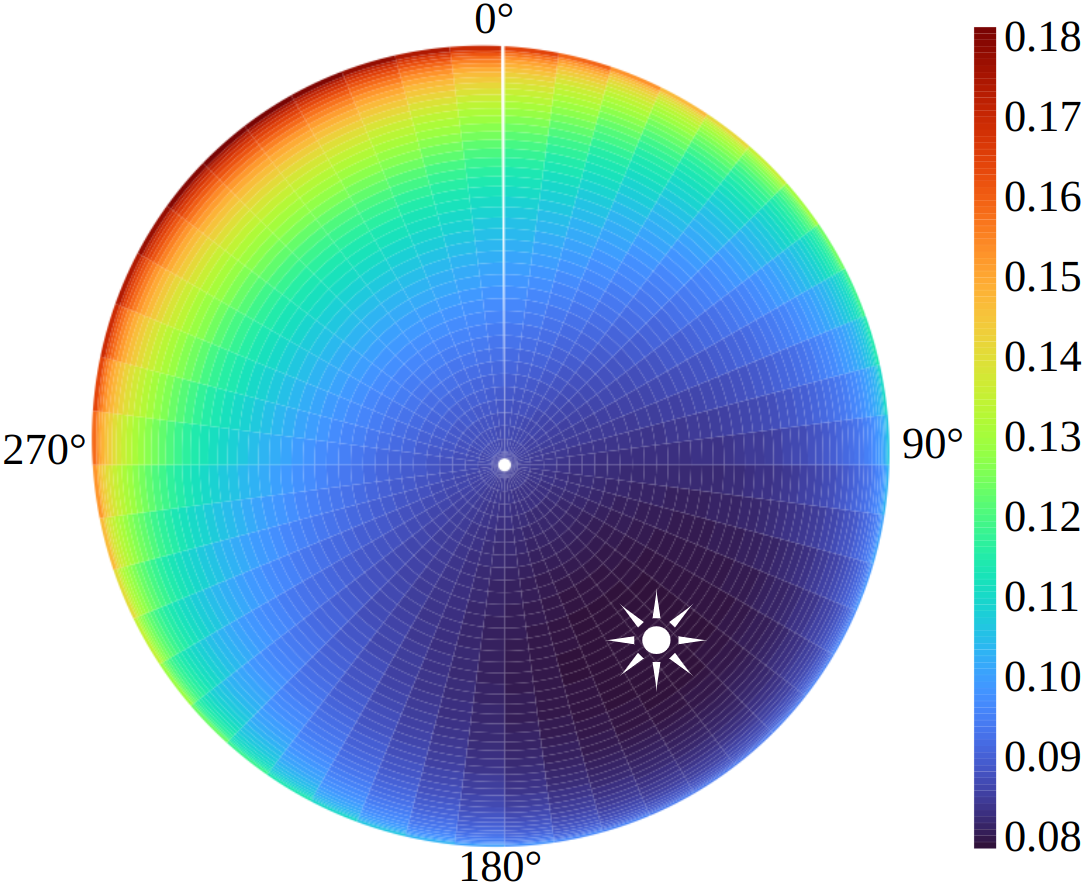}};
                \begin{scope}[x={(image.south east)},y={(image.north west)}]
                \end{scope}
                \end{tikzpicture}
            \end{minipage}%
        }\\
        \subfigure[Bowl shape]{
            \begin{minipage}[t]{0.235\linewidth}
                \begin{tikzpicture}
                \node[anchor=south west,inner sep=0] (image) at (0,0) {
                \includegraphics[width=1\linewidth]{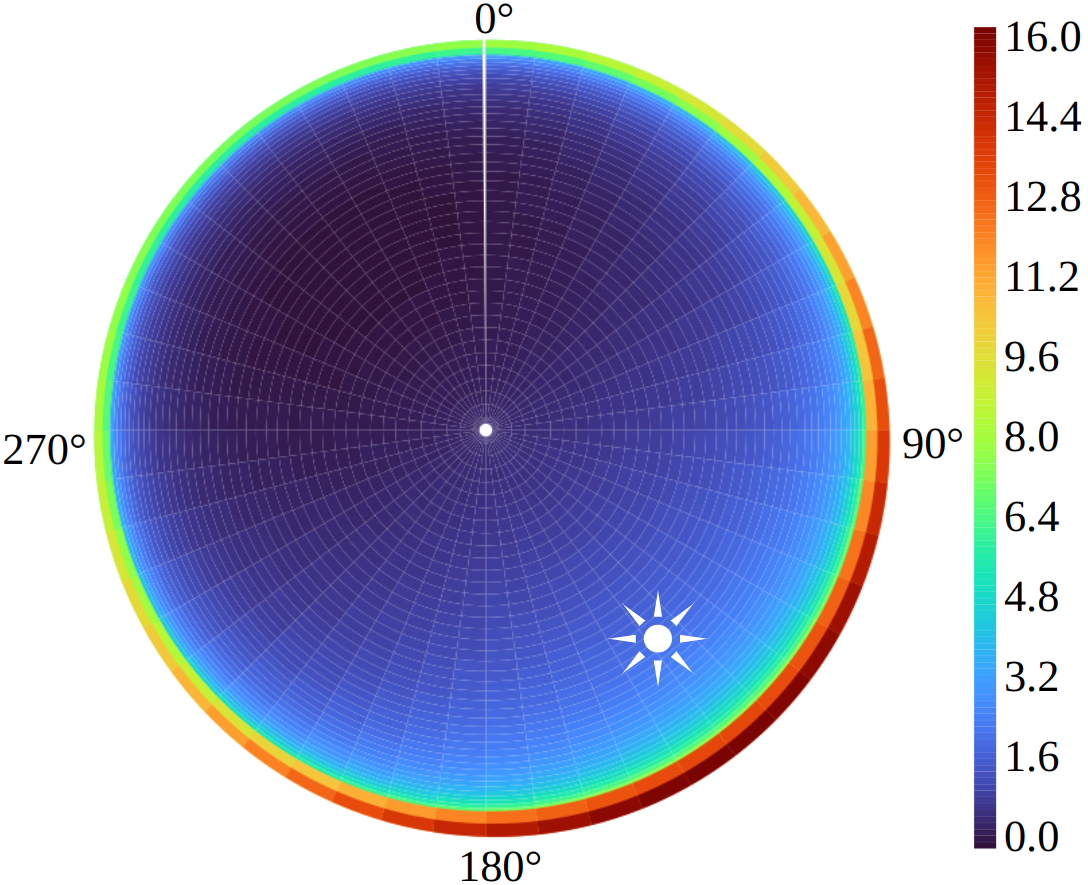}};
                \begin{scope}[x={(image.south east)},y={(image.north west)}]
                \end{scope}
                \end{tikzpicture}
            \end{minipage}%
            }
        \subfigure[Bell shape]{
            \begin{minipage}[t]{0.235\linewidth}
                \begin{tikzpicture}
                \node[anchor=south west,inner sep=0] (image) at (0,0) {
                \includegraphics[width=1\linewidth]{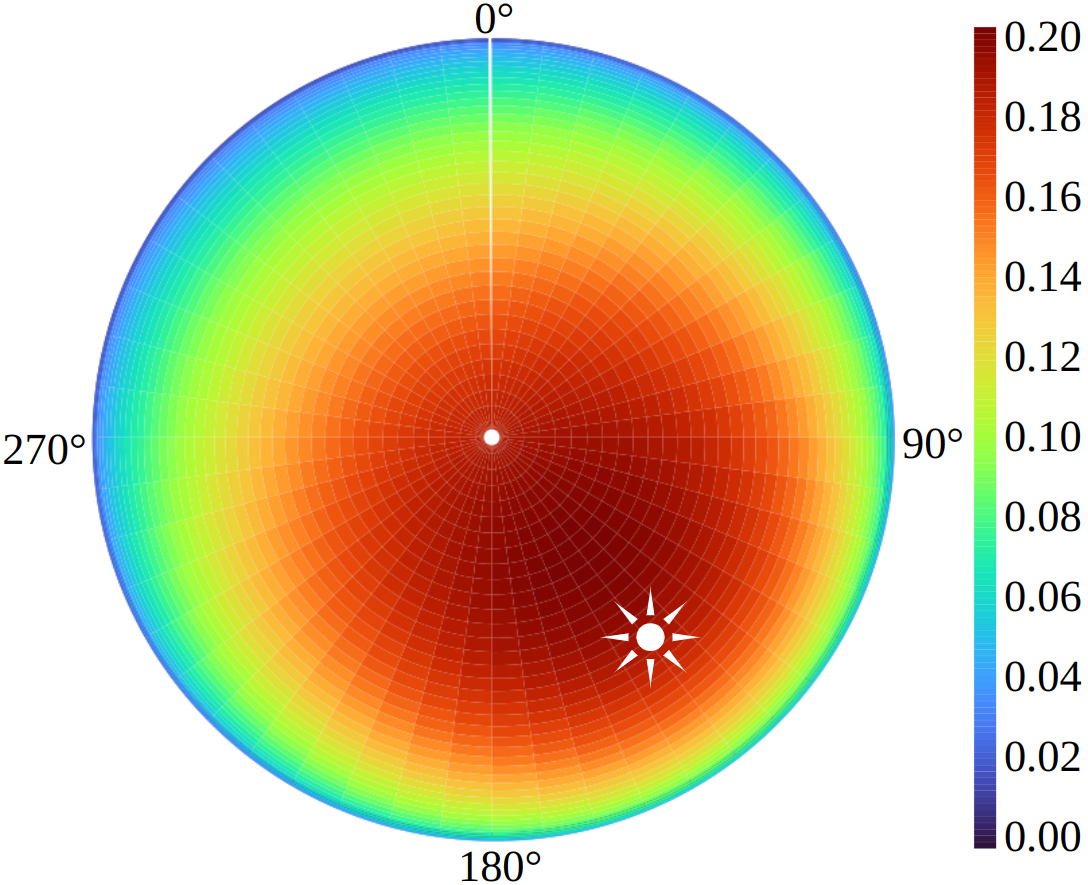}};
                \begin{scope}[x={(image.south east)},y={(image.north west)}]
                \end{scope}
                \end{tikzpicture}
            \end{minipage}%
        }
        \subfigure[Hotspot effect]{
            \begin{minipage}[t]{0.235\linewidth}
                \begin{tikzpicture}
                \node[anchor=south west,inner sep=0] (image) at (0,0) {
                \includegraphics[width=1\linewidth]{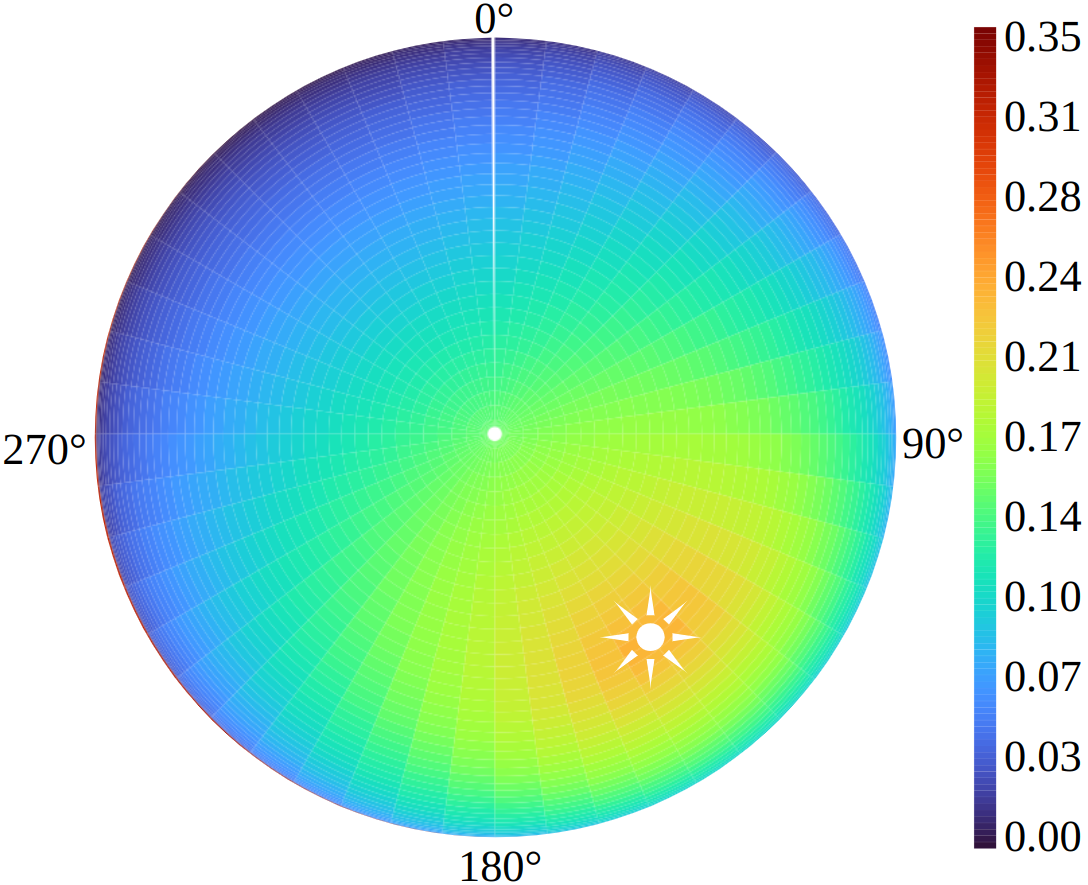}};
                \begin{scope}[x={(image.south east)},y={(image.north west)}]
                \end{scope}
                \end{tikzpicture}
            \end{minipage}%
            \hspace{0mm}
            \begin{minipage}[t]{0.235\linewidth}
                \begin{tikzpicture}
                \node[anchor=south west,inner sep=0] (image) at (0,0) {
                \includegraphics[width=1\linewidth]{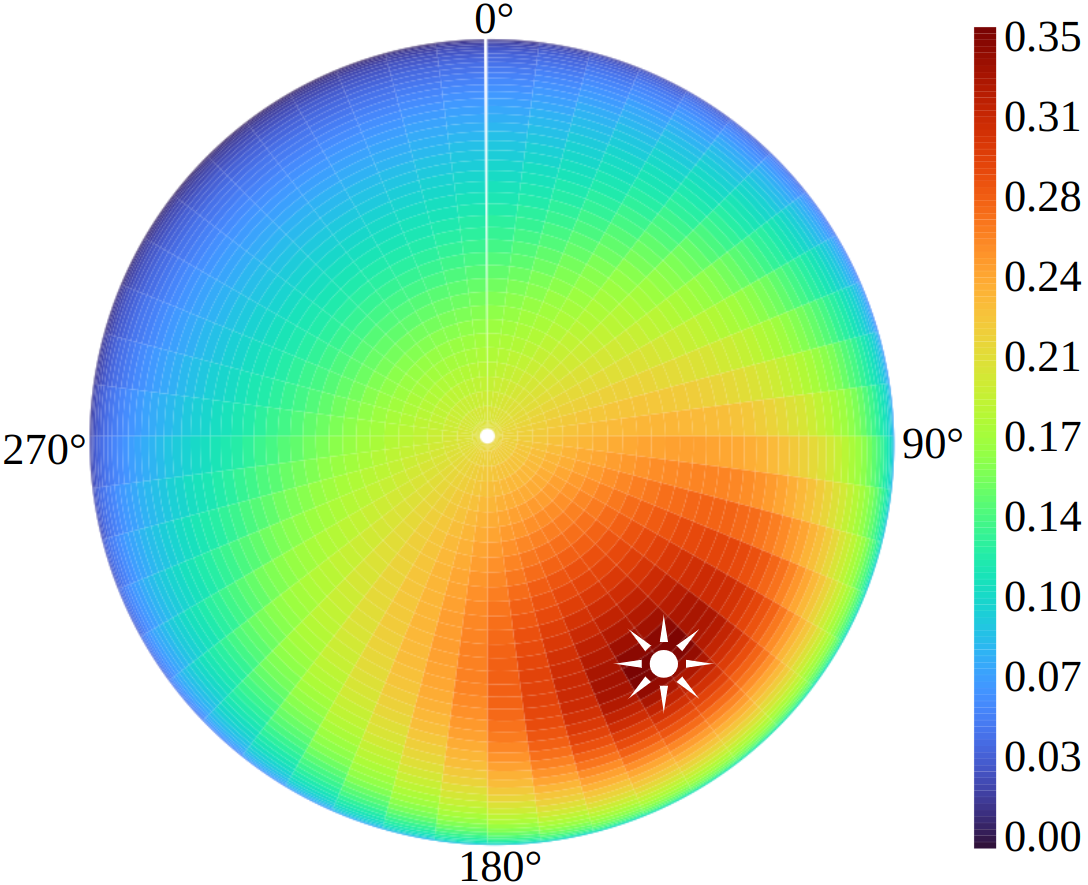}};
                \begin{scope}[x={(image.south east)},y={(image.north west)}]
                \end{scope}
                \end{tikzpicture}
            \end{minipage}%
        }
        \caption{\textbf{RPV Model Parameters -- BRF}. Reflectances displayed in polar coordinates correspond to point A shown in (a). They are generated by evaluating the RPV model over a range of viewing directions, where $\bm{\rho_0}$ and $\textbf{n}$ are predicted by our \OurNeRFShort~and fixed, while the Sun position is fixed and displayed as a white symbol. (b) backward scattering corresponding to the RPV parameters ($\bm{\Theta},\mathbf{k},\bm{\rho_c}$) found by \OurNeRFShort; (c) forward scattering obtained by modifying $\bm{\Theta}$; (d-e) transition between bowl shaped and bell shaped scattering obtained by changing the $\mathbf{k}$ parameter; (f) hotspot effect with modified $\bm{\rho_c}$. The combinations of the six parameters are provided in \Cref{BRF_circle_table}.
        }
        \label{BRF_circle}
    \end{center}
\end{figure}

To give the reader an intuition about how RPV parameters affect reflectance, we analyse the bidirectional reflectance function (BRF) of selected points in one of our datasets (\Cref{BRF_circle}). The BRFs are plotted by varying the viewing directions (zenith and azimuth angles between [0$^{\circ}$, 90$^{\circ}$] and [0$^{\circ}$, 360$^{\circ}$]) and fixing the Sun's direction to $\theta_{ir}=52.1^{\circ}$ and $\Phi_{ir}=142.5^{\circ}$. The pseudo-albedo of the selected surface point estimated by \OurNeRFShort~is $\bm{\rho_0}=[0.122, 0.105, 0.091]$, the normal \textbf{n}=[0, 0, 1]. 
Among the six combinations of ($\bm{\Theta},\mathbf{k},\bm{\rho_c}$) given in \Cref{BRF_circle_table}, backward scattering (\Cref{BRF_circle} (b)) was predicted by \OurNeRFShort. Note that this BRF is consistent with a result generated independently over the same area with 21 Pleiades views \citep{labarre2019retrieving}. 

\begin{table}[tbh!]
	\centering
	\begin{tabular}{|c|c|c|c|c|c|c|}
		\hline
		Param.&Backward&Forward&Bowl shape&Bell shape&\multicolumn{2}{|c|}{Hotspot effect} \\ \hline\hline
		\textbf{k} &0.996& 0.996& 0.500& 1.500& 0.996& 0.996 \\ \hline
		$\bm{\Theta}$&-0.174&0.174& -0.174&-0.174&-0.174&-0.174 \\ \hline
		$\bm{\rho_c}$&0.979& 0.979& 0.979& 0.979& 0.500& 0 \\ \hline
	\end{tabular}
	\caption{\textbf{RPV Model Parameter Sets.} Different RPV parameters \textbf{k}, $\bm{\Theta}$ and $\bm{\rho_c}$ corresponding to the reflectance spectra in Figure \ref{BRF_circle} (b-f).}
	\label{BRF_circle_table}
\end{table}

\newpage
\subsection{Network architecture}
\begin{figure}[htbp!]
	\begin{center}
				\centering
				\includegraphics[width=0.9\linewidth]{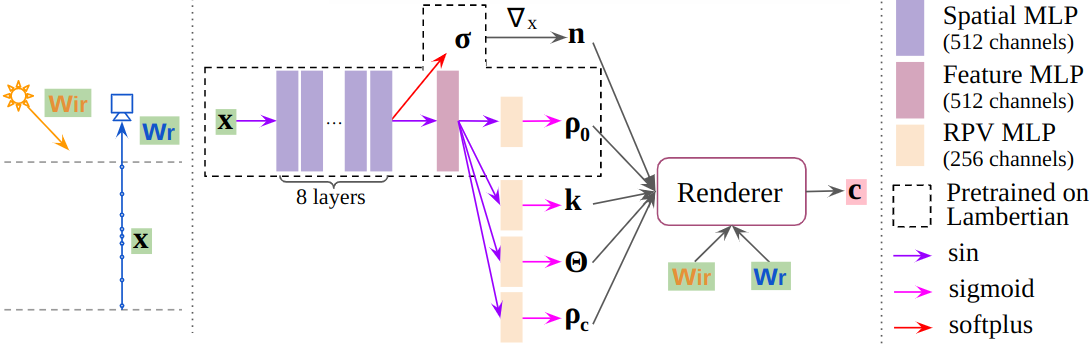}
		\caption{\textbf{\OurNeRFShort~Architecture}. The input 3D locations $\textbf{x}$ are sampled along the camera rays and fed into \OurNeRFShort~to query density and RPV parameters. 
  \OurNeRFShort~consists of shared 8-layer spatial MLPs, 1-layer feature MLP and four 1-layer RPV MLPs. The 8-layer MLPs are followed by a softplus function to predict the density $\sigma$, which is further used to calculate the analytical normal $\textbf{n}$. The RPV MLPs are concatenated with a sigmoid function to predict $\bm{\rho_0}$, $\textbf{k}$, $\bm{\Theta}$ and $\bm{\rho_c}$. The elements within the dashed rectangle are pre-trained on the assumption of a Lambertian surface for the first 20\% of the total training steps. The colour $\textbf{c}$ is predicted with the RPV rendering equation in \Cref{renderingequation}, where $W_{ir}$ and $W_r$ represent the Sun and camera directions, respectively.
  }
  \label{architecture}
	\end{center}
\end{figure}
The network architecture presented in \Cref{architecture} consists of two progressively trained components: the geometric part and the RPV part. Initially, the geometric part is pre-trained on the assumption of a Lambertian surface. After this pre-training phase (i.e., $\sim$20\% of the total duration of the training time), the RPV part is introduced. Three MLPs predicting $\textbf{k}$, $\bm{\Theta}$ and $\bm{\rho_c}$ as well as the analytical normal \textbf{n} engage in training, while the albedo $\bm{\rho_0}$ is finetuned to match the pseudo-albedo of our new rendering equation (see \Cref{renderingequation}). The separation of the geometry part from the RPV part, which handles the case of non-Lambertian surfaces, ensures that the final training stage works on well-initialised normal vectors.



The input spatial locations $\textbf{x}$ are transformed by positional encoding, the activations for $\bm{\rho_0}$, $\textbf{k}$, $\bm{\Theta}$ and $\bm{\rho_c}$ are sigmoid function, and the $\textbf{k}$ and $\bm{\Theta}$ are scaled to [0, 2] and [-1, 1] to match their real value ranges {\citep{koukal2014evaluation}}.
Ultimately, all the parameters of the \OurNeRFShort~are optimized to minimise the combination of (1) the colour loss between the ground truth pixel colour $\textbf{$\overline{\mathbf{C}}$}(\textbf{r})$ and the predicted pixel colours $\textbf{C}(\textbf{r})$, and (2) the depth loss $\mathcal{L}_{depth}(\textbf{r})$ (\Cref{depthloss}):
\begin{equation}
    \mathcal{L} = \sum_{\textbf{r} \in R} \left \| \textbf{C}(\textbf{r}) - \textbf{$\overline{\mathbf{C}}$}(\textbf{r}) \right \| _2 ^2 + \lambda \mathcal{L}_{depth}(\textbf{r})~,
    \label{loss}
\end{equation}
where $\lambda$ is a weight balancing the contribution of colour and depth. See our experiment on finding optimal $\lambda$ in \Cref{Findthebesttrainingstrategy}. 

\section{Numerical experiments}
We conduct a series of experiments to evaluate our \OurNeRFShort~on novel view synthesis (\Cref{Novelviewsynthesis}) and altitude estimation (\Cref{Altitudeestimation}) tasks. In addition, we examine the influence of atmospheric correction {from TOA (Top Of Atmosphere) to TOC (Top Of Canopy)} (\Cref{atmosphericcorrection}) and carry out ablation studies to determine the best training strategy (\Cref{Findthebesttrainingstrategy}), and the most optimal way of rendering (\Cref{sub:rendering}).
    
\paragraph{Evaluation Metrics} Precision metrics are Peak Signal-to-Noise Ratio (PSNR) and Structural Similarity Index measure (SSIM) \citep{wang2004image} for view synthesis, and Mean Altitude Error (MAE) for altitude extraction. Ground truth (GT) images are \textit{true} images not seen during training, while the GT surface is a DSM generated with high-resolution Pleiades panchromatic images, with a ground sampling distance ($GSD$) of 0.5 m, and stereo image matching \citep{mpd:06:sgm}. \OurNeRFShort~is also compared to competitions \SatNeRF~\citep{mari2022sat}, \SpSNeRF~\citep{zhang2023spsnerf1} {as well as \acc{DSM}s generated with \acc{SGM} using full-resolution images (i.e., $SGM_{Z1}$ with $GSD=2$m)}. The source code for EO-NeRF \citep{Mari_2023_CVPR}  was not available for comparison. The RPV parameters are indirectly validated by examining quantitative metrics for novel view synthesis. Additionally, one can compare the BRF in \Cref{BRF_circle} (b) with an independent result derived from 21 Pléiades images in \citep{labarre2019retrieving}.

\subsection{Implementation Details} 

\paragraph{Training} Our network is trained with the Adam Optimizer (lr=5e-4, decay=0.9, batch size=1024). We use \SpSNeRF's ray sampling strategy to sample 64 stratified points along each ray, accompanied with 64 guided points following a Gaussian distribution. We optimize \OurNeRFShort~for 100k iterations, which takes $\sim$10 hours on NVIDIA GPU with 40GB RAM. For fair comparison, the competitive methods {(i.e., \SatNeRF~and \SpSNeRF)} are also trained for 100k iterations with 64 + 64 points along each ray, which takes $\sim$6 hours. \OurNeRFShort~is less efficient in training than competitive methods, mainly due to the normal analytical computation. Computational efficiency was not the aim of this work and could be improved in the future with techniques such as tensor decomposition.

\paragraph{\Lightvi} It is important to take into account the visibility of samples when scenes contain occlusions, for example when acquiring images in mountainous or urban areas. Visibility describes the transmission of a sample between the light source and the query point. Brute-force computation of the \lightvi~is computationally expensive, as it requires marching rays from all the query points along the camera ray to the light source. Previous work treats the \lightvi~with different strategies: (1) assuming that \lightvi~is the same everywhere \citep{mai2023neural, boss2021nerd}; in this scenario, a shadow is embedded into the albedo colour; (2) with a fully analytical approach provided that the camera-light configuration is collocated, i.e., the light rays are aligned with the camera rays \citep{bi2020neural}; (3) with a semi-analytical approach where visibility is calculated by ray tracing from the light source to the estimated surface \citep{Zhang_2021, li2022neural, yang2022s, Mari_2023_CVPR}; (4) with MLP-learnt visibility,
supervised by photometric loss \citep{verbin2022refnerf}, or by \lightvi~calculated by ray tracing \citep{srinivasan2020nerv, yang2022psnerf, derksen2021shadow, mari2022sat}.

The fully analytical strategy is intractable in situations where light and view are not collocated, while the semi-analytical strategy {can lead} to noisy results, especially when few input views are used. The learning-based approach gives the network superfluous freedom that could confuse \lightvi~with other phenomena, such as albedo colour. Since our main objective is to model natural surfaces, such as bare soil, from aerial images, we  can safely assume that the \lightvi~is constant everywhere, i.e., equal to 1.


\subsection{Dataset}
\begin{figure}[htb!]
	\begin{center}
		\subfigure[Djibouti]{
			\begin{minipage}[t]{0.430\linewidth}
				\centering
				\includegraphics[width=1.0\linewidth]{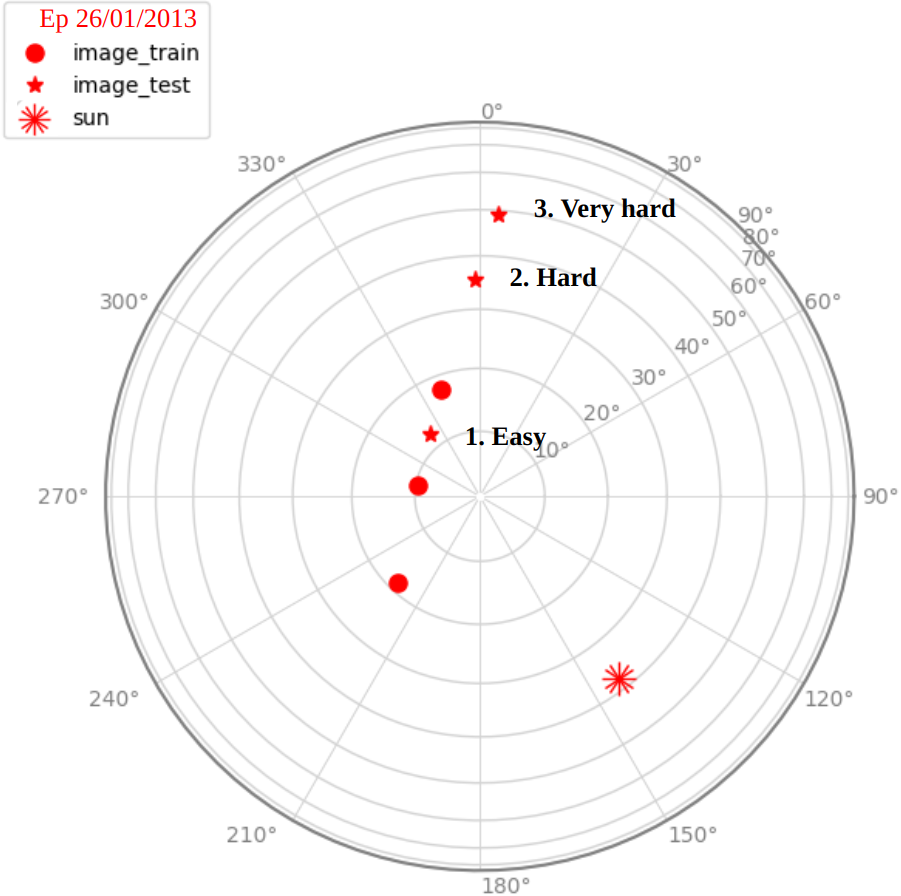}
			\end{minipage}%
		}
		\subfigure[Lanzhou]{
			\begin{minipage}[t]{0.430\linewidth}
				\centering
				\includegraphics[width=1.0\linewidth]{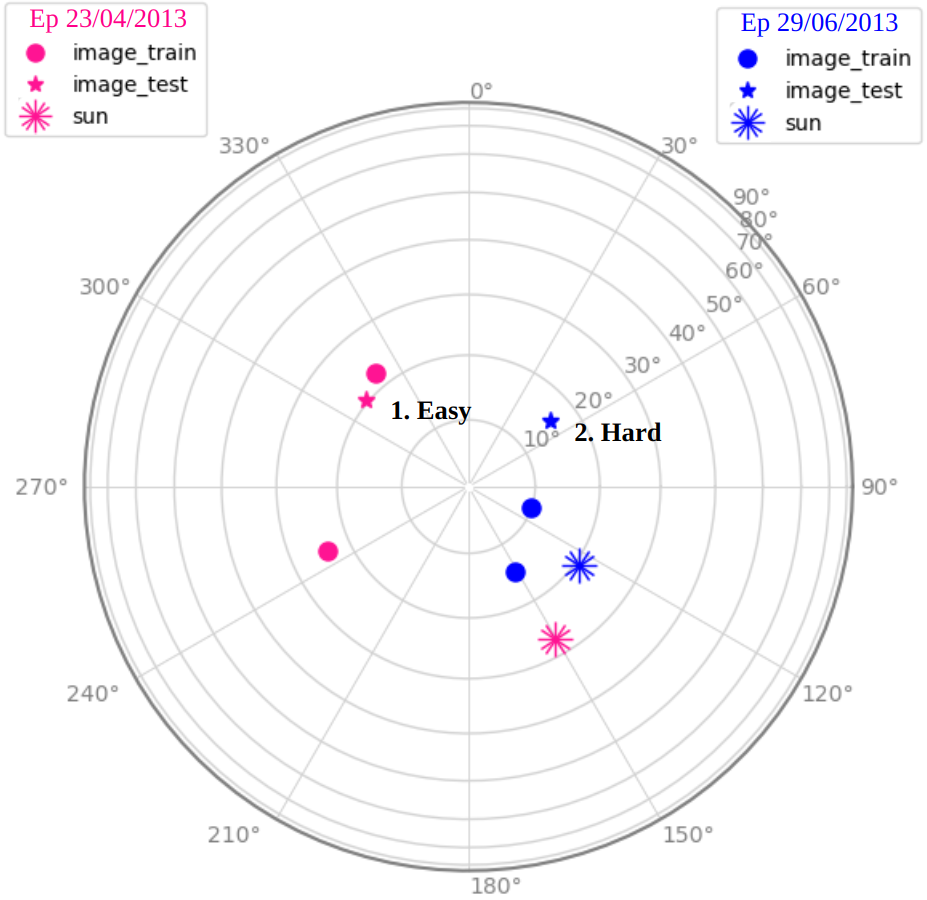}
			\end{minipage}%
		}
		\caption{\textbf{Testing Scenarios -- Sun and Viewing Angles.}  
  The angles are displayed in polar coordinate systems, where the radius and angular coordinate represent zenith and azimuth, respectively. We design up to three test scenarios, which differ in the viewing angle of the test images. \textit{Easy} corresponds to a case where the test images are interpolated from the training images, while \textit{hard} and \textit{very hard} correspond to extrapolation. Three training images are used in Djibouti, four in Lanzhou (two of each epoch).}
		\label{sun_view_angle}
	\end{center}
\end{figure}

Evaluations are carried out on two sites (Djibouti, Lanzhou). We extract two regions of interest ($\sim$ 1.5 km $\times$ 1.5 km) in each site, which we refer to as $A$ and $B$ (e.g., Dji-A and Dji-B). In each dataset, we distinguish three test scenarios ranging from \textit{easy} (novel view interpolation) to \textit{very hard} (novel view extrapolation). The input images are RGB with a GSD of 2m and undergo atmospheric correction prior to processing (see \Cref{atmosphericcorrection}). Sadly, we could not perform experiments on the open DFC dataset \citep{bosch2019semantic} representing urban scenes since RPV is designed for natural surfaces.
\paragraph{Djibouti Dataset} It is located in the Asal-Ghoubbet rift (Republic of Djibouti) \citep{labarre2019retrieving} and consists of 21 multiangular Pleiades 1B images collected in a single flyby on January 26, 2013. Three quasi-nadir images are chosen for training, and three other images for testing, including interpolation and extrapolation scenarios (\Cref{sun_view_angle} (a)). 
\paragraph{Lanzhou Dataset}  It is located in Lanzhou (China) and consists of 3 Pleiades 1B images acquired on April 23, 2013 and 3 Pleiades 1A images acquired on June 29, 2013. This is a multi-date dataset where the position of the Sun changes between acquisitions (\Cref{sun_view_angle} (b)). 

\subsection{The effect of \ac}
\label{atmosphericcorrection}
\begin{figure}[htb!]
    \begin{center}
        \subfigure[Without atmospheric correction]{
            \begin{minipage}[t]{0.212\linewidth}
                \centering
                \includegraphics[width=0.8\linewidth]{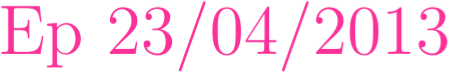}
                \includegraphics[width=1.0\linewidth]{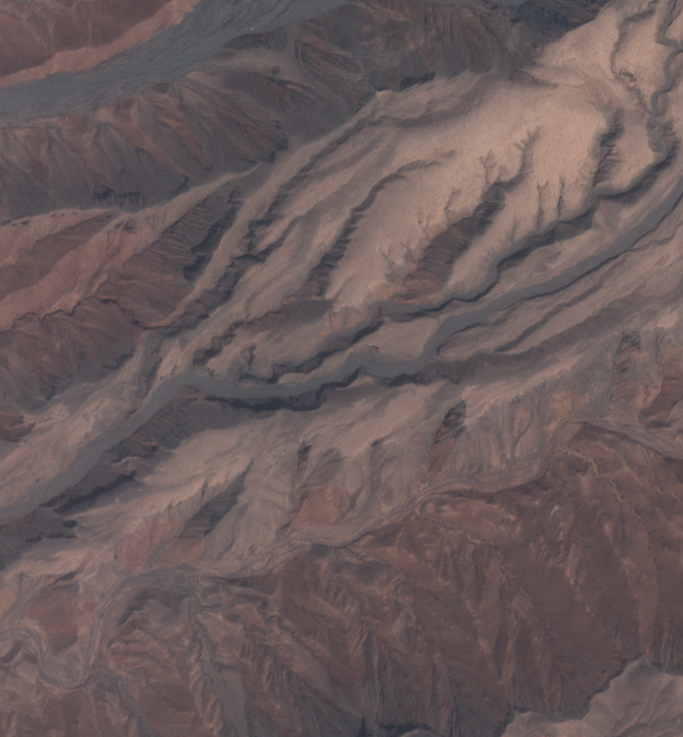}
            \end{minipage}%
            \hspace{0mm}
            \begin{minipage}[t]{0.225\linewidth}
                \centering
                \includegraphics[width=0.8\linewidth]{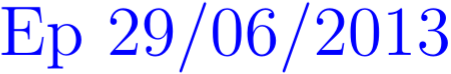}                
                \includegraphics[width=1.0\linewidth]{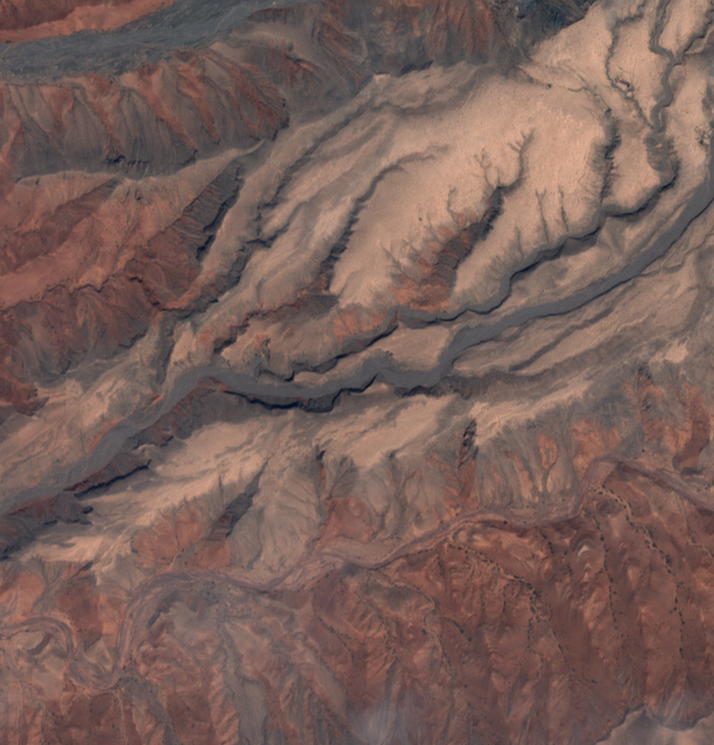}
                \end{minipage}%
        }
        \subfigure[With atmospheric correction]{
            \begin{minipage}[t]{0.212\linewidth}
                \centering
                \includegraphics[width=0.8\linewidth]{ep04.png}                
                \includegraphics[width=1.0\linewidth]{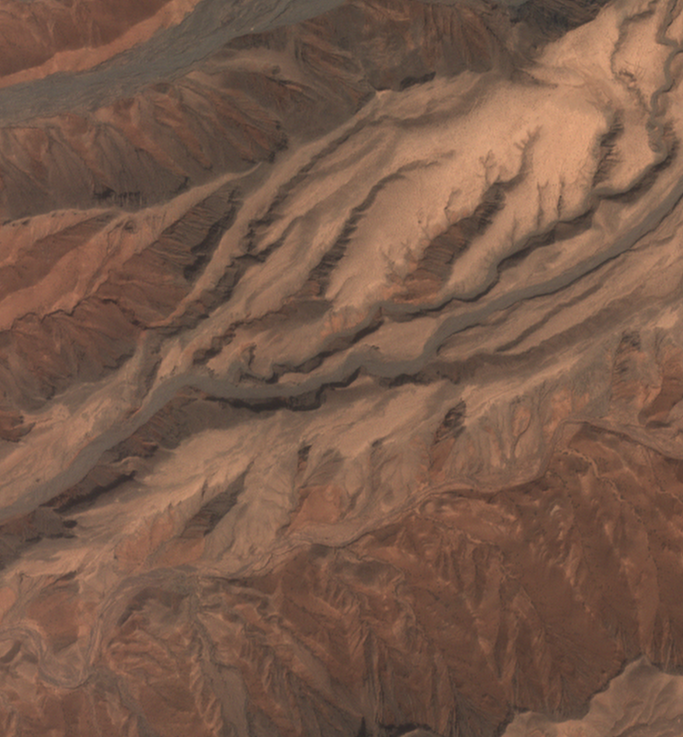}
            \end{minipage}%
            \hspace{0mm}
            \begin{minipage}[t]{0.225\linewidth}
                \centering
                \includegraphics[width=0.8\linewidth]{ep06.png}                                
                \includegraphics[width=1.0\linewidth]{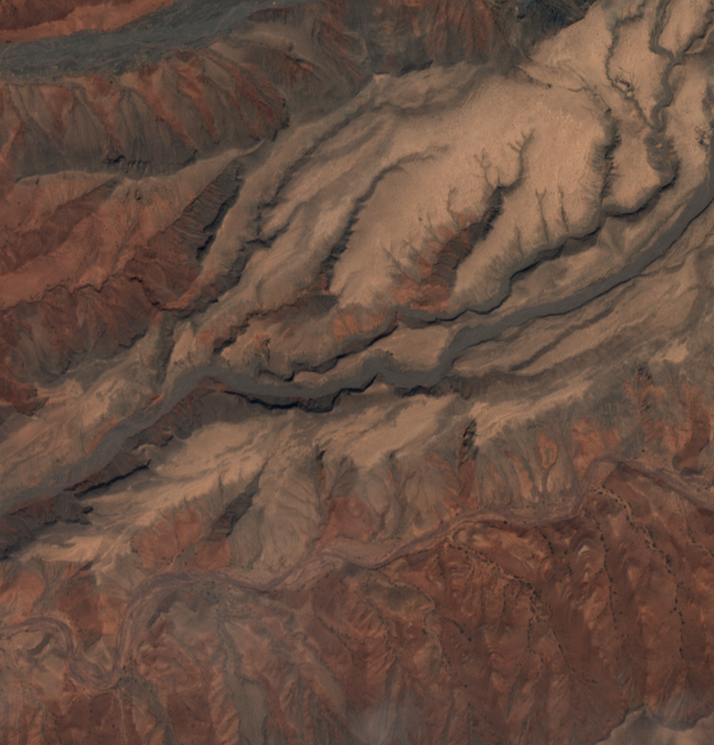}
                \end{minipage}%
        }
        \caption{\textbf{Atmospheric Correction -- Visualisations.} Comparison between {input} images without and with atmospheric correction for the Lzh-A multi-date dataset. The similarity of images tones between different epochs is improved after \ac.}
        \label{compatmo}
    \end{center}
\end{figure}

\begin{figure}[!t]
    \begin{center}
        \subfigure[\SpSNeRF]{
            \begin{minipage}[t]{0.259\linewidth}
                \centering
                \includegraphics[width=1.0\linewidth]{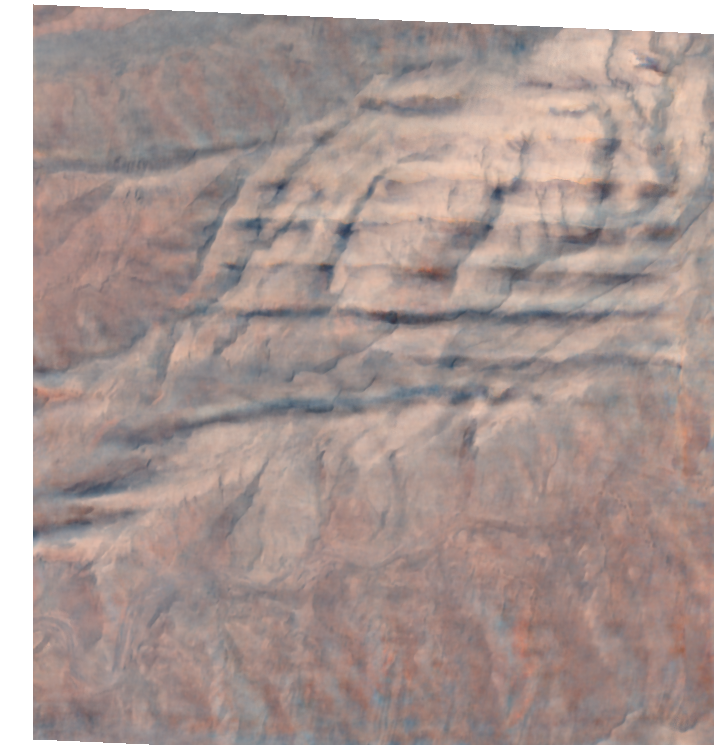}
            \end{minipage}%
        }
        \subfigure[\OurNeRFShort]{
            \begin{minipage}[t]{0.259\linewidth}
                \centering
                \includegraphics[width=1.0\linewidth]{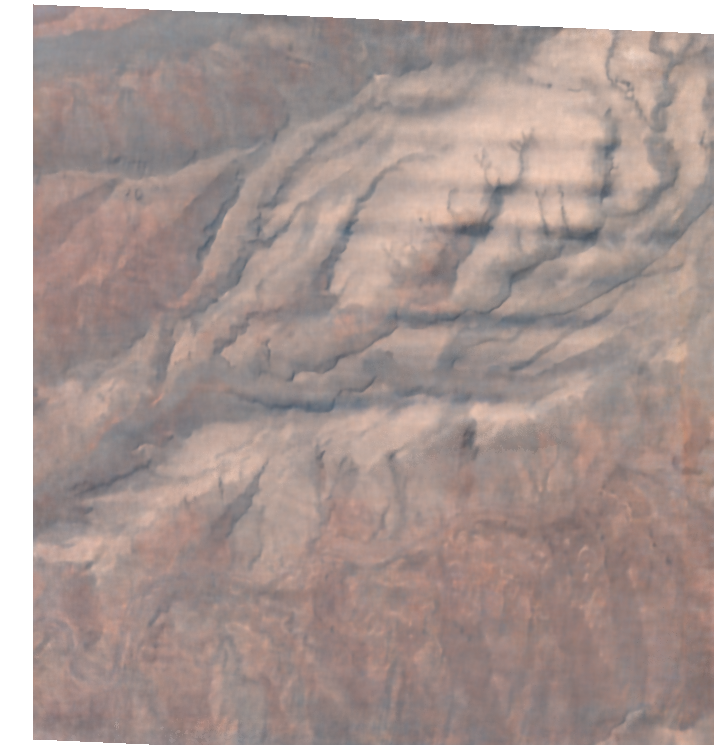}
            \end{minipage}%
        }
        \subfigure[GT]{
            \begin{minipage}[t]{0.259\linewidth}
                \centering
                \includegraphics[width=1.0\linewidth]{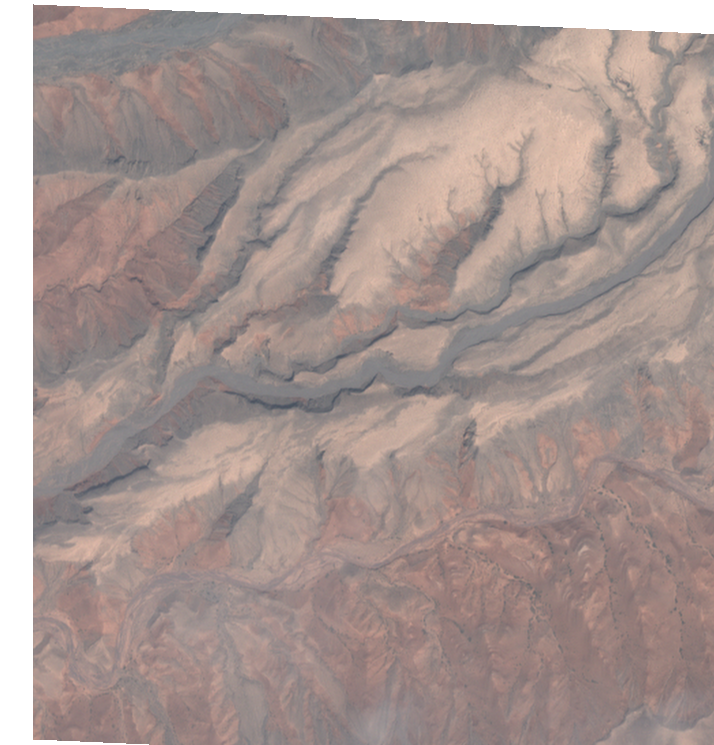}
            \end{minipage}%
        }
        \subfigure[\SpSNeRF]{
            \begin{minipage}[t]{0.259\linewidth}
                \centering
                \includegraphics[width=1.0\linewidth]{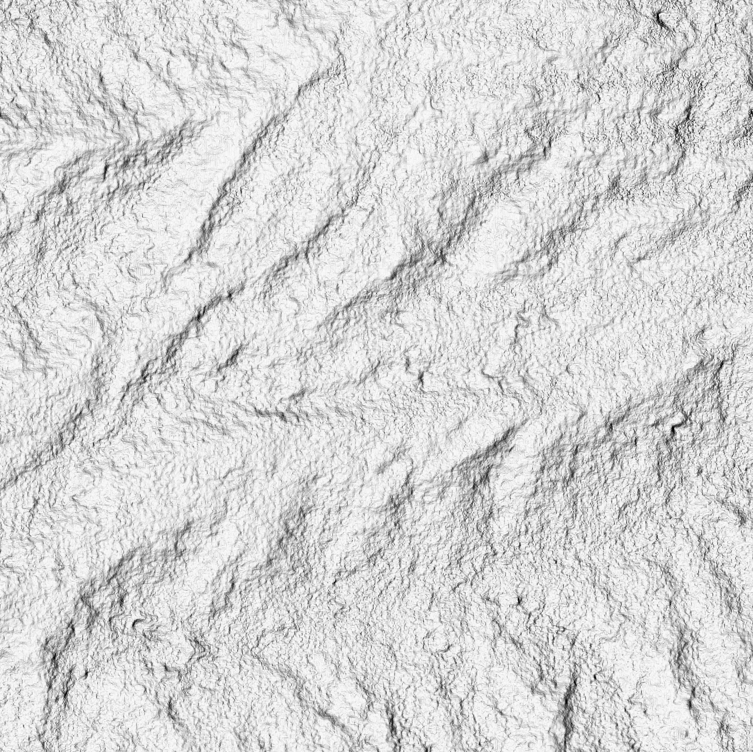}
            \end{minipage}%
        }
        \subfigure[\OurNeRFShort]{
            \begin{minipage}[t]{0.259\linewidth}
                \centering
                \includegraphics[width=1.0\linewidth]{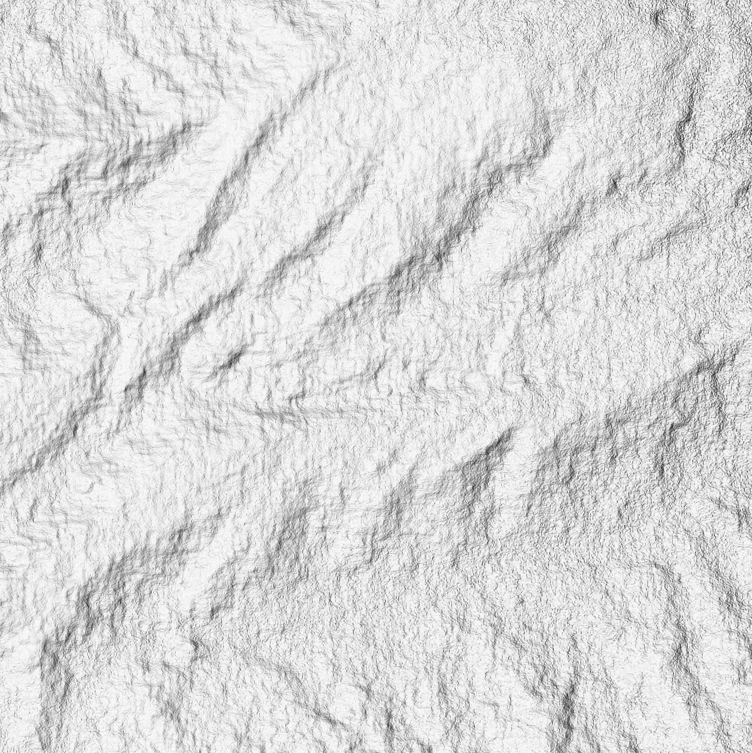}
            \end{minipage}%
        }
        \subfigure[GT]{
            \begin{minipage}[t]{0.259\linewidth}
                \centering
                \includegraphics[width=1.0\linewidth]{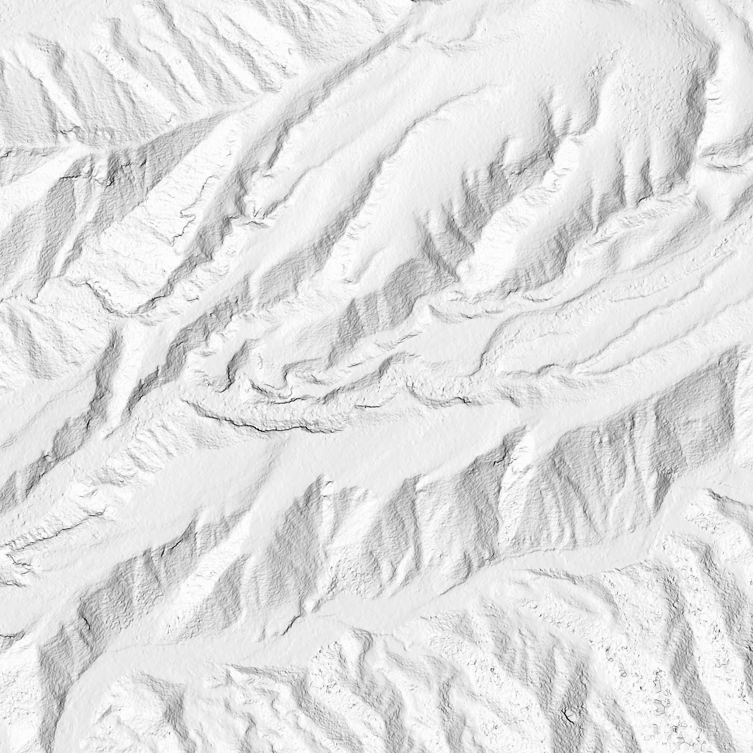}
            \end{minipage}%
        }
        \caption{\textbf{Ablation of Atmospheric Correction -- View Synthesis and Altitude Estimation Visualisation}. Results on the Lzh-A site using images without \ac~(\redcross \textbf{\acshort}). The view synthesis is displayed in the first line, and the altitude estimation in the second. \OurNeRFShort~synthesises images with fewer artifacts than \SpSNeRF~and estimates surface closer to GT. Performance is further improved on images with \acshort~(\Cref{maintestnovalLzh} (e) and \Cref{maintestaltLzh} (e)).}
        \label{figwoac}
    \end{center}
\end{figure}

Atmospheric correction is important for two reasons. Firstly, the atmosphere affects the signal received by the satellite sensor. Secondly, the RPV model, like all BRDF models, describes TOC reflectances whereas Pleiades images are typically supplied as calibrated at TOA. We apply the atmospheric correction using an Orfeo ToolBox (OTB) software library \citep{OTB2002} which is based on the 6S radiative transfer code \citep{vermote1997second}. The correction model requires four atmospheric parameters: ozone content $U_{O3}$ (cm-atm), water vapor content $U_{H2O}$ (g/cm2), aerosol optical thickness $\tau_A$ (unitless) and atmospheric pressure $P_a$ (hPa). The first three parameters are estimated from ancillary datasets corresponding to the day of image acquisition, available on NASA's Earth observation website (https://neo.gsfc.nasa.gov/). The atmospheric pressure is approximated using the formula $P_a = 1013.25 \cdot (1 - 0.0065 \cdot Z / 288.15)^{5.31}$, where $Z$ is the surface altitude expressed in meter. The four parameters, along with the adjacency radius for both epochs in the Lanzhou dataset, are shown in Table \ref{acparameters}. The Djibouti dataset had been corrected for atmosphere by the satellite image provider.
\Cref{compatmo} shows an example of comparison between input images with and without performing \ac. Images tones between different epochs become more similar with \ac. 
\Cref{figwoac} and \Cref{tablewoac} compared results based on images with and without \ac~on novel view synthesis and altitude estimation. Results without \ac~gained generally worse metrics and displayed artifacts in synthetic images.



\begin{table}[tb!]
\centering
\begin{tabular}{|c|c|c|c|c|c|}
\hline
\multirow{2}*{Epoch} & $U_{O3}$ & $U_{H2O}$ & $\tau_A$ & $P_a$ & adjacency \\ 
& (cm-atm) & (g/cm$^2$) & (unitless) & (hPa) & radius (-) \\ \hline\hline
23/04/2013 & 0.3220 & 1.7333 & 0.4665 & 783 & 1.0 \\ \hline
29/06/2013 & 0.2969 & 2.5625 & 0.0980 & 783 & 1.0 \\ \hline
\end{tabular}
\caption{\textbf{Atmospheric Correction Parameters.} Values for ozone content $U_{O3}$, water vapor content $U_{H2O}$, aerosol optical thickness $\tau_A$ and atmospheric pressure $P_a$, as well as the adjacency radius used for atmospheric correction in the Lanzhou dataset.}
\label{acparameters}
\end{table}

\newcommand{\gacpsnr}[1]{\gradientcelld{#1}{24}{29}{34}{gLow}{gMid}{gHigh}{70}}
\newcommand{\gacssim}[1]{\gradientcelld{#1}{0.8}{0.89}{0.98}{gLow}{gMid}{gHigh}{70}}
\newcommand{\gacgeom}[1]{\gradientcelld{#1}{3}{3.5}{4}{gHigh}{gMid}{gLow}{70}}

\begin{table}[h!]
\centering
\begin{tabular}{|c|c|c|c|c|c|c|}
\hline
\multirow{2}*{Method} & \multirow{2}*{AC} & \multirow{2}*{MAE $\downarrow$} & \multicolumn{2}{|c}{PSNR $\uparrow$} & \multicolumn{2}{|c|}{SSIM $\uparrow$} \\ \cline{4-7}
&&&Easy&Hard&Easy&Hard\\\hline\hline
\SpSNeRF&\redcross&\gacgeom{3.697}&\gacpsnr{28.282}&\gacpsnr{26.131}&\gacssim{0.880}&\gacssim{0.858}\\\hline
\SpSNeRF&\greencheck&\gacgeom{3.558}&\gacpsnr{29.548}&\gacpsnr{24.755}&\gacssim{0.965}&\gacssim{0.900}\\\hline
\OurNeRFShort&\redcross&\gacgeom{3.439}&\textbf{\gacpsnr{33.315}}&\textbf{\gacpsnr{29.857}}&\gacssim{0.946}&\gacssim{0.923}\\\hline
\OurNeRFShort&\greencheck&\textbf{\gacgeom{3.420}}&\gacpsnr{32.196}&\gacpsnr{28.420}&\textbf{\gacssim{0.979}}&\textbf{\gacssim{0.94}}\\\hline
\end{tabular}
\caption{\textbf{Ablation of Atmospheric Correction -- Quantitative Evaluation on Lzh-A.} Experiments with \greencheck and without \redcross atmospheric correction (AC). \OurNeRFShort~\greencheck performs best, with the smallest MAE and biggest SSIM. {Although \OurNeRFShort~\redcross shows slightly higher PSNR, qualitative visualisation in \Cref{figwoac}(b) reveal stripe artefacts, which are not present in \OurNeRFShort~\greencheck in \Cref{maintestnovalLzh}(e).}}
\label{tablewoac}
\end{table}

\newpage
\subsection{Novel view synthesis}
\label{Novelviewsynthesis}
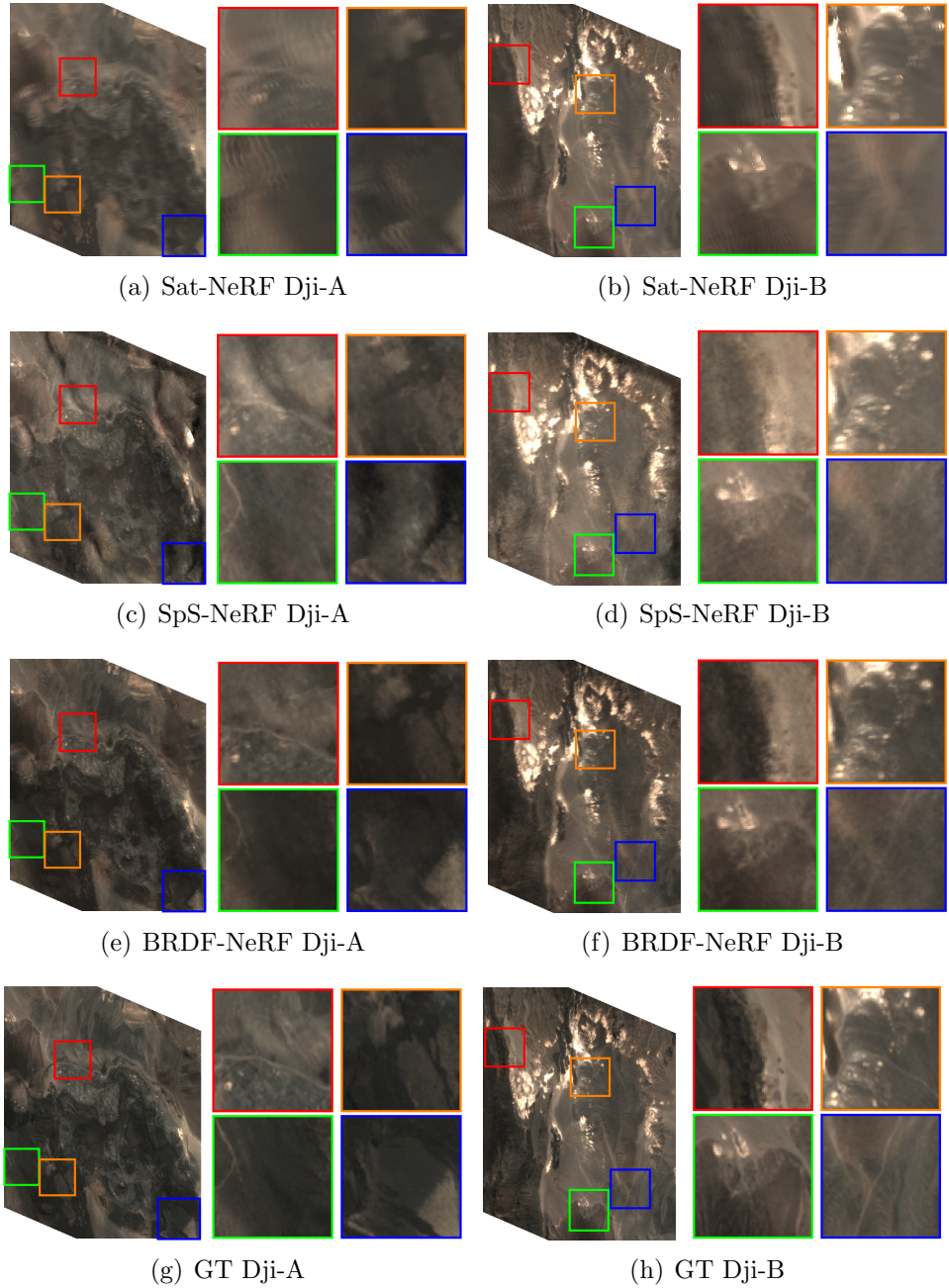
\begin{figure*}[!htbp]
    \begin{center}
        \subfigure[\SatNeRF~Dji-A]{
            \begin{minipage}[t]{0.205\linewidth}
                \centering
                    \begin{tikzpicture}
                    \node[anchor=south west,inner sep=0] (image) at (0,0) {
                \includegraphics[width=1\linewidth]{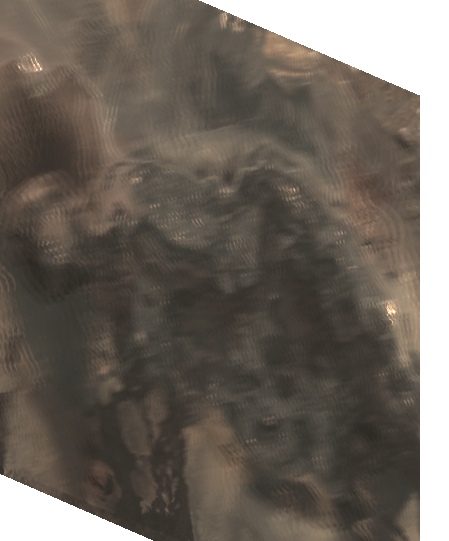}};
                    \begin{scope}[x={(image.south east)},y={(image.north west)}]
                    \draw[green,thick] (0.0,0.357) rectangle (0.165,0.214);
                    \draw[orange,thick] (0.168,0.315) rectangle (0.336,0.173);
                    \draw[red,thick] (0.240,0.635) rectangle (0.410,0.782);
                    \draw[blue,thick] (0.730,0.000) rectangle (0.930,0.160);
                    \end{scope}
                    \end{tikzpicture}
            \end{minipage}%
            \begin{minipage}[t]{0.245\linewidth}
                \centering
                \includegraphics[width=1\linewidth]{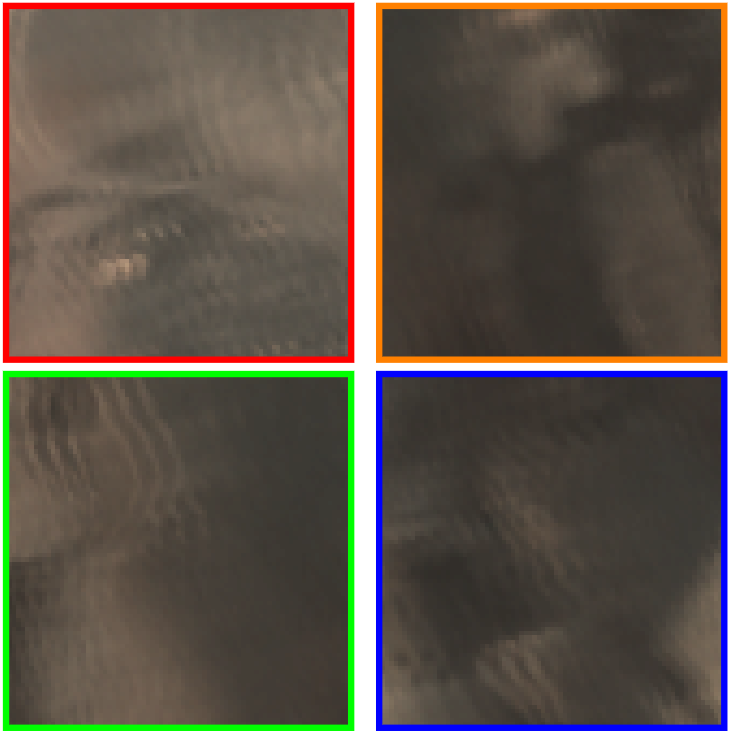}
            \end{minipage}%
        }
        \subfigure[\SatNeRF~Dji-B]{
            \begin{minipage}[t]{0.205\linewidth}
                \centering
                    \begin{tikzpicture}
                    \node[anchor=south west,inner sep=0] (image) at (0,0) {
                \includegraphics[width=1\linewidth]{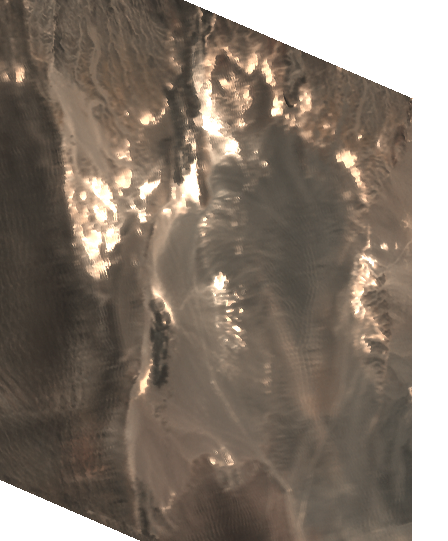}};
                    \begin{scope}[x={(image.south east)},y={(image.north west)}]
                    \draw[red,thick] (0.013,0.687) rectangle (0.199,0.838);
                    \draw[green,thick] (0.415,0.04) rectangle (0.599,0.200);
                    \draw[blue,thick] (0.61,0.128) rectangle (0.797,0.280);
                    \draw[orange,thick] (0.42,0.571) rectangle (0.605,0.72);
                    \end{scope}
                    \end{tikzpicture}
            \end{minipage}%
            \begin{minipage}[t]{0.245\linewidth}
                \centering
                \includegraphics[width=1\linewidth]{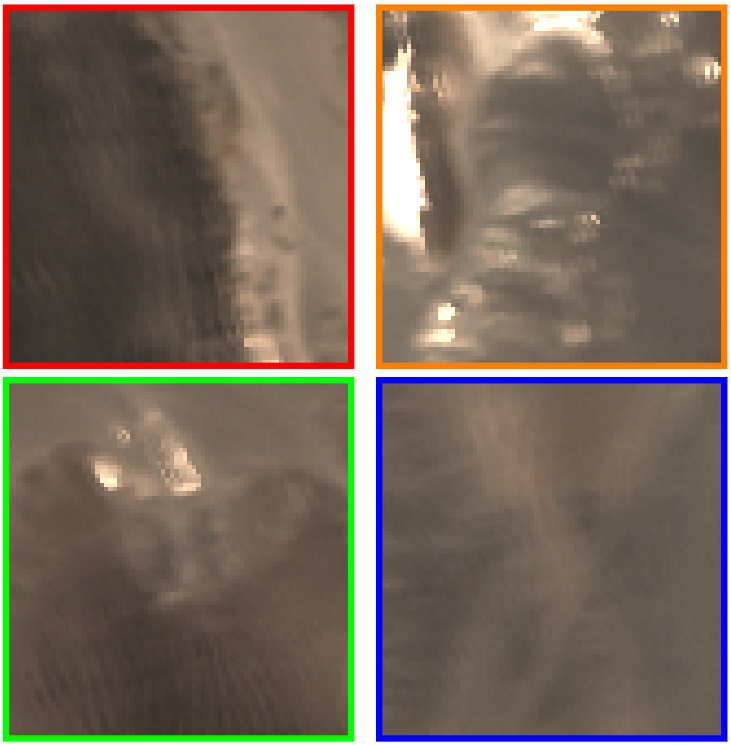}
            \end{minipage}%
        }
        \subfigure[\SpSNeRF~Dji-A]{
            \begin{minipage}[t]{0.205\linewidth}
                \centering
                    \begin{tikzpicture}
                    \node[anchor=south west,inner sep=0] (image) at (0,0) {
                \includegraphics[width=1\linewidth]{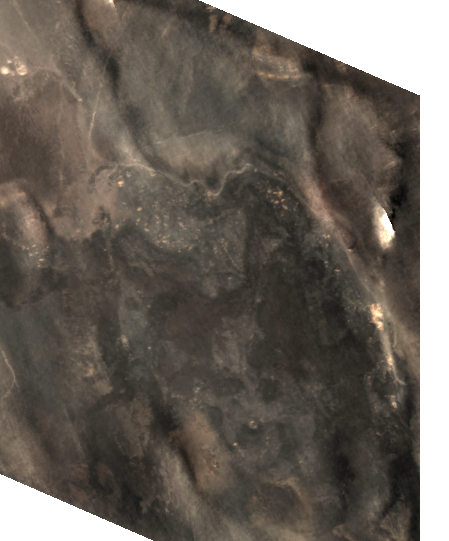}};
                    \begin{scope}[x={(image.south east)},y={(image.north west)}]
                    \draw[green,thick] (0.0,0.357) rectangle (0.165,0.214);
                    \draw[orange,thick] (0.168,0.315) rectangle (0.336,0.173);
                    \draw[red,thick] (0.240,0.635) rectangle (0.410,0.782);
                    \draw[blue,thick] (0.730,0.000) rectangle (0.930,0.160);
                    \end{scope}
                    \end{tikzpicture}
            \end{minipage}%
            \begin{minipage}[t]{0.245\linewidth}
                \centering
                \includegraphics[width=1\linewidth]{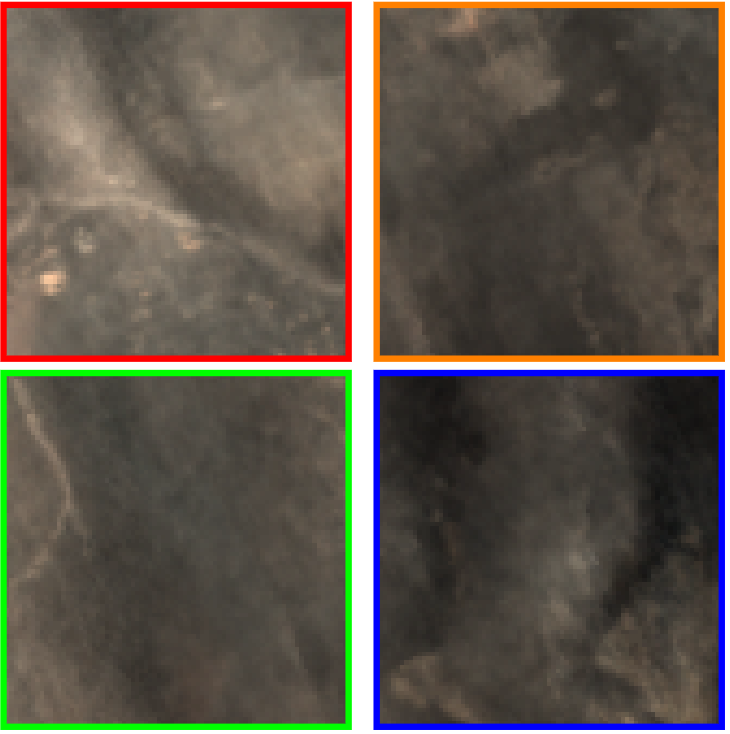}
            \end{minipage}%
        }
        \subfigure[\SpSNeRF~Dji-B]{
            \begin{minipage}[t]{0.205\linewidth}
                \centering
                    \begin{tikzpicture}
                    \node[anchor=south west,inner sep=0] (image) at (0,0) {
                \includegraphics[width=1\linewidth]{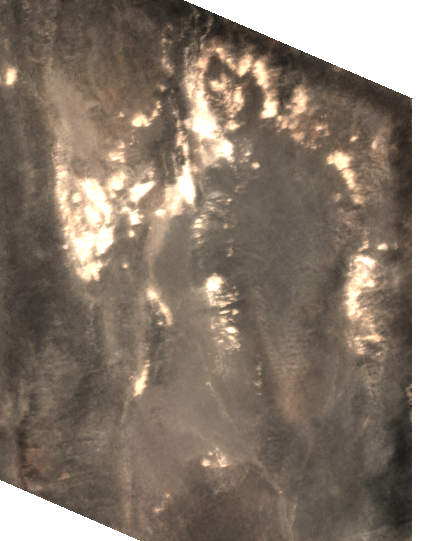}};
                    \begin{scope}[x={(image.south east)},y={(image.north west)}]
                    \draw[red,thick] (0.013,0.687) rectangle (0.199,0.838);
                    \draw[green,thick] (0.415,0.04) rectangle (0.599,0.200);
                    \draw[blue,thick] (0.61,0.128) rectangle (0.797,0.280);
                    \draw[orange,thick] (0.42,0.571) rectangle (0.605,0.72);
                    \end{scope}
                    \end{tikzpicture}
            \end{minipage}%
            \begin{minipage}[t]{0.245\linewidth}
                \centering
                \includegraphics[width=1\linewidth]{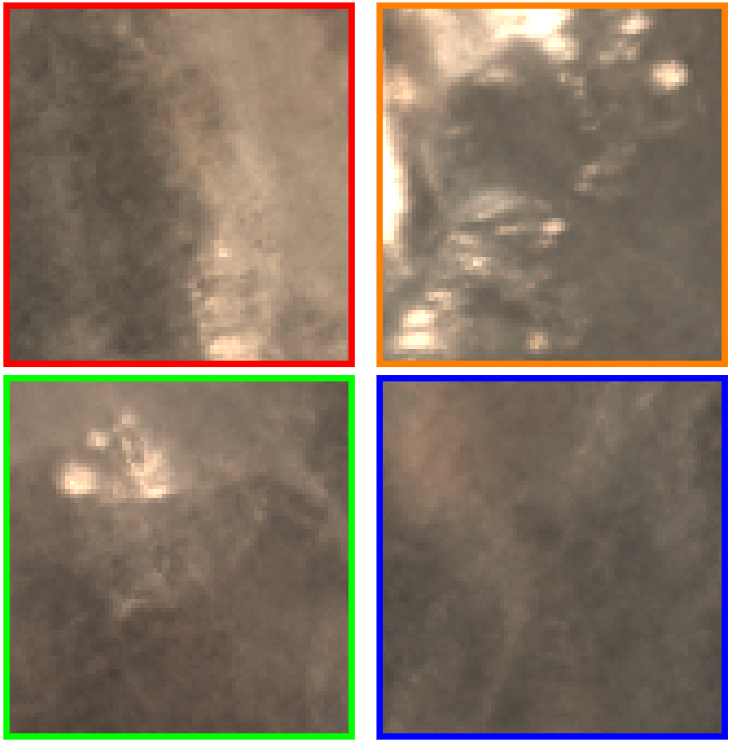}
            \end{minipage}%
        }
        \subfigure[\OurNeRFShort~Dji-A]{
            \begin{minipage}[t]{0.205\linewidth}
                \centering
                    \begin{tikzpicture}
                    \node[anchor=south west,inner sep=0] (image) at (0,0) {
                \includegraphics[width=1\linewidth]{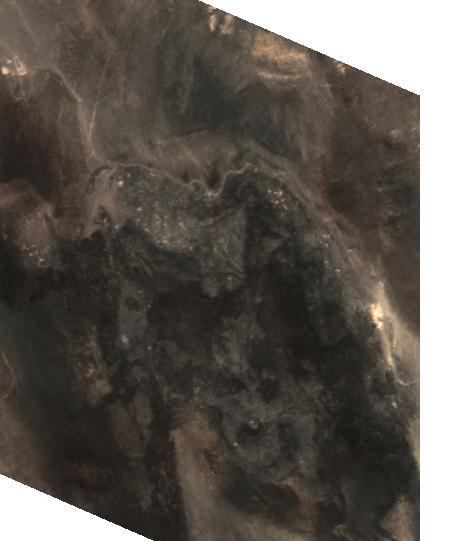}};
                    \begin{scope}[x={(image.south east)},y={(image.north west)}]
                    \draw[green,thick] (0.0,0.357) rectangle (0.165,0.214);
                    \draw[orange,thick] (0.168,0.315) rectangle (0.336,0.173);
                    \draw[red,thick] (0.240,0.635) rectangle (0.410,0.782);
                    \draw[blue,thick] (0.730,0.000) rectangle (0.930,0.160);
                    \end{scope}
                    \end{tikzpicture}
            \end{minipage}%
            \begin{minipage}[t]{0.245\linewidth}
                \centering
                \includegraphics[width=1\linewidth]{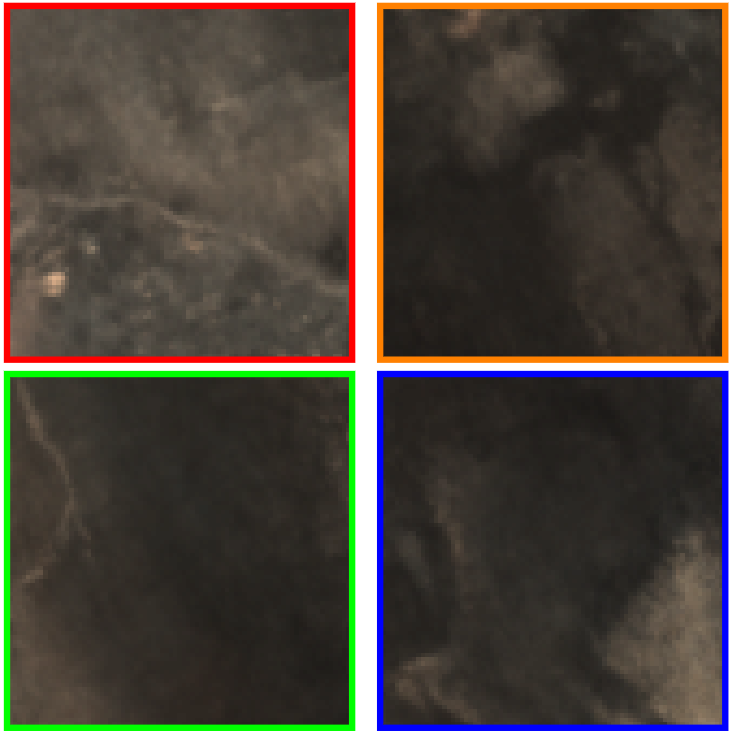}
            \end{minipage}%
        }
        \subfigure[\OurNeRFShort~Dji-B]{
            \begin{minipage}[t]{0.205\linewidth}
                \centering
                    \begin{tikzpicture}
                    \node[anchor=south west,inner sep=0] (image) at (0,0) {
                \includegraphics[width=1\linewidth]{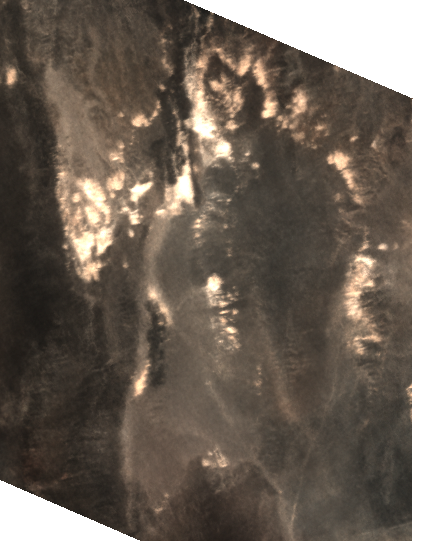}};
                    \begin{scope}[x={(image.south east)},y={(image.north west)}]
                    \draw[red,thick] (0.013,0.687) rectangle (0.199,0.838);
                    \draw[green,thick] (0.415,0.04) rectangle (0.599,0.200);
                    \draw[blue,thick] (0.61,0.128) rectangle (0.797,0.280);
                    \draw[orange,thick] (0.42,0.571) rectangle (0.605,0.72);
                    \end{scope}
                    \end{tikzpicture}
            \end{minipage}%
            \begin{minipage}[t]{0.245\linewidth}
                \centering
                \includegraphics[width=1\linewidth]{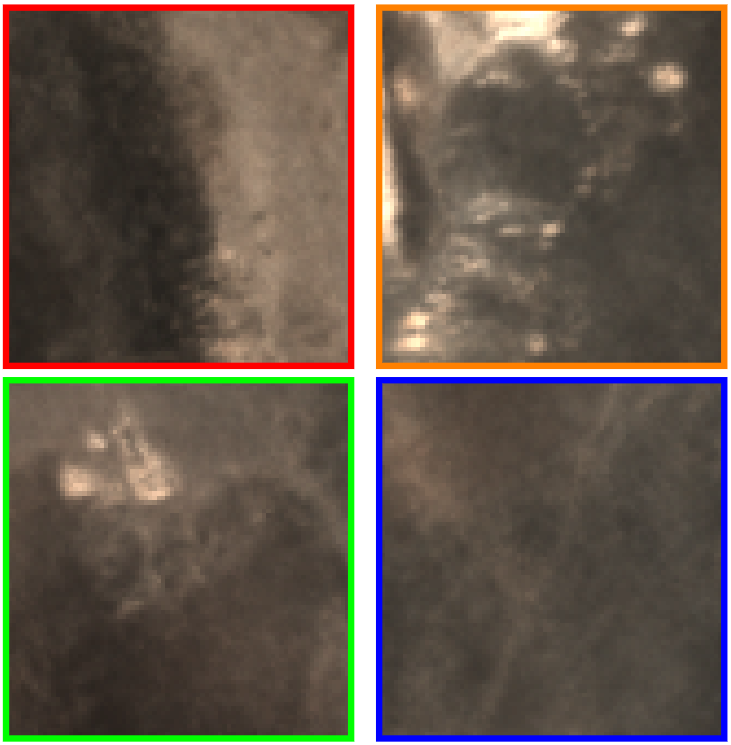}
            \end{minipage}%
        }
        \subfigure[GT Dji-A]{
            \begin{minipage}[t]{0.205\linewidth}
                \centering
                    \begin{tikzpicture}
                    \node[anchor=south west,inner sep=0] (image) at (0,0) {
                \includegraphics[width=1\linewidth]{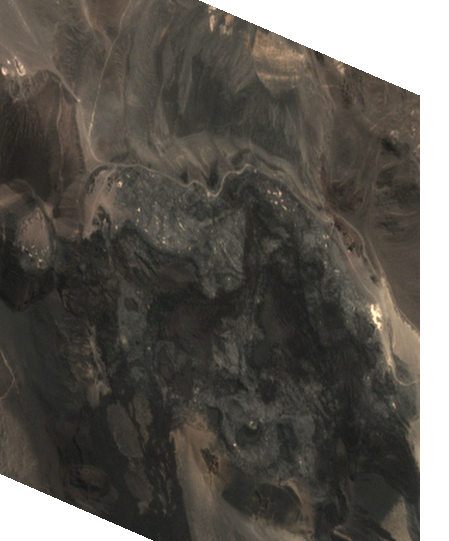}};
                    \begin{scope}[x={(image.south east)},y={(image.north west)}]
                    \draw[green,thick] (0.0,0.357) rectangle (0.165,0.214);
                    \draw[orange,thick] (0.168,0.315) rectangle (0.336,0.173);
                    \draw[red,thick] (0.240,0.635) rectangle (0.410,0.782);
                    \draw[blue,thick] (0.730,0.000) rectangle (0.930,0.160);
                    \end{scope}
                    \end{tikzpicture}
            \end{minipage}%
            \begin{minipage}[t]{0.245\linewidth}
                \centering
                \includegraphics[width=1\linewidth]{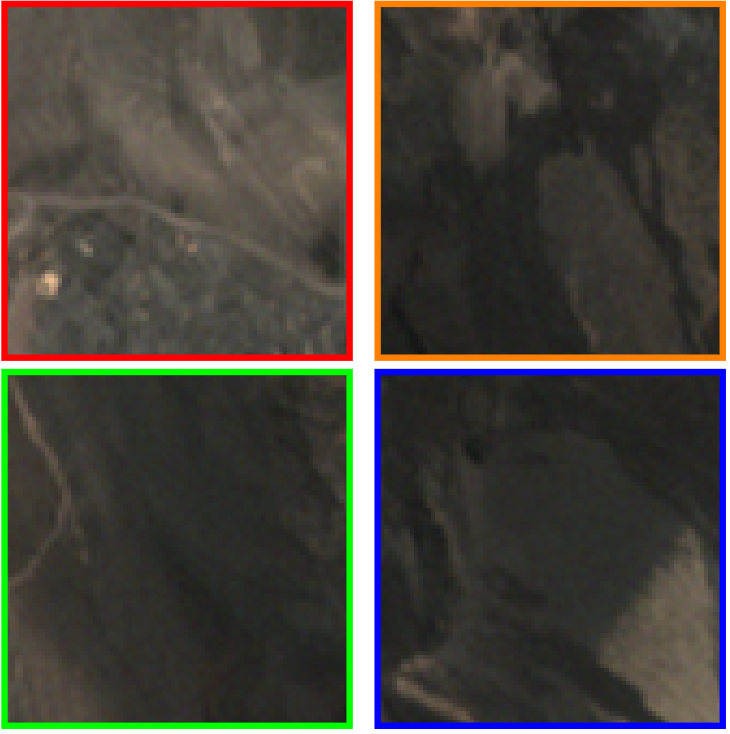}
            \end{minipage}%
        }
        \subfigure[GT Dji-B]{
            \begin{minipage}[t]{0.205\linewidth}
                \centering
                    \begin{tikzpicture}
                    \node[anchor=south west,inner sep=0] (image) at (0,0) {
                \includegraphics[width=1\linewidth]{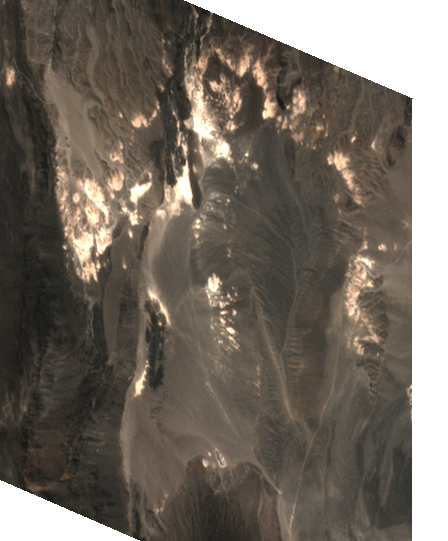}};
                    \begin{scope}[x={(image.south east)},y={(image.north west)}]
                    \draw[red,thick] (0.013,0.687) rectangle (0.199,0.838);
                    \draw[green,thick] (0.415,0.04) rectangle (0.599,0.200);
                    \draw[blue,thick] (0.61,0.128) rectangle (0.797,0.280);
                    \draw[orange,thick] (0.42,0.571) rectangle (0.605,0.72);
                    \end{scope}
                    \end{tikzpicture}
            \end{minipage}%
            \begin{minipage}[t]{0.245\linewidth}
                \centering
                \includegraphics[width=1\linewidth]{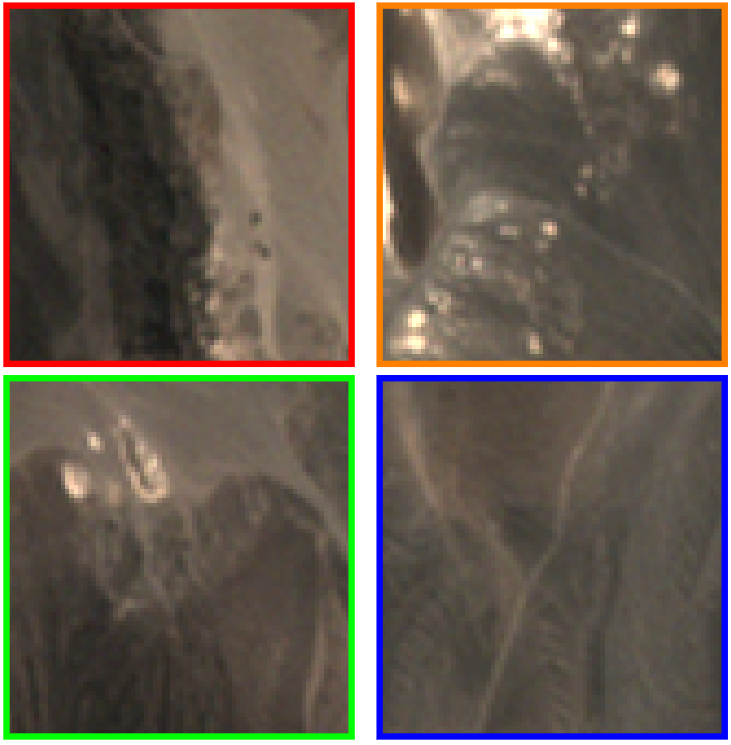}
            \end{minipage}%
        }
        \caption{\textbf{Novel View Synthesis -- Djibouti Dataset.} Renderings of sites A and B of the Djibouti dataset in the \textit{very hard} scenario (\Cref{sun_view_angle}). \SatNeRF~and \SpSNeRF~renderings are less sharp and detailed than \OurNeRFShort. Note that the colour tone of \OurNeRFShort~is closest to that of ground truth (GT).}
        \label{maintestnovalDji}
    \end{center}
\end{figure*}

\begin{figure*}[!htbp]
    \begin{center}
            \subfigure[\SatNeRF~Lzh-A]{
            \begin{minipage}[t]{0.205\linewidth}
                \centering
                    \begin{tikzpicture}
                    \node[anchor=south west,inner sep=0] (image) at (0,0) {
                \includegraphics[width=1\linewidth]{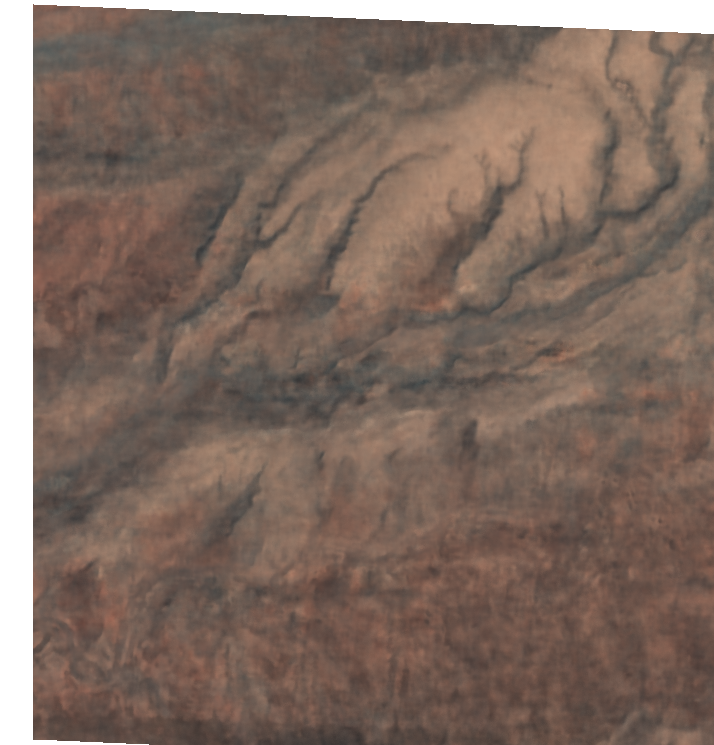}};
                    \begin{scope}[x={(image.south east)},y={(image.north west)}]
                    \draw[red,thick] (0.296,0.442) rectangle (0.504,0.656);
                    \draw[orange,thick] (0.575,0.564) rectangle (0.787,0.756);
                    \draw[green,thick] (0.090,0.380) rectangle (0.293,0.180);
                    \draw[blue,thick] (0.254,0.000) rectangle (0.465,0.198);
                    \end{scope}
                    \end{tikzpicture}
            \end{minipage}%
            \hspace{0mm}
            \begin{minipage}[t]{0.245\linewidth}
                \centering
                \includegraphics[width=1\linewidth]{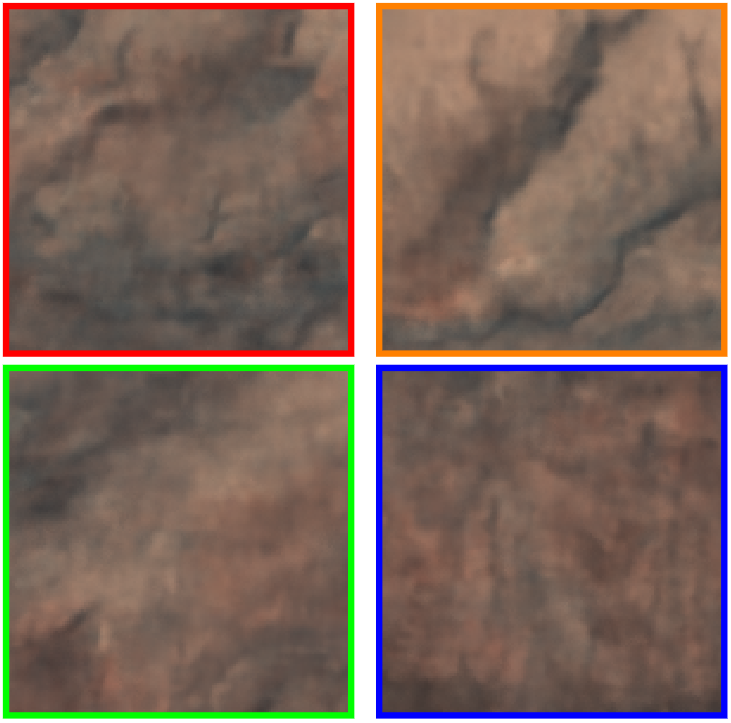}
            \end{minipage}%
        }
        \subfigure[\SatNeRF~Lzh-B]{
            \begin{minipage}[t]{0.205\linewidth}
                \centering
                    \begin{tikzpicture}
                    \node[anchor=south west,inner sep=0] (image) at (0,0) {
                \includegraphics[width=1\linewidth]{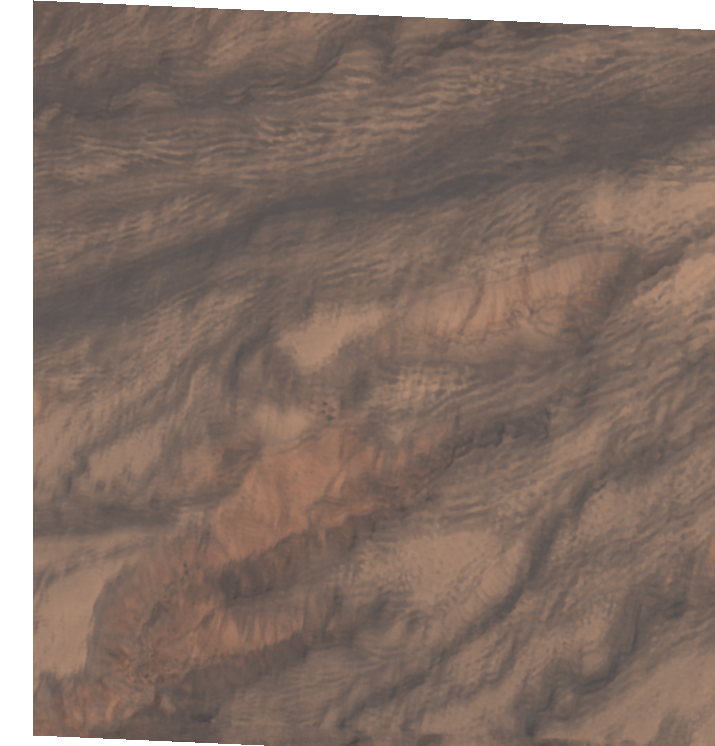}};
                    \begin{scope}[x={(image.south east)},y={(image.north west)}]
                    \draw[red,thick] (0.078,0.700) rectangle (0.307,0.924);
                    \draw[orange,thick] (0.742,0.693) rectangle (0.948,0.888);
                    \draw[green,thick] (0.053,0.196) rectangle (0.262,0.412);
                    \draw[blue,thick] (0.778,0.060) rectangle (0.99,0.258);
                    \end{scope}
                    \end{tikzpicture}
            \end{minipage}%
            \hspace{0mm}
            \begin{minipage}[t]{0.245\linewidth}
                \centering
                \includegraphics[width=1\linewidth]{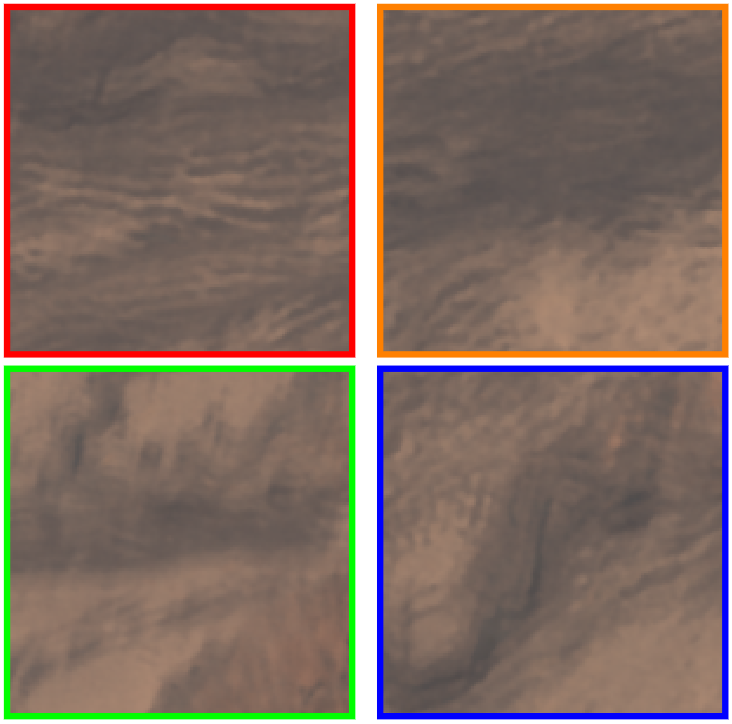}
            \end{minipage}%
        }
        \subfigure[\SpSNeRF~Lzh-A]{
            \begin{minipage}[t]{0.205\linewidth}
                \centering
                    \begin{tikzpicture}
                    \node[anchor=south west,inner sep=0] (image) at (0,0) {
                \includegraphics[width=1\linewidth]{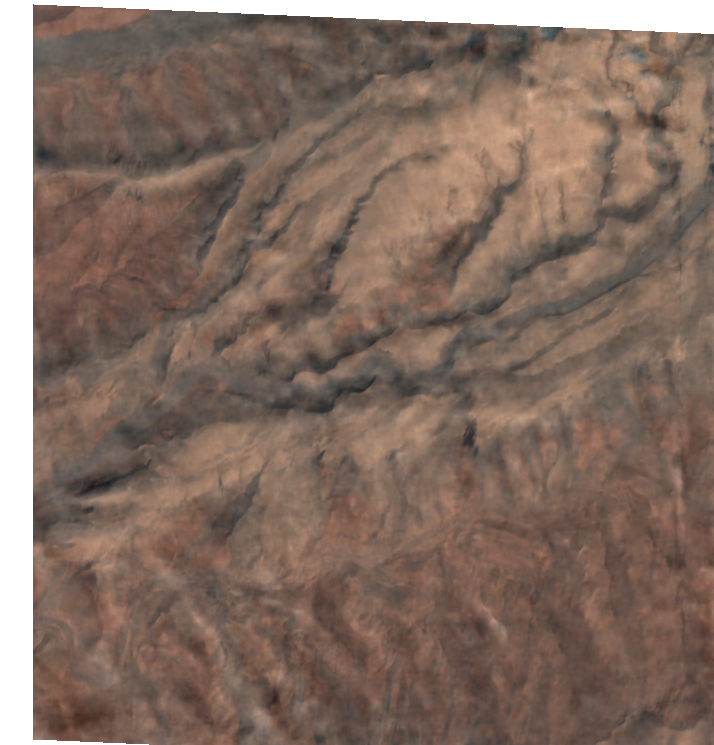}};
                    \begin{scope}[x={(image.south east)},y={(image.north west)}]
                    \draw[red,thick] (0.296,0.442) rectangle (0.504,0.656);
                    \draw[orange,thick] (0.575,0.564) rectangle (0.787,0.756);
                    \draw[green,thick] (0.090,0.380) rectangle (0.293,0.180);
                    \draw[blue,thick] (0.254,0.000) rectangle (0.465,0.198);
                    \end{scope}
                    \end{tikzpicture}
            \end{minipage}%
            \hspace{0mm}
            \begin{minipage}[t]{0.245\linewidth}
                \centering
                \includegraphics[width=1\linewidth]{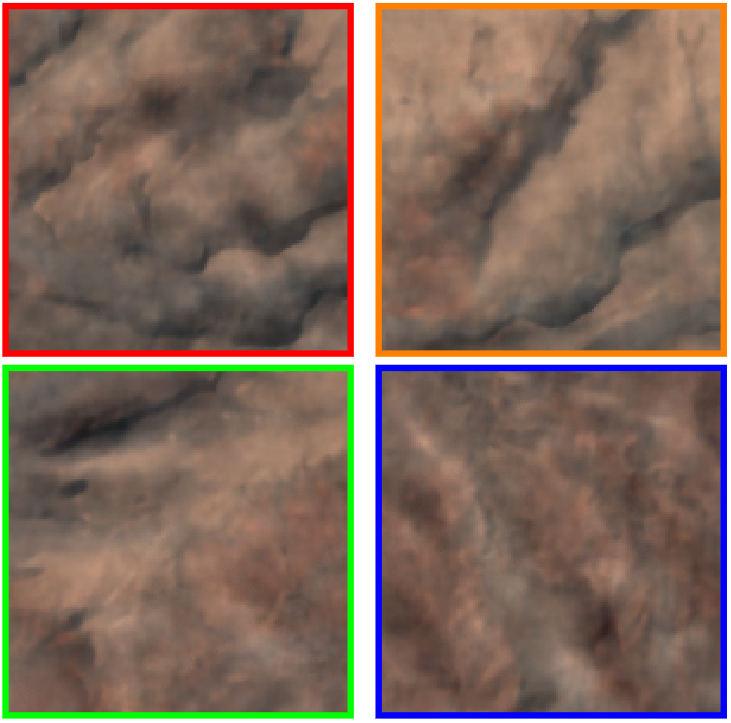}
            \end{minipage}%
        }
        \subfigure[\SpSNeRF~Lzh-B]{
            \begin{minipage}[t]{0.205\linewidth}
                \centering
                    \begin{tikzpicture}
                    \node[anchor=south west,inner sep=0] (image) at (0,0) {
                \includegraphics[width=1\linewidth]{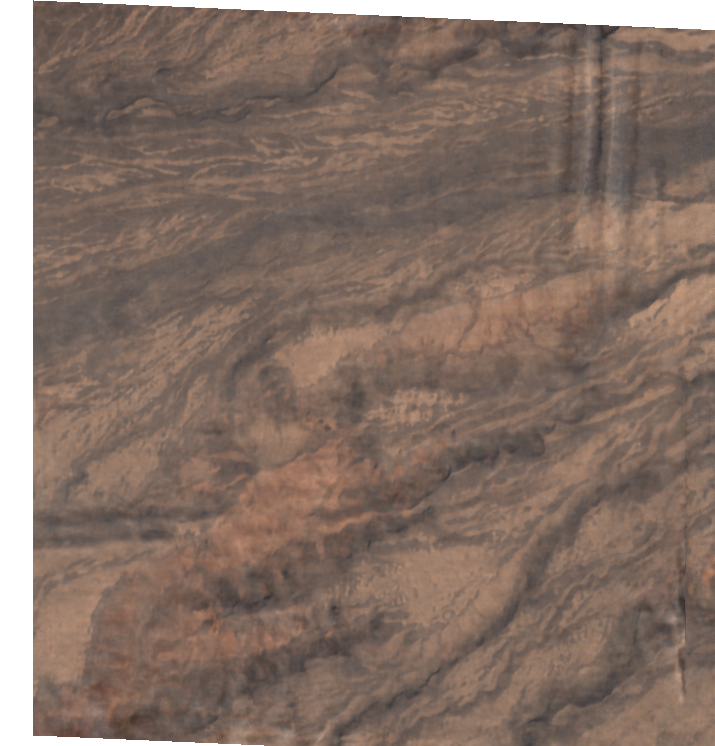}};
                    \begin{scope}[x={(image.south east)},y={(image.north west)}]
                    \draw[red,thick] (0.078,0.700) rectangle (0.307,0.924);
                    \draw[orange,thick] (0.742,0.693) rectangle (0.948,0.888);
                    \draw[green,thick] (0.053,0.196) rectangle (0.262,0.412);
                    \draw[blue,thick] (0.778,0.060) rectangle (0.99,0.258);
                    \end{scope}
                    \end{tikzpicture}
            \end{minipage}%
            \hspace{0mm}
            \begin{minipage}[t]{0.245\linewidth}
                \centering
                \includegraphics[width=1\linewidth]{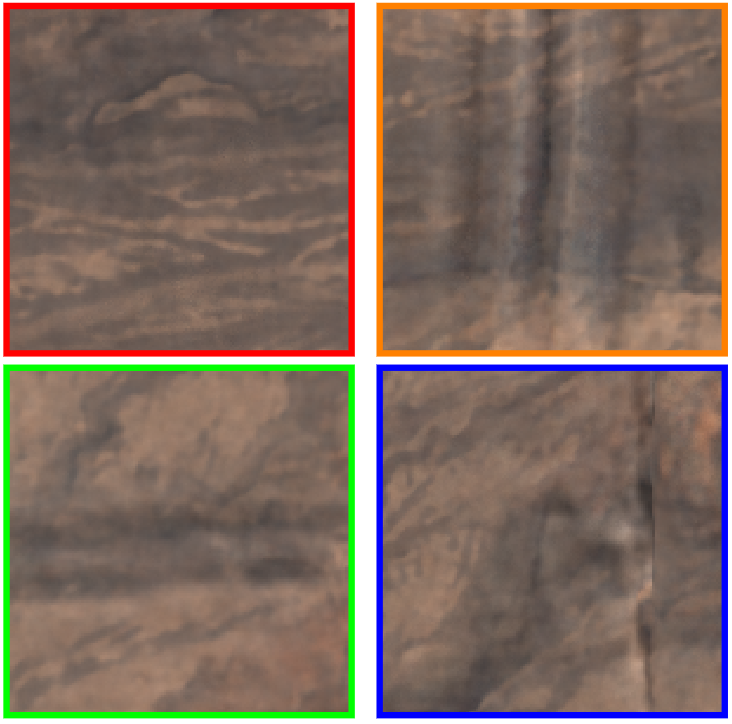}
            \end{minipage}%
        }        
        \subfigure[\OurNeRFShort~Lzh-A]{
            \begin{minipage}[t]{0.205\linewidth}
                \centering
                    \begin{tikzpicture}
                    \node[anchor=south west,inner sep=0] (image) at (0,0) {
                \includegraphics[width=1\linewidth]{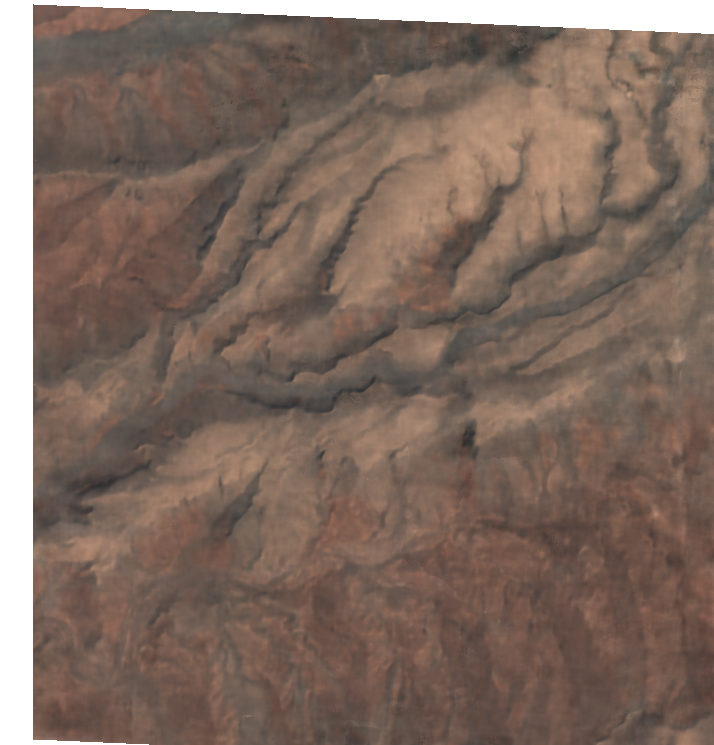}};
                    \begin{scope}[x={(image.south east)},y={(image.north west)}]
                    \draw[red,thick] (0.296,0.442) rectangle (0.504,0.656);
                    \draw[orange,thick] (0.575,0.564) rectangle (0.787,0.756);
                    \draw[green,thick] (0.090,0.380) rectangle (0.293,0.180);
                    \draw[blue,thick] (0.254,0.000) rectangle (0.465,0.198);
                    \end{scope}
                    \end{tikzpicture}
            \end{minipage}%
            \hspace{0mm}
            \begin{minipage}[t]{0.245\linewidth}
                \centering
                \includegraphics[width=1\linewidth]{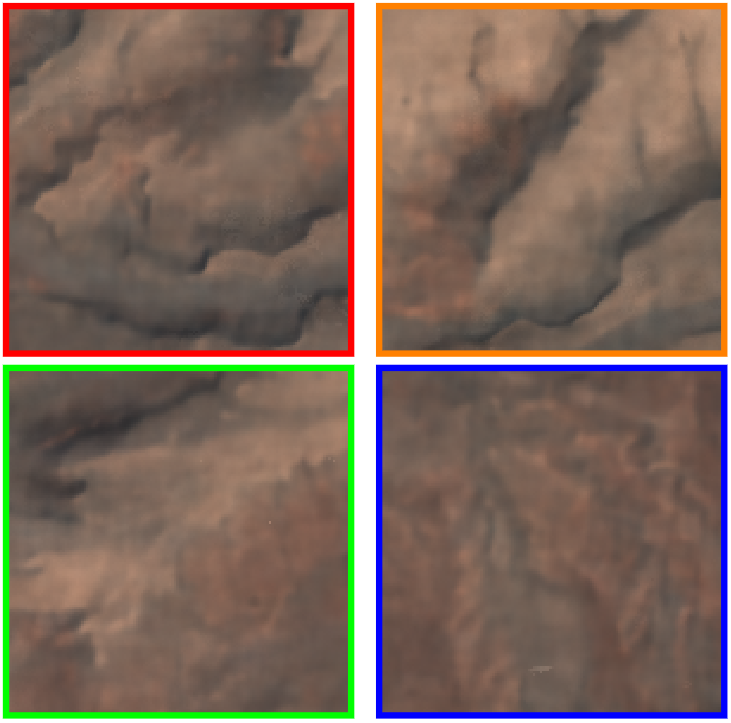}
            \end{minipage}%
        }
        \subfigure[\OurNeRFShort~Lzh-B]{
            \begin{minipage}[t]{0.205\linewidth}
                \centering
                    \begin{tikzpicture}
                    \node[anchor=south west,inner sep=0] (image) at (0,0) {
                \includegraphics[width=1\linewidth]{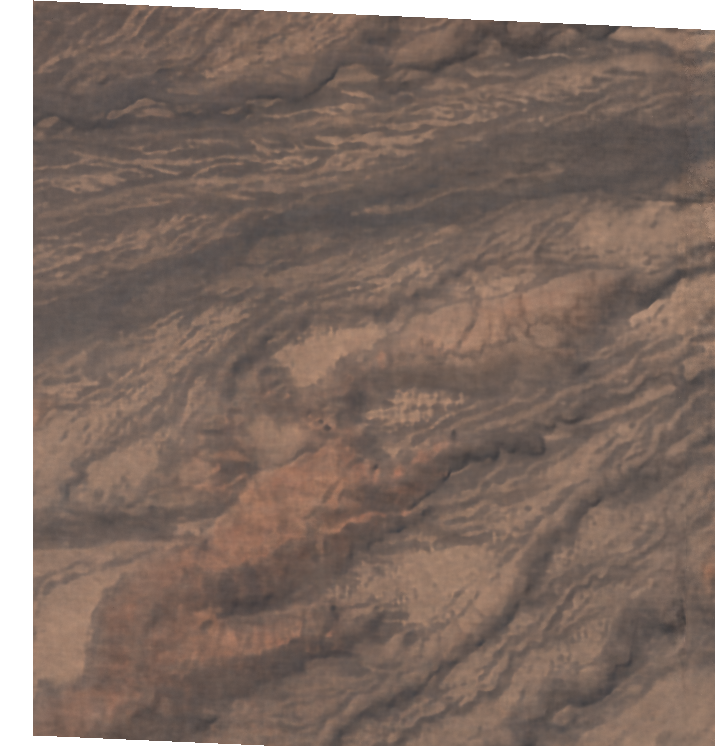}};
                    \begin{scope}[x={(image.south east)},y={(image.north west)}]
                    \draw[red,thick] (0.078,0.700) rectangle (0.307,0.924);
                    \draw[orange,thick] (0.742,0.693) rectangle (0.948,0.888);
                    \draw[green,thick] (0.053,0.196) rectangle (0.262,0.412);
                    \draw[blue,thick] (0.778,0.060) rectangle (0.99,0.258);
                    \end{scope}
                    \end{tikzpicture}
            \end{minipage}%
            \hspace{0mm}
            \begin{minipage}[t]{0.245\linewidth}
                \centering
                \includegraphics[width=1\linewidth]{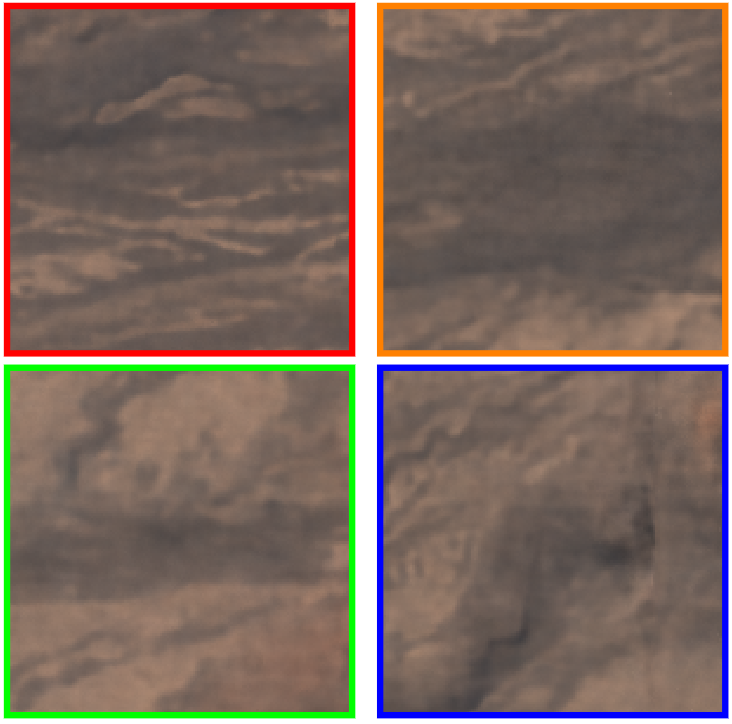}
            \end{minipage}%
        }
        \subfigure[GT Lzh-A]{
            \begin{minipage}[t]{0.205\linewidth}
                \centering
                    \begin{tikzpicture}
                    \node[anchor=south west,inner sep=0] (image) at (0,0) {
                \includegraphics[width=1\linewidth]{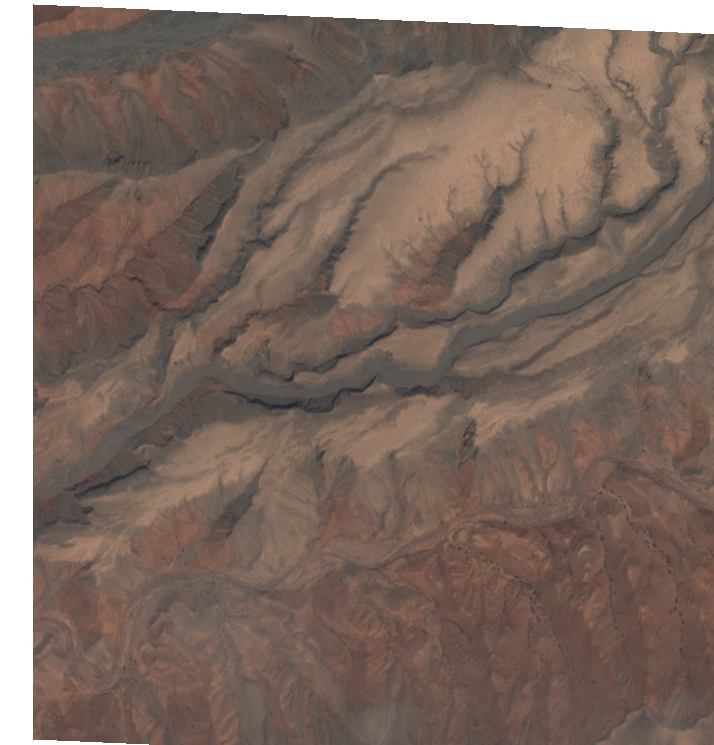}};
                    \begin{scope}[x={(image.south east)},y={(image.north west)}]
                    \draw[red,thick] (0.296,0.442) rectangle (0.504,0.656);
                    \draw[orange,thick] (0.575,0.564) rectangle (0.787,0.756);
                    \draw[green,thick] (0.090,0.380) rectangle (0.293,0.180);
                    \draw[blue,thick] (0.254,0.000) rectangle (0.465,0.198);
                    \end{scope}
                    \end{tikzpicture}
            \end{minipage}%
            \hspace{0mm}
            \begin{minipage}[t]{0.245\linewidth}
                \centering
                \includegraphics[width=1\linewidth]{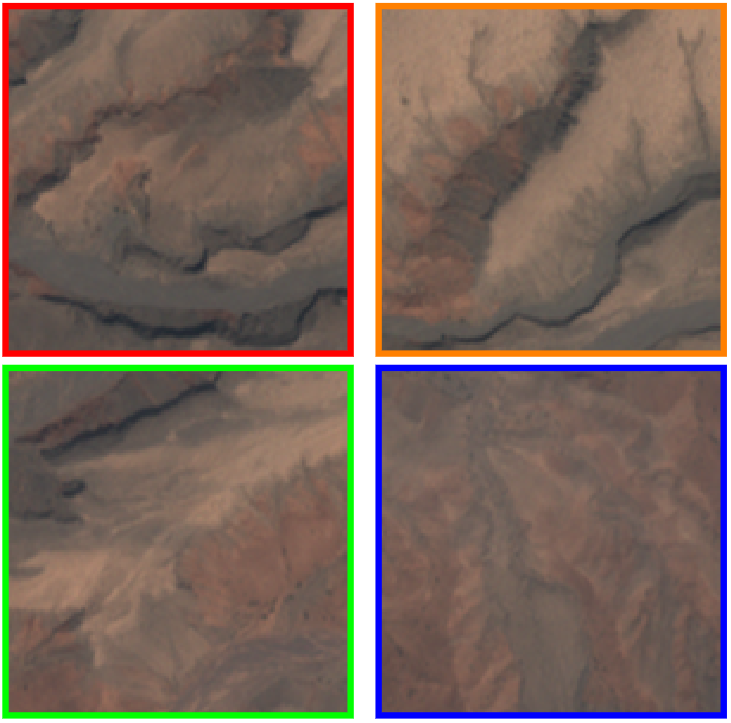}
            \end{minipage}%
        }
        \subfigure[GT Lzh-B]{
            \begin{minipage}[t]{0.205\linewidth}
                \centering
                    \begin{tikzpicture}
                    \node[anchor=south west,inner sep=0] (image) at (0,0) {
                \includegraphics[width=1\linewidth]{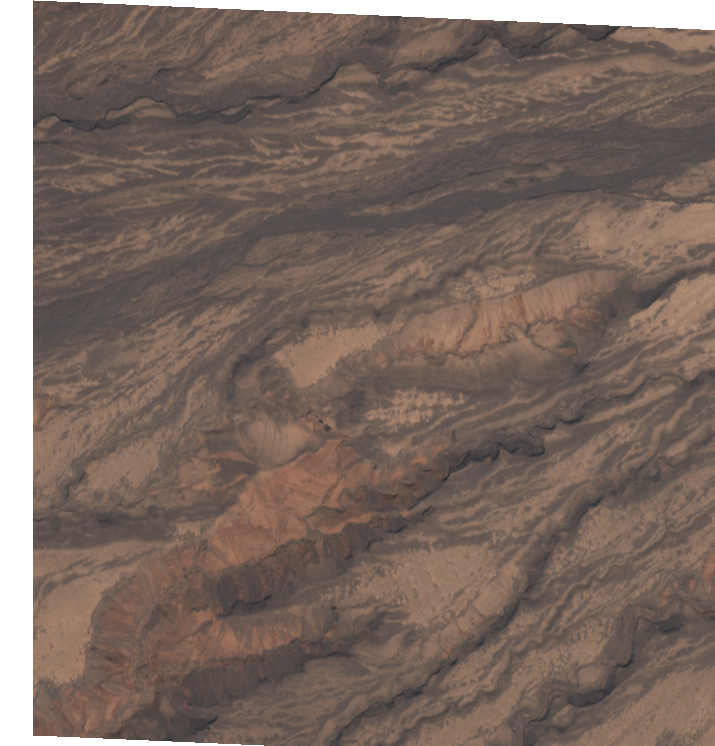}};
                    \begin{scope}[x={(image.south east)},y={(image.north west)}]
                    \draw[red,thick] (0.078,0.700) rectangle (0.307,0.924);
                    \draw[orange,thick] (0.742,0.693) rectangle (0.948,0.888);
                    \draw[green,thick] (0.053,0.196) rectangle (0.262,0.412);
                    \draw[blue,thick] (0.778,0.060) rectangle (0.99,0.258);
                    \end{scope}
                    \end{tikzpicture}
            \end{minipage}%
            \hspace{0mm}
            \begin{minipage}[t]{0.245\linewidth}
                \centering
                \includegraphics[width=1\linewidth]{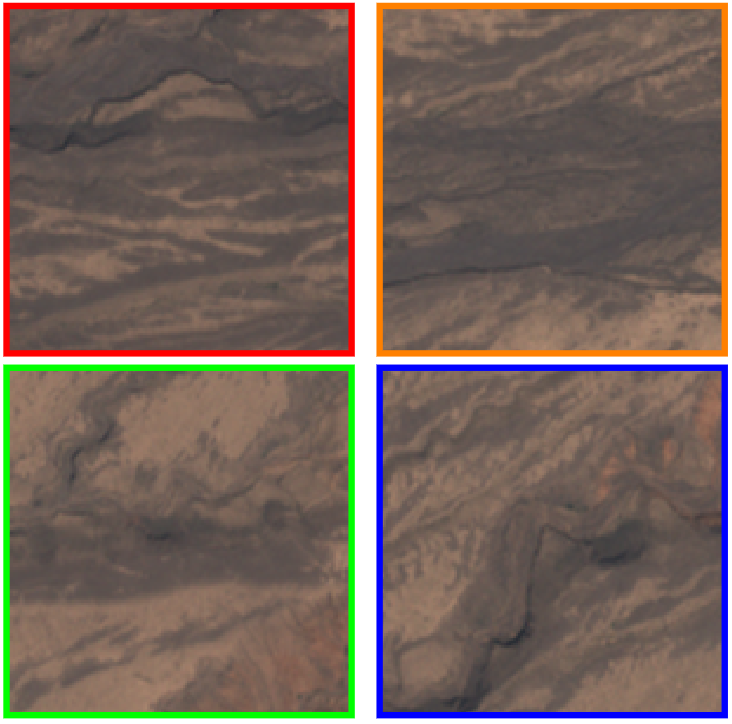}
            \end{minipage}%
        }
        \caption{\textbf{Novel View Synthesis -- Lanzhou Dataset.} Renderings of sites A and B of the multi-date Lanzhou dataset in the \textit{hard} scenario. Hallucination artefacts are revealed in both \SatNeRF~and \SpSNeRF~but are more pronounced in the latter. Despite this, \SpSNeRF~renders sharper surface details than \SatNeRF. \OurNeRFShort~generates images with fine details and much less artifacts.
        }
        \label{maintestnovalLzh}
    \end{center}
\end{figure*}

Quantitative metrics are presented in \Cref{maintestquantitativemetrics} while qualitative visualisations are provided in \Cref{maintestnovalDji,maintestnovalLzh}. The visualisations here are limited to the most challenging scenario (\textit{very hard} for Djibouti and \textit{hard} for Lanzhou) due to space limitation. Our \OurNeRFShort~outperforms \SatNeRF~and \SpSNeRF. Among competitive methods, \SpSNeRF~produces less blurry renderings  than \SatNeRF~(compare \Cref{maintestnovalDji} (a) and (c) as well as \Cref{maintestnovalLzh} (b) and (d)). Nevertheless, both methods produce minimal hallucination effects (\Cref{maintestnovalDji} (a), \Cref{maintestnovalLzh} (b,d)). The two competitive methods remain less sharp and far from the colour tone of the \NERF~which includes our realistic RPV-based BRDF (compare \Cref{maintestnovalDji} (c) and (e)). PSNR and SSIM metrics are best for \OurNeRFShort, followed by \SpSNeRF~in second place.
{The margins between \OurNeRFShort~and competitive approaches increase from \textit{easy} to \textit{very hard} mode, indicating greater robustness of \OurNeRFShort. From single to multiple epoch datasets (i.e., from Djibouti to Lanzhou), the image quality rendered by \SatNeRF~and \SpSNeRF~decreased significantly, while \OurNeRFShort~ recovered photorealistic images in both cases.}



\newcommand{\gmainpsnr}[1]{\gradientcelld{#1}{24}{29}{34}{gLow}{gMid}{gHigh}{70}}
\newcommand{\gmainssim}[1]{\gradientcelld{#1}{0.8}{0.89}{0.98}{gLow}{gMid}{gHigh}{70}}
\newcommand{\gmaingeom}[1]{\gradientcelld{#1}{1.0}{3.0}{5.0}{gHigh}{gMid}{gLow}{70}}

\begin{table}[htbp!]
\centering
\begin{tabular}{|c|c|c|c|c|c|}
\hline
{Metric} & {Method} & {Dji-A} & {Dji-B} & {Lzh-A} & {Lzh-B} \\ \hline
PSNR&{\SatNeRF~}&\gmainpsnr{32.747}&\gmainpsnr{31.818}&\textcolor{magenta}{\gmainpsnr{29.979}}&\gmainpsnr{25.470}\\\cline{2-6}
(Easy)&{\SpSNeRF~}&\textcolor{magenta}{\gmainpsnr{38.832}}&\textcolor{magenta}{\gmainpsnr{38.174}}&\gmainpsnr{29.548}&\textcolor{magenta}{\gmainpsnr{30.904}}\\\cline{2-6}
$\uparrow$&{\OurNeRFShort~}&{\textcolor{blue}{\gmainpsnr{41.844}}}&{\textcolor{blue}{\gmainpsnr{40.823}}}&{\textcolor{blue}{\gmainpsnr{32.196}}}&{\textcolor{blue}{\gmainpsnr{32.165}}}\\\hline\hline
PSNR&\SatNeRF~&\gmainpsnr{25.542}&\gmainpsnr{23.699}&\textcolor{magenta}{\gmainpsnr{25.963}}&\gmainpsnr{20.811}\\\cline{2-6}
(Hard)&\SpSNeRF~&\textcolor{magenta}{\gmainpsnr{28.348}}&\textcolor{magenta}{\gmainpsnr{27.468}}&\gmainpsnr{24.755}&\textcolor{magenta}{\gmainpsnr{24.148}}\\\cline{2-6}
$\uparrow$&\OurNeRFShort~&\textcolor{blue}{\gmainpsnr{36.232}}&\textcolor{blue}{\gmainpsnr{35.448}}&\textcolor{blue}{\gmainpsnr{28.420}}&\textcolor{blue}{\gmainpsnr{27.814}}\\\hline\hline
PSNR&\SatNeRF~&\textcolor{magenta}{\gmainpsnr{23.581}}&\gmainpsnr{21.288}&/&/\\\cline{2-6}
(VHard)&\SpSNeRF~&\gmainpsnr{23.144}&\textcolor{magenta}{\gmainpsnr{22.31}}&/&/\\\cline{2-6}
$\uparrow$&\OurNeRFShort~&\textcolor{blue}{\gmainpsnr{33.35}}&\textcolor{blue}{\gmainpsnr{32.376}}&/&/\\\hline\hline

SSIM&\SatNeRF~&\gmainssim{0.927}&\gmainssim{0.950}&\gmainssim{0.962}&\gmainssim{0.925}\\\cline{2-6}
(Easy)&\SpSNeRF~&\textcolor{magenta}{\gmainssim{0.975}}&\textcolor{magenta}{\gmainssim{0.979}}&\textcolor{magenta}{\gmainssim{0.965}}&\textcolor{magenta}{\gmainssim{0.970}}\\\cline{2-6}
$\uparrow$&\OurNeRFShort~&\textcolor{blue}{\gmainssim{0.985}}&\textcolor{blue}{\gmainssim{0.988}}&\textcolor{blue}{\gmainssim{0.979}}&\textcolor{blue}{\gmainssim{0.975}}\\\hline\hline
SSIM&\SatNeRF~&\gmainssim{0.766}&\gmainssim{0.825}&\textcolor{magenta}{\gmainssim{0.909}}&\gmainssim{0.760}\\\cline{2-6}
(Hard)&\SpSNeRF~&\textcolor{magenta}{\gmainssim{0.840}}&\textcolor{magenta}{\gmainssim{0.887}}&\gmainssim{0.900}&\textcolor{magenta}{\gmainssim{0.928}}\\\cline{2-6}
$\uparrow$&\OurNeRFShort~&\textcolor{blue}{\gmainssim{0.957}}&\textcolor{blue}{\gmainssim{0.965}}&\textcolor{blue}{\gmainssim{0.94}}&\textcolor{blue}{\gmainssim{0.953}}\\\hline\hline
SSIM&\SatNeRF~&\textcolor{magenta}{\gmainssim{0.676}}&\textcolor{magenta}{\gmainssim{0.768}}&/&/\\\cline{2-6}
(VHard)&\SpSNeRF~&\gmainssim{0.614}&\gmainssim{0.72}&/&/\\\cline{2-6}
$\uparrow$&\OurNeRFShort~&\textcolor{blue}{\gmainssim{0.918}}&\textcolor{blue}{\gmainssim{0.942}}&/&/\\\hline\hline

 & \SatNeRF~& \gmaingeom{12.85} & \gmaingeom{18.059} & \gmaingeom{61.299} & \gmaingeom{27.489} \\ \cline{2-6}
MAE & \SpSNeRF~& \gmaingeom{1.438} & \gmaingeom{1.761} & \gmaingeom{3.558} & \gmaingeom{3.235} \\ \cline{2-6}
$\downarrow$ & \OurNeRFShort~& \textcolor{magenta}{\gmaingeom{1.378}} & \textcolor{magenta}{\gmaingeom{1.614}} & \textcolor{magenta}{\gmaingeom{3.42}} & \textcolor{magenta}{\gmaingeom{2.941}} \\ \cline{2-6}
& SGM$_{Z1}$ & \textcolor{blue}{\gmaingeom{1.061}} & \textcolor{blue}{\gmaingeom{1.052}} & \textcolor{blue}{\gmaingeom{1.409}} & \textcolor{blue}{\gmaingeom{1.220}} \\ \hline
\end{tabular}
\caption{\textbf{Quantitative Evaluation}. The best and second best performing metrics are in \textcolor{blue}{blue} and \textcolor{magenta}{magenta}. {For each dataset, we train a \OurNeRFShort~to render three images in easy, hard (and very hard when existing) modes at the same time.} \OurNeRFShort~achieves better PSNR, SSIM and MAE than \SatNeRF~and \SpSNeRF. However, \OurNeRFShort~has higher MAEs than SGM$_{Z1}$, which we attribute {to \NERF's design that handles pixels individually without taking context into account.}}
\label{maintestquantitativemetrics}
\end{table}

\subsection{Altitude estimation}
\label{Altitudeestimation}
\begin{figure*}[!htbp]
    \begin{center}
        \subfigure[\SatNeRF~Dji-A]{
            \begin{minipage}[t]{0.223\linewidth}
                \centering
                    \begin{tikzpicture}
                    \node[anchor=south west,inner sep=0] (image) at (0,0) {
                \includegraphics[width=1\linewidth]{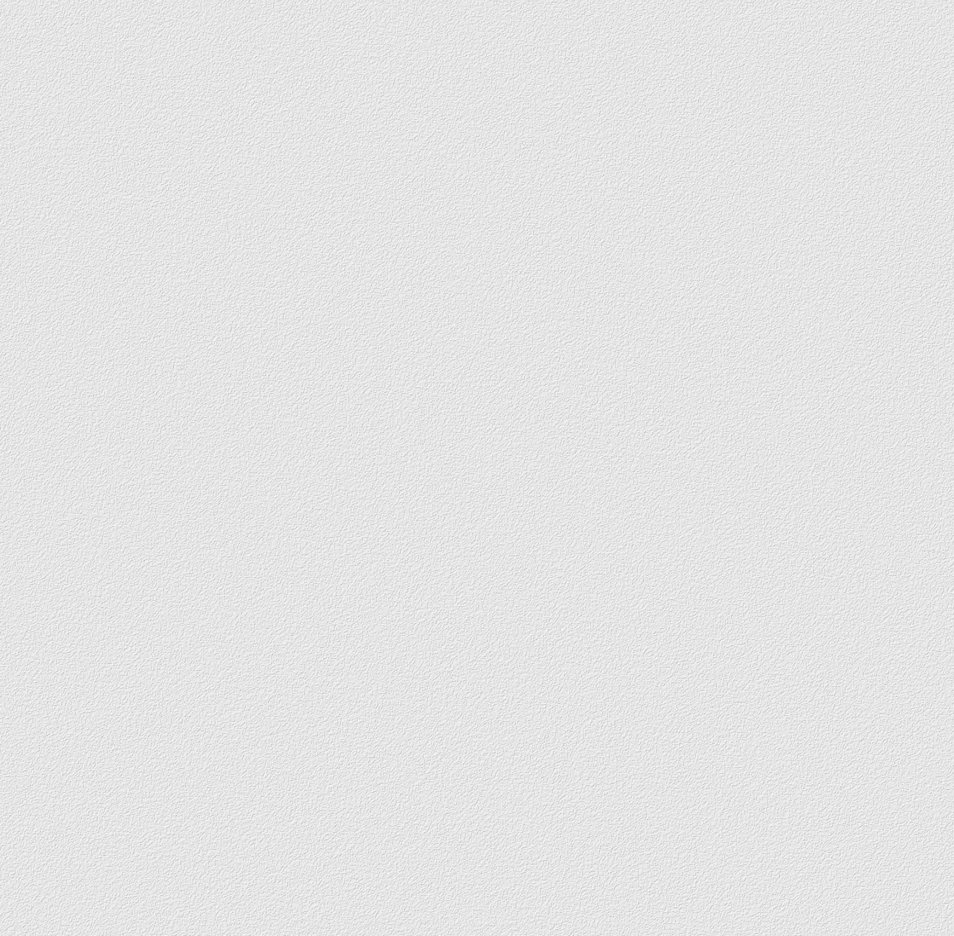}};
                    \begin{scope}[x={(image.south east)},y={(image.north west)}]
                    \draw[green,thick] (0.023,0.154) rectangle (0.186,0.323);
                    \draw[blue,thick] (0.840,0.140) rectangle (0.999,0.306);
                    \draw[orange,thick] (0.830,0.491) rectangle (0.992,0.655);
                    \draw[red,thick] (0.645,0.795) rectangle (0.802,0.628);
                    \end{scope}
                    \end{tikzpicture}
            \end{minipage}%
            \hspace{0mm}
            \begin{minipage}[t]{0.227\linewidth}
                \centering
                \includegraphics[width=1\linewidth]{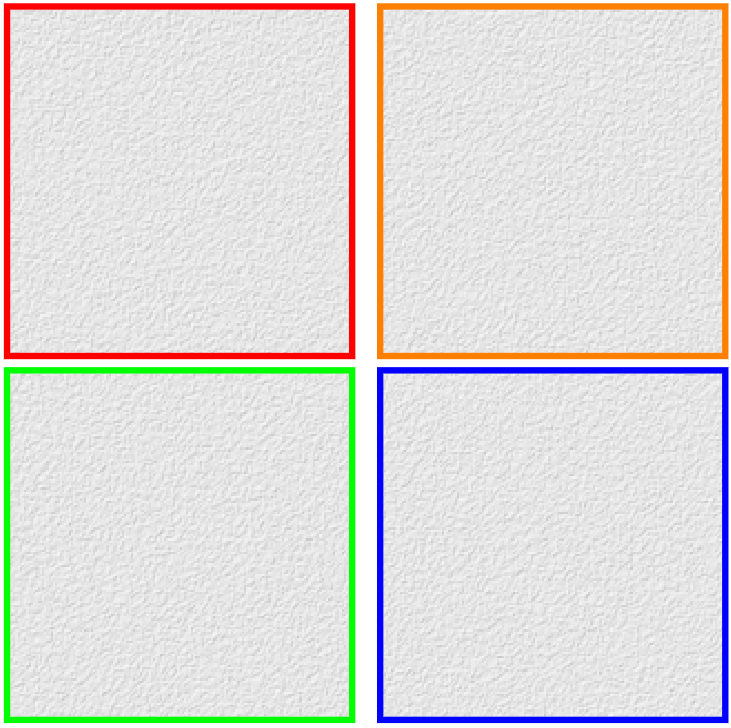}
            \end{minipage}%
            }
        \subfigure[\SatNeRF~Dji-B]{
            \begin{minipage}[t]{0.223\linewidth}
                \centering
                    \begin{tikzpicture}
                    \node[anchor=south west,inner sep=0] (image) at (0,0) {
                \includegraphics[width=1\linewidth]{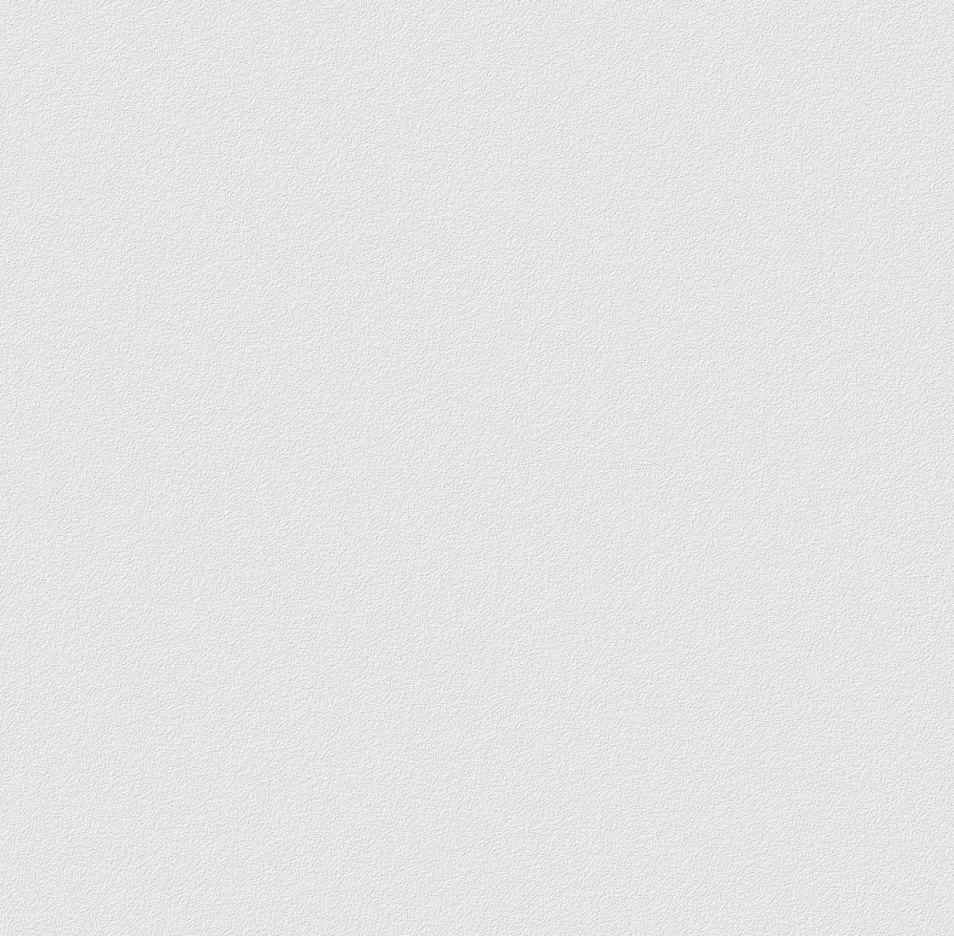}};
                    \begin{scope}[x={(image.south east)},y={(image.north west)}]
                    \draw[red,thick] (0.175,0.634) rectangle (0.345,0.460);
                    \draw[orange,thick] (0.595,0.816) rectangle (0.781,0.995);
                    \draw[green,thick] (0.420,0.271) rectangle (0.587,0.113);
                    \draw[blue,thick] (0.750,0.468) rectangle (0.919,0.310);
                    \end{scope}
                    \end{tikzpicture}
            \end{minipage}%
            \hspace{0mm}
            \begin{minipage}[t]{0.227\linewidth}
                \centering
                \includegraphics[width=1\linewidth]{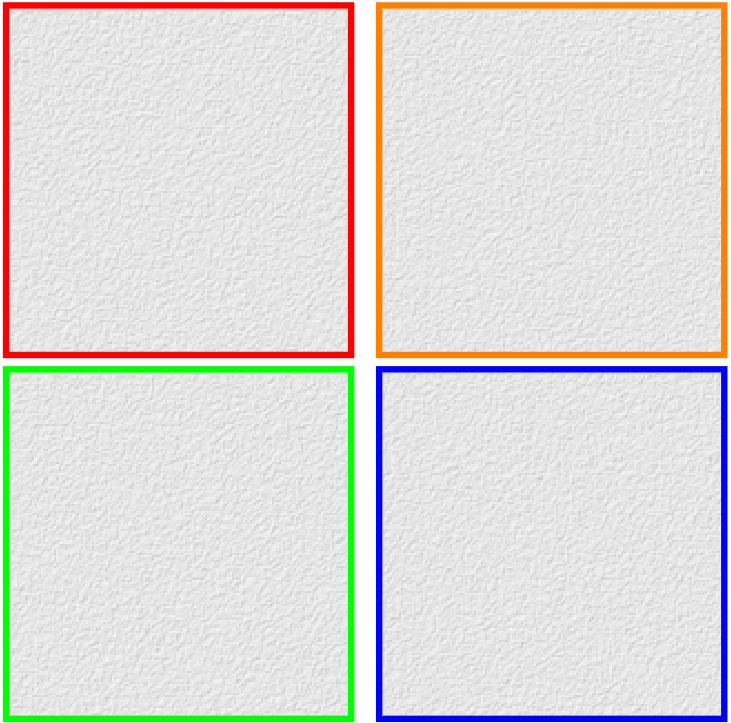}
            \end{minipage}%
            }
        \subfigure[\SpSNeRF~Dji-A]{
            \begin{minipage}[t]{0.223\linewidth}
                \centering
                    \begin{tikzpicture}
                    \node[anchor=south west,inner sep=0] (image) at (0,0) {
                \includegraphics[width=1\linewidth]{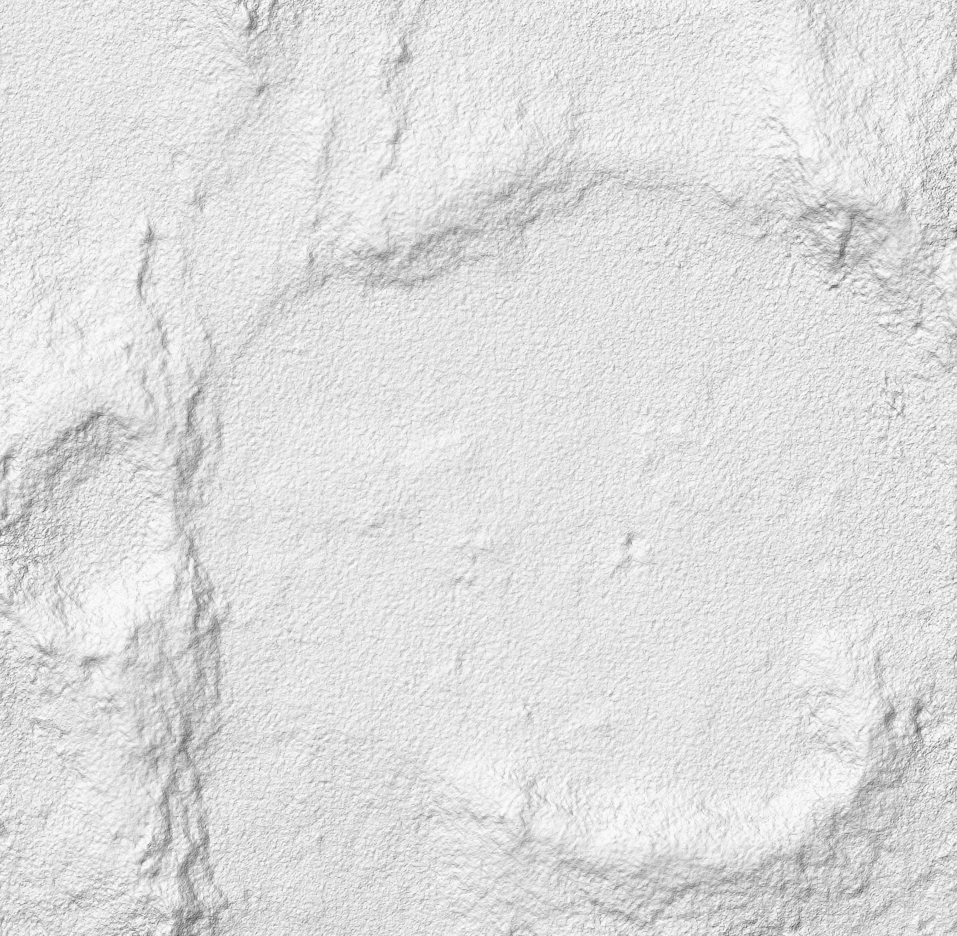}};
                    \begin{scope}[x={(image.south east)},y={(image.north west)}]
                    \draw[green,thick] (0.023,0.154) rectangle (0.186,0.323);
                    \draw[blue,thick] (0.840,0.140) rectangle (0.999,0.306);
                    \draw[orange,thick] (0.830,0.491) rectangle (0.992,0.655);
                    \draw[red,thick] (0.645,0.795) rectangle (0.802,0.628);
                    \end{scope}
                    \end{tikzpicture}
            \end{minipage}%
            \hspace{0mm}
            \begin{minipage}[t]{0.227\linewidth}
                \centering
                \includegraphics[width=1\linewidth]{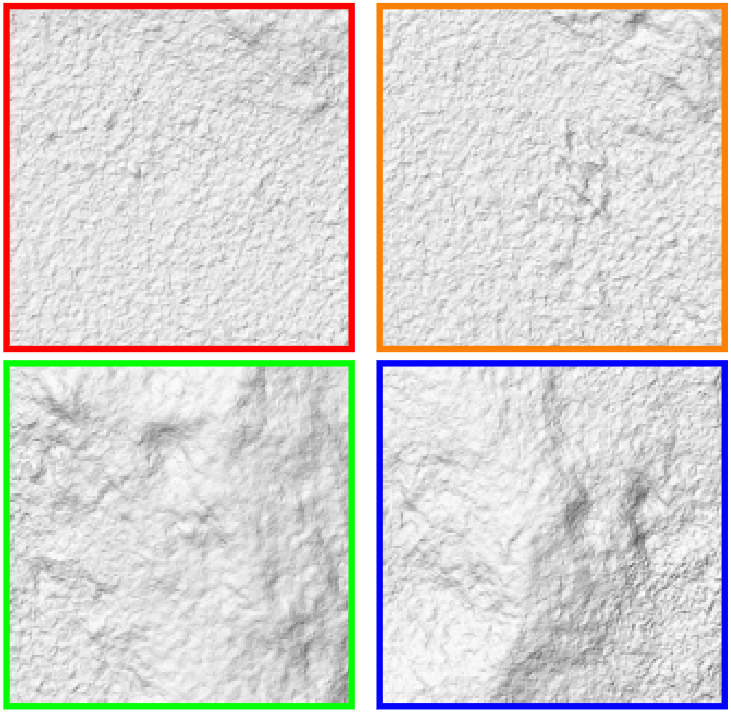}
            \end{minipage}%
            }
        \subfigure[\SpSNeRF~Dji-B]{
            \begin{minipage}[t]{0.223\linewidth}
                \centering
                    \begin{tikzpicture}
                    \node[anchor=south west,inner sep=0] (image) at (0,0) {
                \includegraphics[width=1\linewidth]{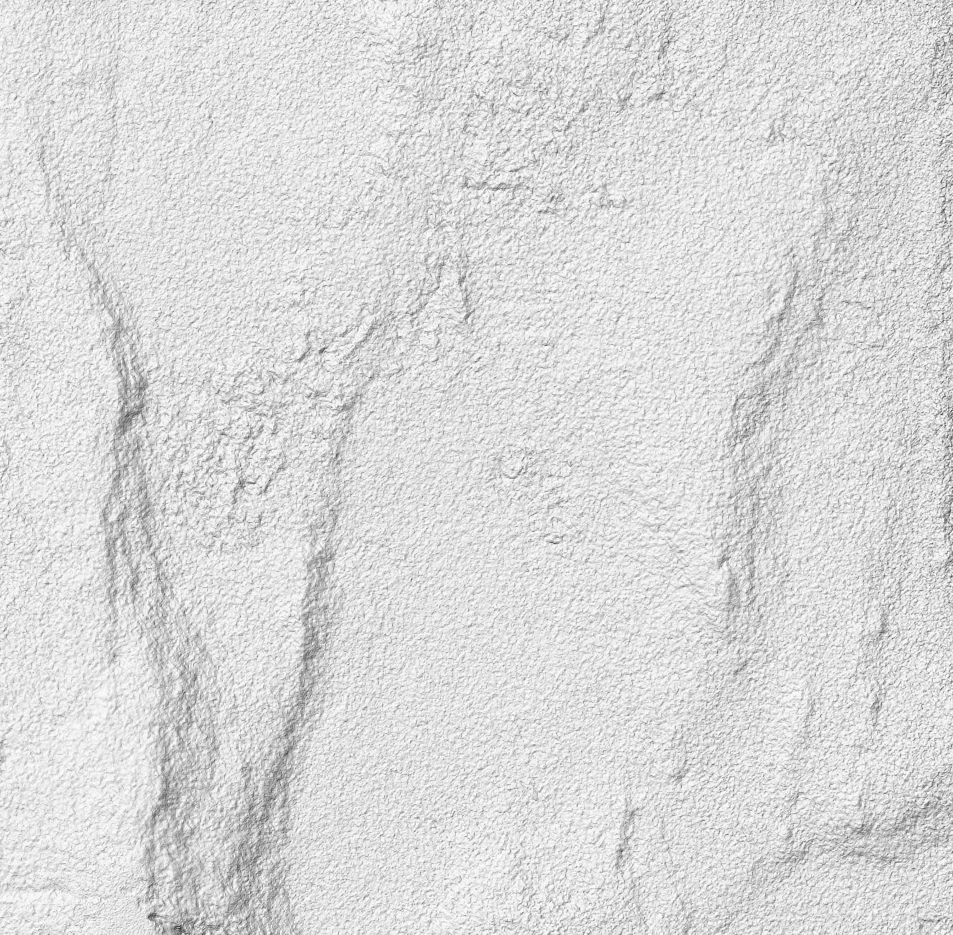}};
                    \begin{scope}[x={(image.south east)},y={(image.north west)}]
                    \draw[red,thick] (0.175,0.634) rectangle (0.345,0.460);
                    \draw[orange,thick] (0.595,0.816) rectangle (0.781,0.995);
                    \draw[green,thick] (0.420,0.271) rectangle (0.587,0.113);
                    \draw[blue,thick] (0.750,0.468) rectangle (0.919,0.310);
                    \end{scope}
                    \end{tikzpicture}
            \end{minipage}%
            \hspace{0mm}
            \begin{minipage}[t]{0.227\linewidth}
                \centering
                \includegraphics[width=1\linewidth]{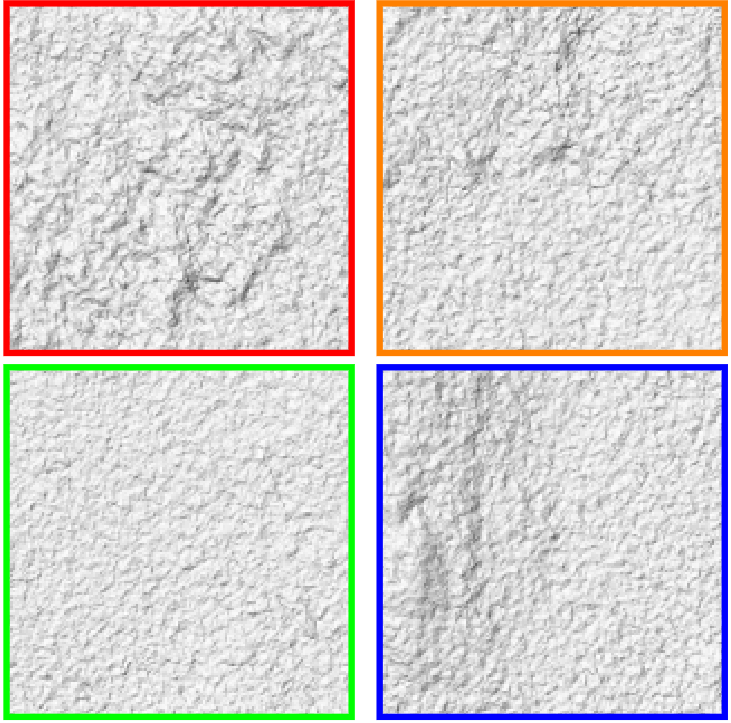}
            \end{minipage}%
            }
        \subfigure[\OurNeRFShort~Dji-A]{
            \begin{minipage}[t]{0.223\linewidth}
                \centering
                    \begin{tikzpicture}
                    \node[anchor=south west,inner sep=0] (image) at (0,0) {
                \includegraphics[width=1\linewidth]{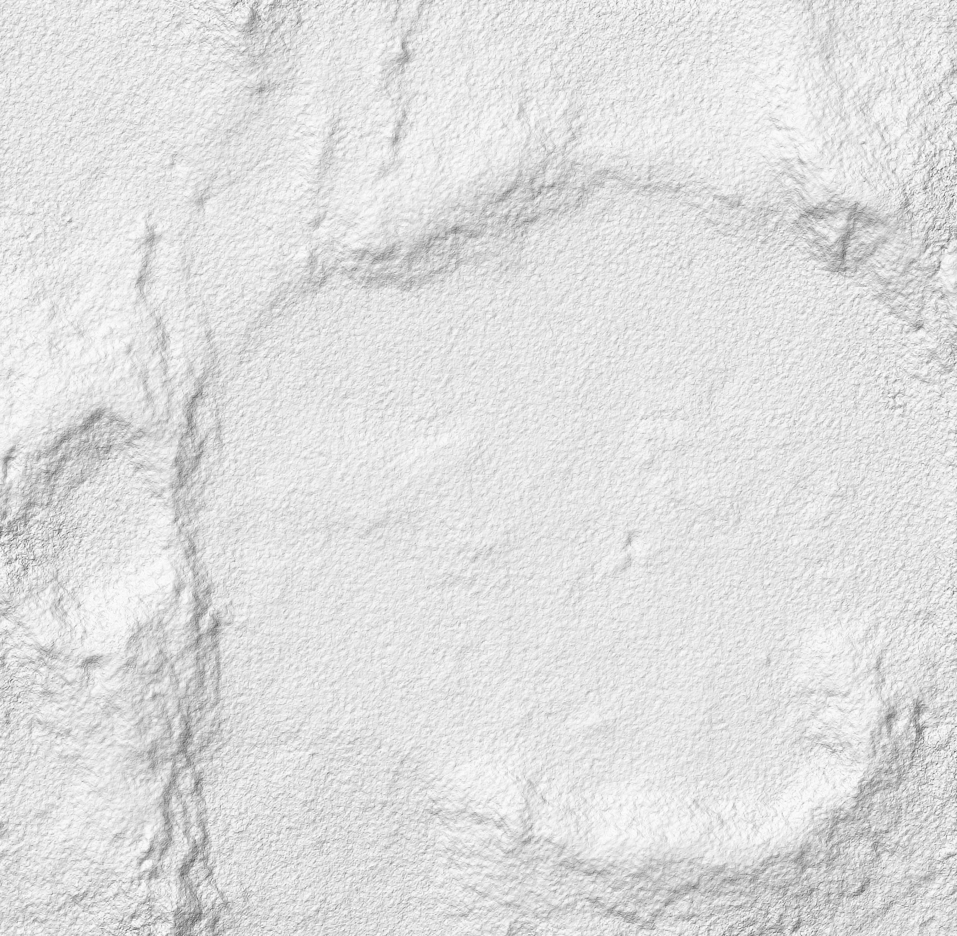}};
                    \begin{scope}[x={(image.south east)},y={(image.north west)}]
                    \draw[green,thick] (0.023,0.154) rectangle (0.186,0.323);
                    \draw[blue,thick] (0.840,0.140) rectangle (0.999,0.306);
                    \draw[orange,thick] (0.830,0.491) rectangle (0.992,0.655);
                    \draw[red,thick] (0.645,0.795) rectangle (0.802,0.628);
                    \end{scope}
                    \end{tikzpicture}
            \end{minipage}%
            \hspace{0mm}
            \begin{minipage}[t]{0.227\linewidth}
                \centering
                \includegraphics[width=1\linewidth]{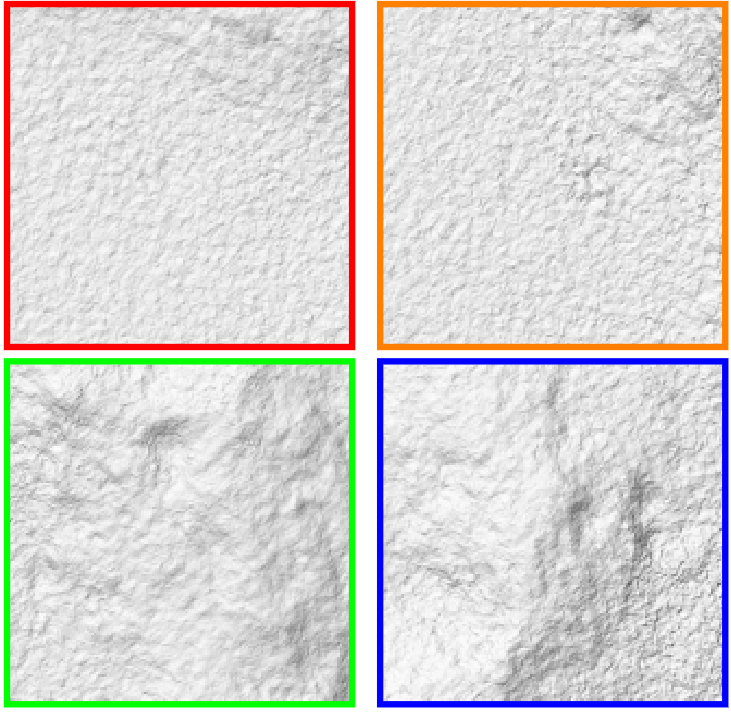}
            \end{minipage}%
            }
        \subfigure[\OurNeRFShort~Dji-B]{
            \begin{minipage}[t]{0.223\linewidth}
                \centering
                    \begin{tikzpicture}
                    \node[anchor=south west,inner sep=0] (image) at (0,0) {
                \includegraphics[width=1\linewidth]{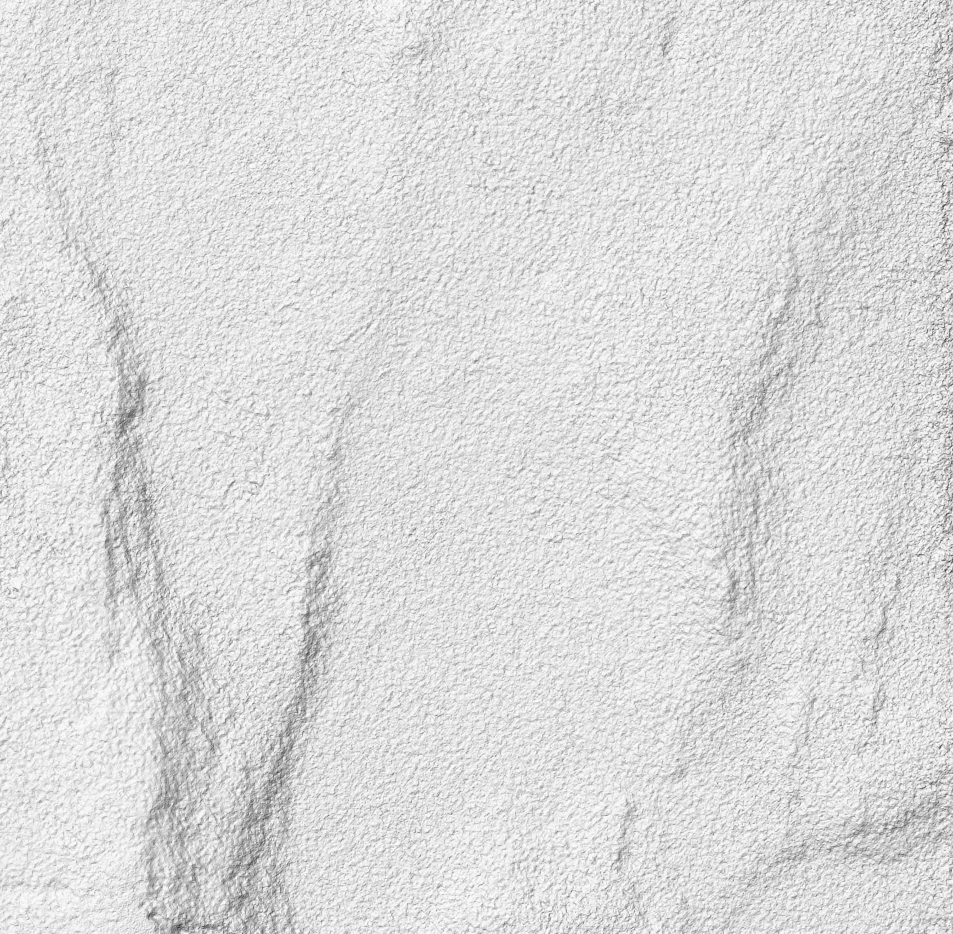}};
                    \begin{scope}[x={(image.south east)},y={(image.north west)}]
                    \draw[red,thick] (0.175,0.634) rectangle (0.345,0.460);
                    \draw[orange,thick] (0.595,0.816) rectangle (0.781,0.995);
                    \draw[green,thick] (0.420,0.271) rectangle (0.587,0.113);
                    \draw[blue,thick] (0.750,0.468) rectangle (0.919,0.310);
                    \end{scope}
                    \end{tikzpicture}
            \end{minipage}%
            \hspace{0mm}
            \begin{minipage}[t]{0.227\linewidth}
                \centering
                \includegraphics[width=1\linewidth]{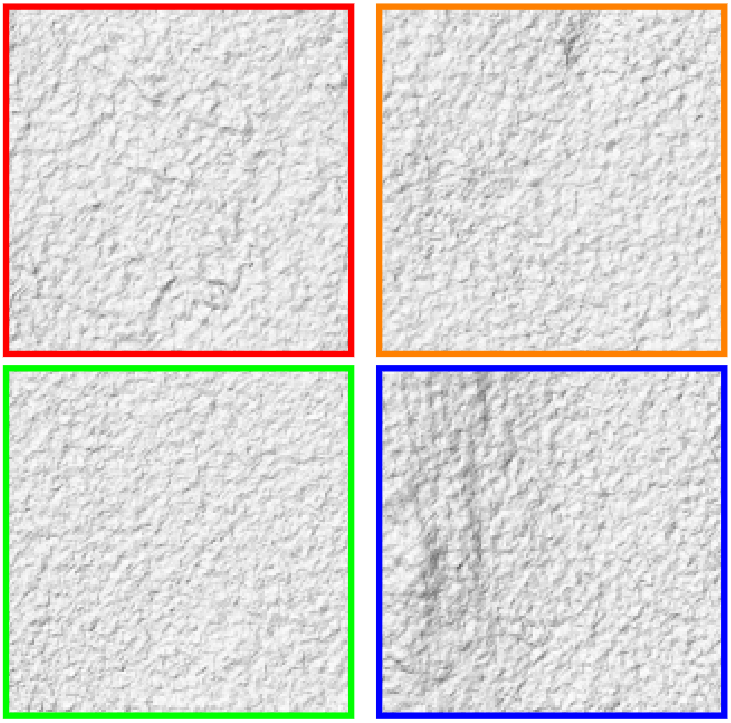}
            \end{minipage}%
            }
        \subfigure[SGM$_{Z1}$ Dji-A]{
            \begin{minipage}[t]{0.223\linewidth}
                \centering
                    \begin{tikzpicture}
                    \node[anchor=south west,inner sep=0] (image) at (0,0) {
                \includegraphics[width=1\linewidth]{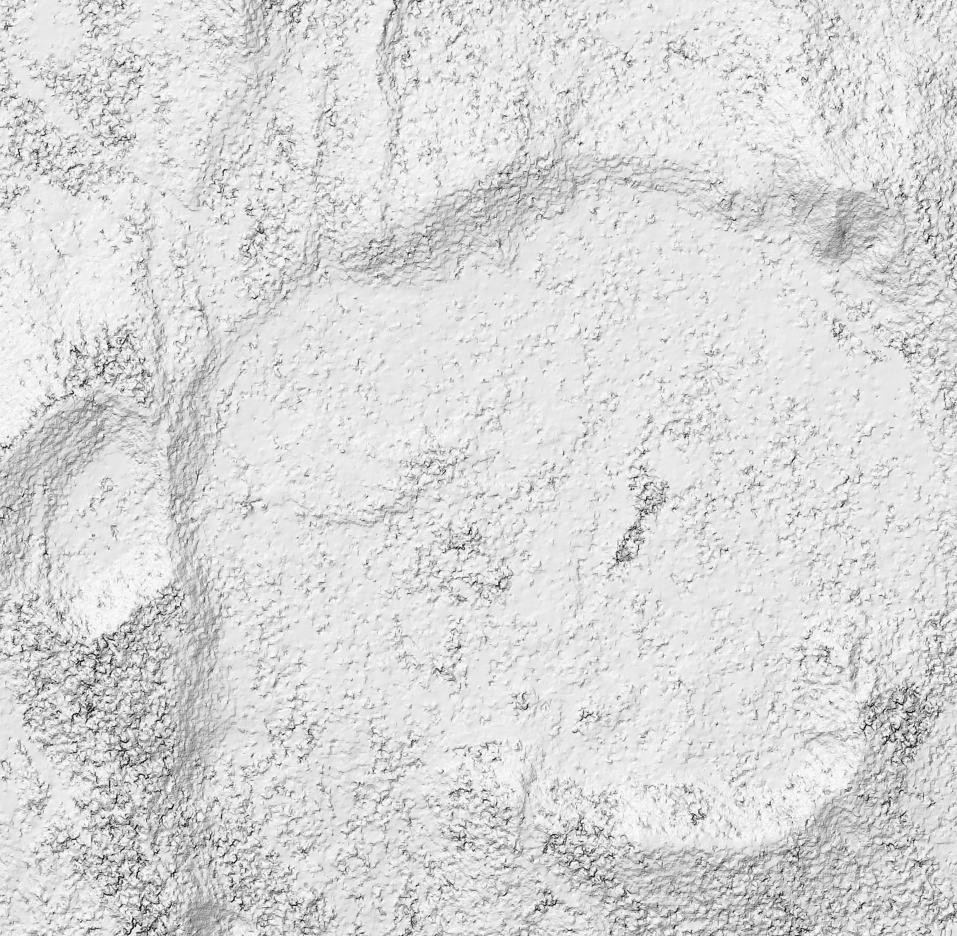}};
                    \begin{scope}[x={(image.south east)},y={(image.north west)}]
                    \draw[green,thick] (0.023,0.154) rectangle (0.186,0.323);
                    \draw[blue,thick] (0.840,0.140) rectangle (0.999,0.306);
                    \draw[orange,thick] (0.830,0.491) rectangle (0.992,0.655);
                    \draw[red,thick] (0.645,0.795) rectangle (0.802,0.628);
                    \end{scope}
                    \end{tikzpicture}
            \end{minipage}%
            \hspace{0mm}
            \begin{minipage}[t]{0.227\linewidth}
                \centering
                \includegraphics[width=1\linewidth]{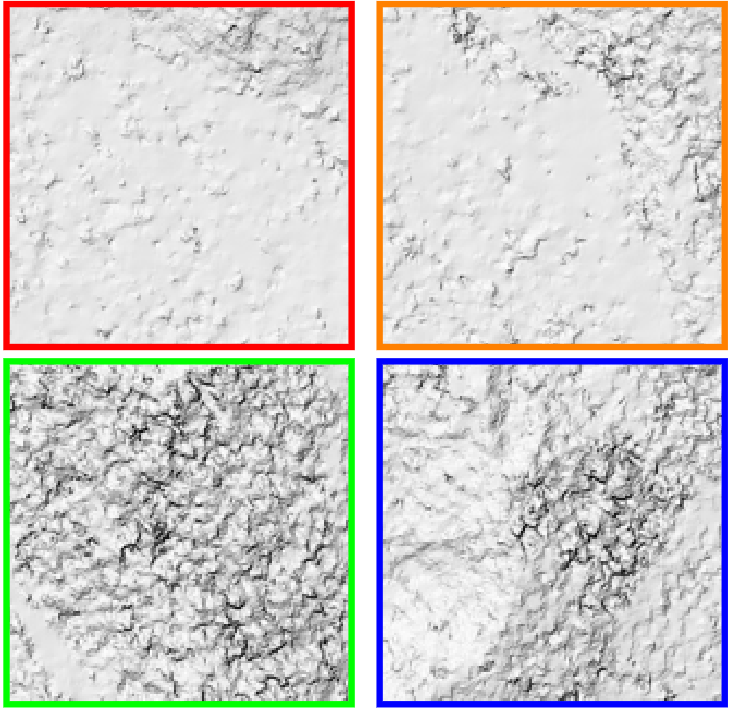}
            \end{minipage}%
            }
        \subfigure[SGM$_{Z1}$ Dji-B]{
            \begin{minipage}[t]{0.223\linewidth}
                \centering
                    \begin{tikzpicture}
                    \node[anchor=south west,inner sep=0] (image) at (0,0) {
                \includegraphics[width=1\linewidth]{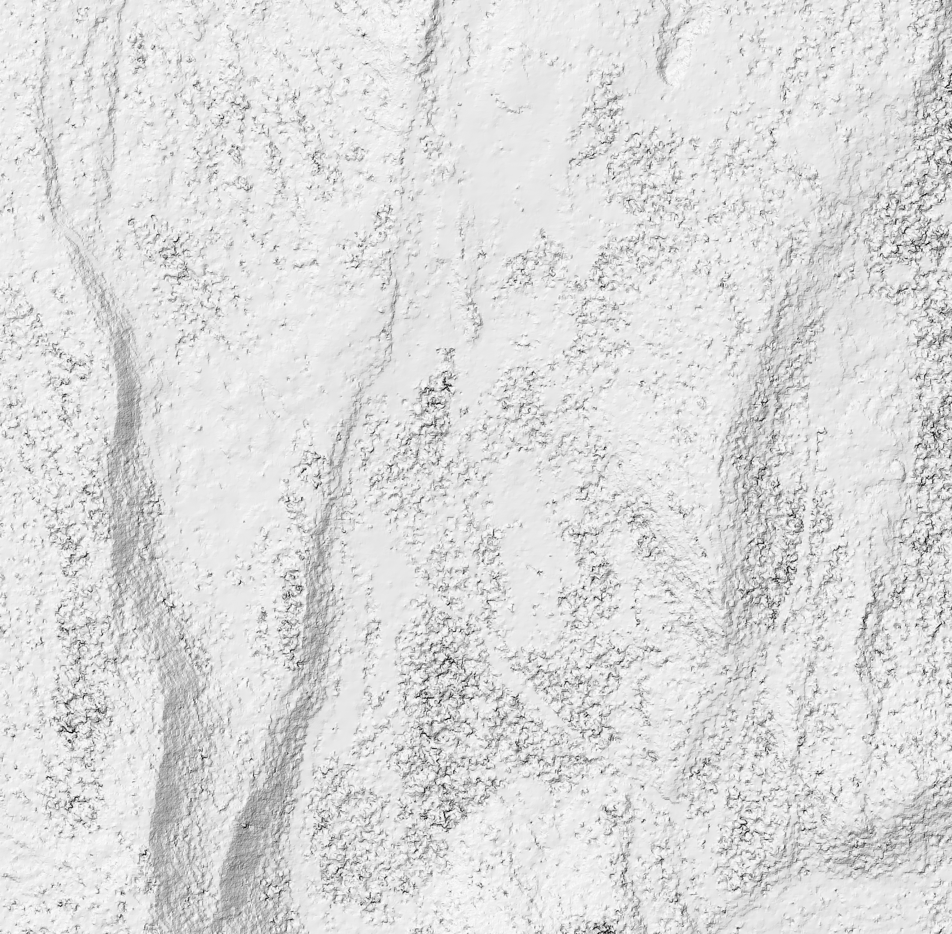}};
                    \begin{scope}[x={(image.south east)},y={(image.north west)}]
                    \draw[red,thick] (0.175,0.634) rectangle (0.345,0.460);
                    \draw[orange,thick] (0.595,0.816) rectangle (0.781,0.995);
                    \draw[green,thick] (0.420,0.271) rectangle (0.587,0.113);
                    \draw[blue,thick] (0.750,0.468) rectangle (0.919,0.310);
                    \end{scope}
                    \end{tikzpicture}
            \end{minipage}%
            \hspace{0mm}
            \begin{minipage}[t]{0.227\linewidth}
                \centering
                \includegraphics[width=1\linewidth]{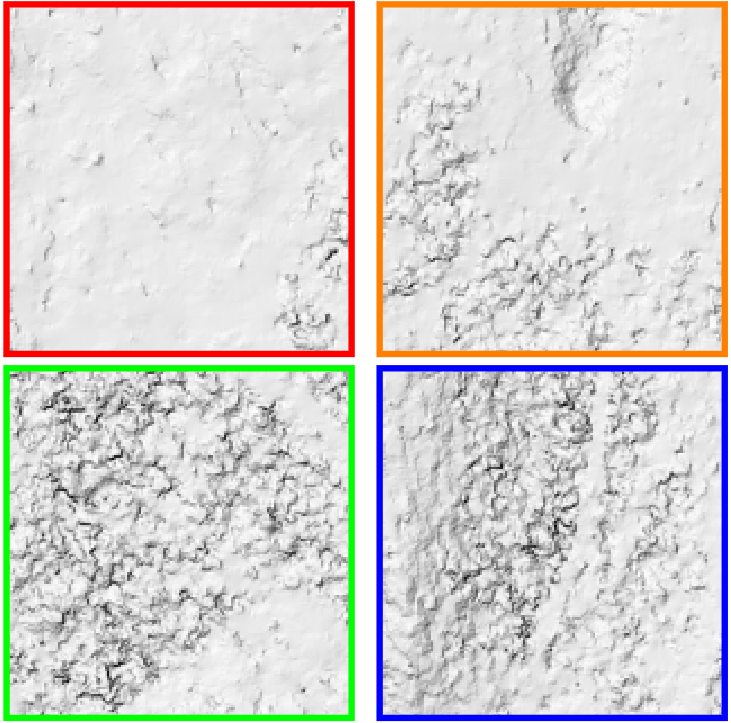}
            \end{minipage}%
            }    
        \subfigure[GT Dji-A]{
            \begin{minipage}[t]{0.223\linewidth}
                \centering
                    \begin{tikzpicture}
                    \node[anchor=south west,inner sep=0] (image) at (0,0) {
                \includegraphics[width=1\linewidth]{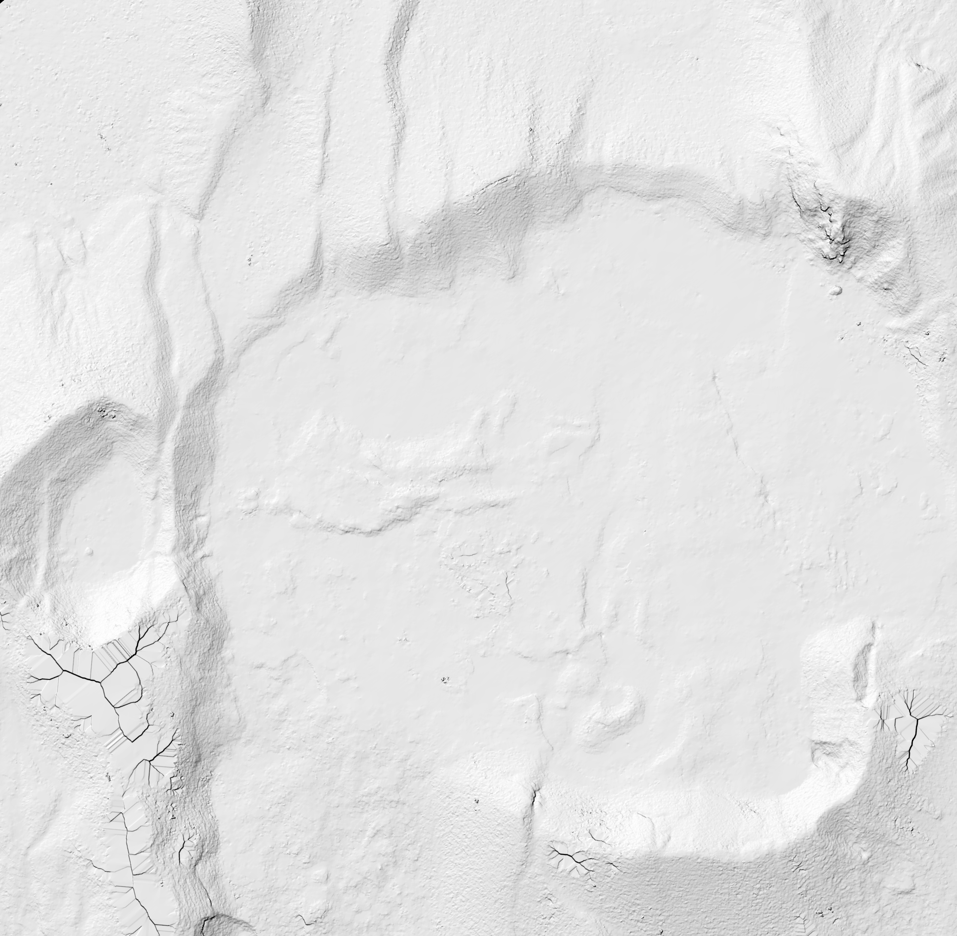}};
                    \begin{scope}[x={(image.south east)},y={(image.north west)}]
                    \draw[green,thick] (0.023,0.154) rectangle (0.186,0.323);
                    \draw[blue,thick] (0.840,0.140) rectangle (0.999,0.306);
                    \draw[orange,thick] (0.830,0.491) rectangle (0.992,0.655);
                    \draw[red,thick] (0.645,0.795) rectangle (0.802,0.628);
                    \end{scope}
                    \end{tikzpicture}
            \end{minipage}%
            \hspace{0mm}
            \begin{minipage}[t]{0.227\linewidth}
                \centering
                \includegraphics[width=1\linewidth]{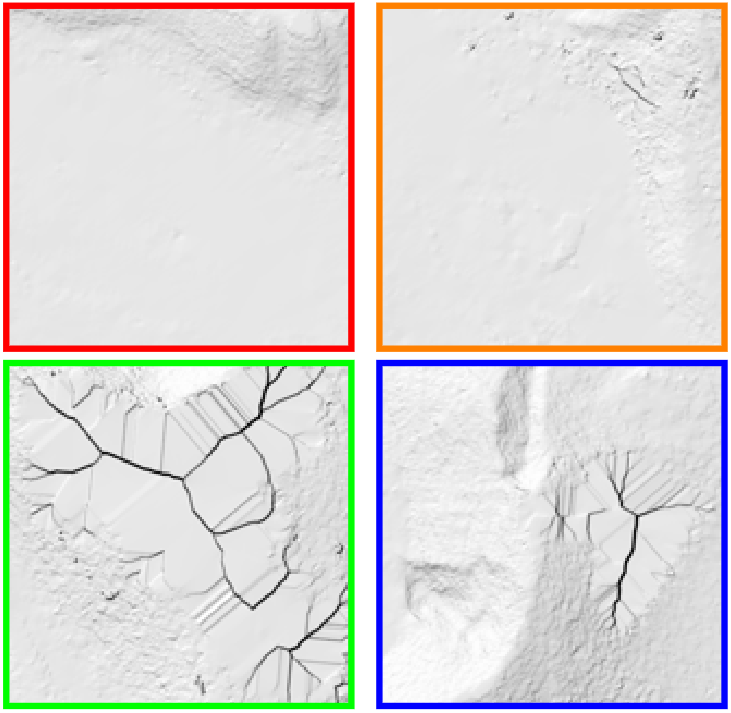}
            \end{minipage}%
            }
        \subfigure[GT Dji-B]{
            \begin{minipage}[t]{0.223\linewidth}
                \centering
                    \begin{tikzpicture}
                    \node[anchor=south west,inner sep=0] (image) at (0,0) {
                \includegraphics[width=1\linewidth]{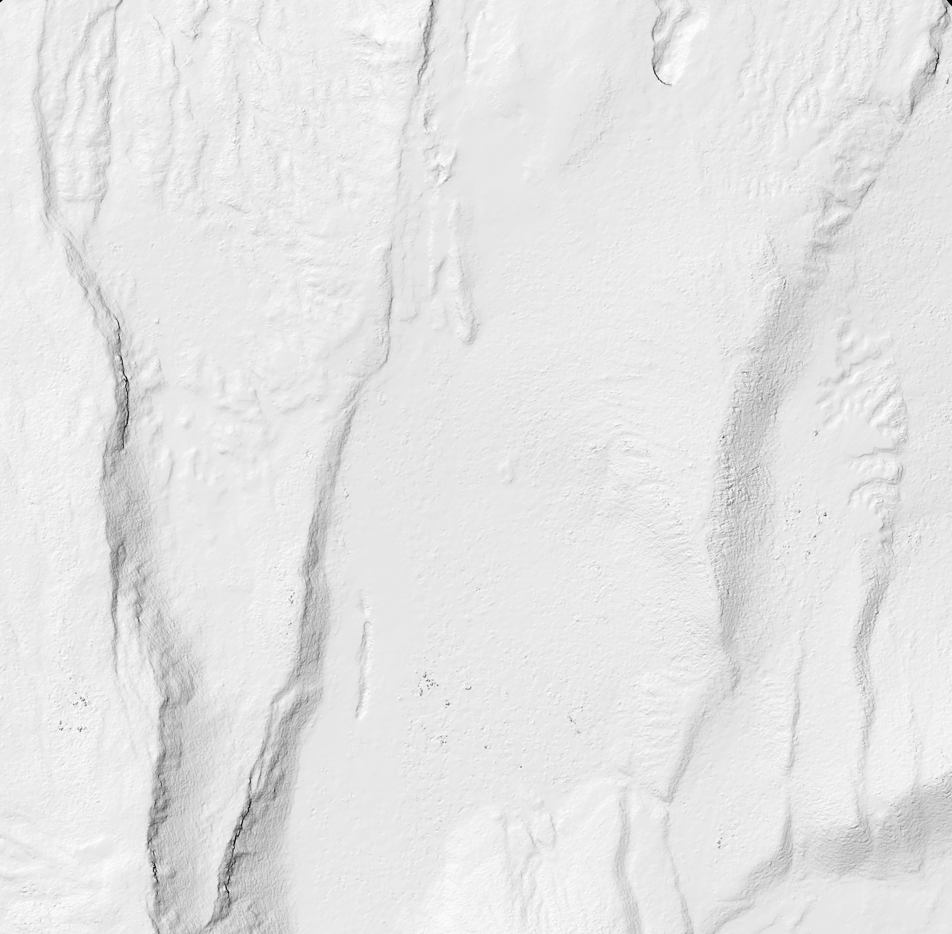}};
                    \begin{scope}[x={(image.south east)},y={(image.north west)}]
                    \draw[red,thick] (0.175,0.634) rectangle (0.345,0.460);
                    \draw[orange,thick] (0.595,0.816) rectangle (0.781,0.995);
                    \draw[green,thick] (0.420,0.271) rectangle (0.587,0.113);
                    \draw[blue,thick] (0.750,0.468) rectangle (0.919,0.310);
                    \end{scope}
                    \end{tikzpicture}
            \end{minipage}%
            \hspace{0mm}
            \begin{minipage}[t]{0.227\linewidth}
                \centering
                \includegraphics[width=1\linewidth]{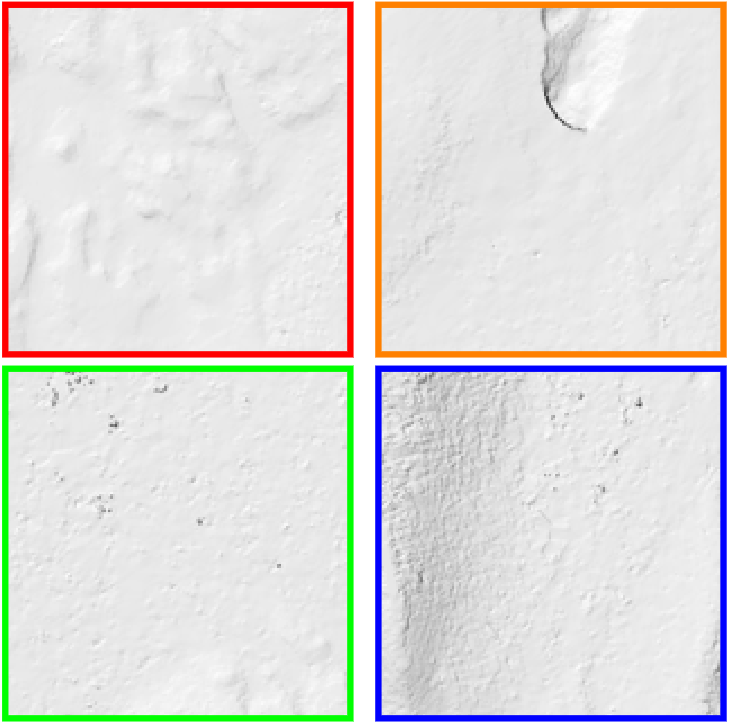}
            \end{minipage}%
            }
        \caption{\textbf{Altitude Estimation -- Djibouti Dataset.} \SatNeRF~fails to recover the altitudes. \SpSNeRF~recovers noisy altitudes. The noise is suppressed to some extent in \OurNeRFShort. SGM$_{Z1}$ generates generally more detailed surfaces, but underperforms in weakly textured areas (see green rectangle).}
        \label{maintestaltDji}
    \end{center}
\end{figure*}

\begin{figure*}[!htbp]
    \begin{center}
        \subfigure[\SatNeRF~Lzh-A]{
            \begin{minipage}[t]{0.223\linewidth}
                \centering
                    \begin{tikzpicture}
                    \node[anchor=south west,inner sep=0] (image) at (0,0) {
                \includegraphics[width=1\linewidth]{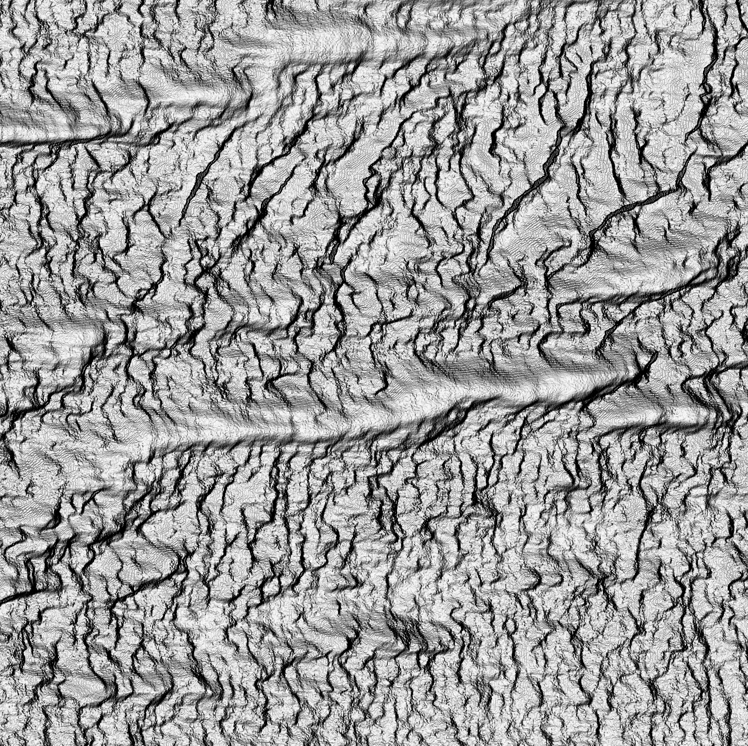}};
                    \begin{scope}[x={(image.south east)},y={(image.north west)}]
                    \draw[red,thick] (0.286,0.301) rectangle (0.486,0.494);
                    \draw[orange,thick] (0.567,0.737) rectangle (0.765,0.553);
                    \draw[green,thick] (0.307,0.000) rectangle (0.507,0.204);
                    \draw[blue,thick] (0.773,0.535) rectangle (0.975,0.728);
                    \end{scope}
                    \end{tikzpicture}
            \end{minipage}%
            \hspace{0mm}
            \begin{minipage}[t]{0.227\linewidth}
                \centering
                \includegraphics[width=1\linewidth]{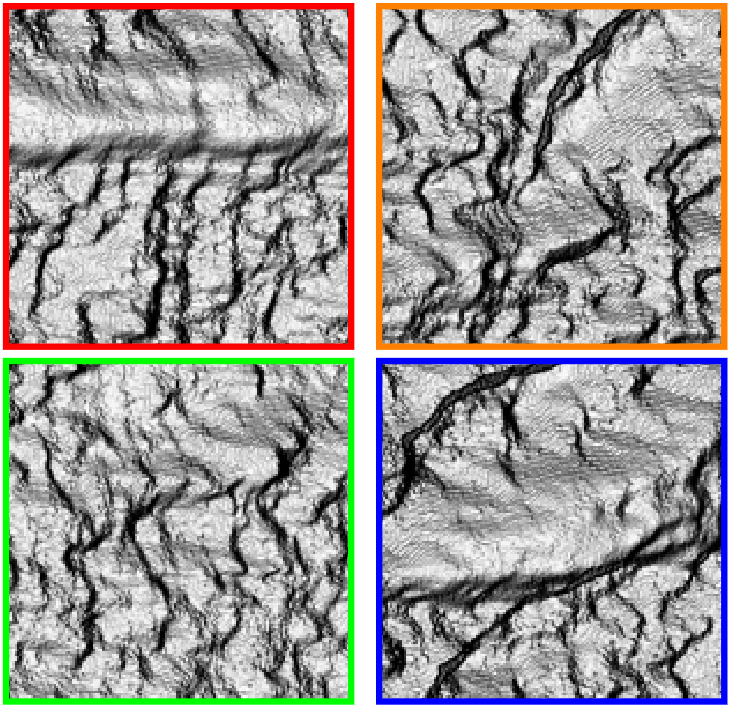}
            \end{minipage}%
            }
        \subfigure[\SatNeRF~Lzh-B]{
            \begin{minipage}[t]{0.223\linewidth}
                \centering
                    \begin{tikzpicture}
                    \node[anchor=south west,inner sep=0] (image) at (0,0) {
                \includegraphics[width=1\linewidth]{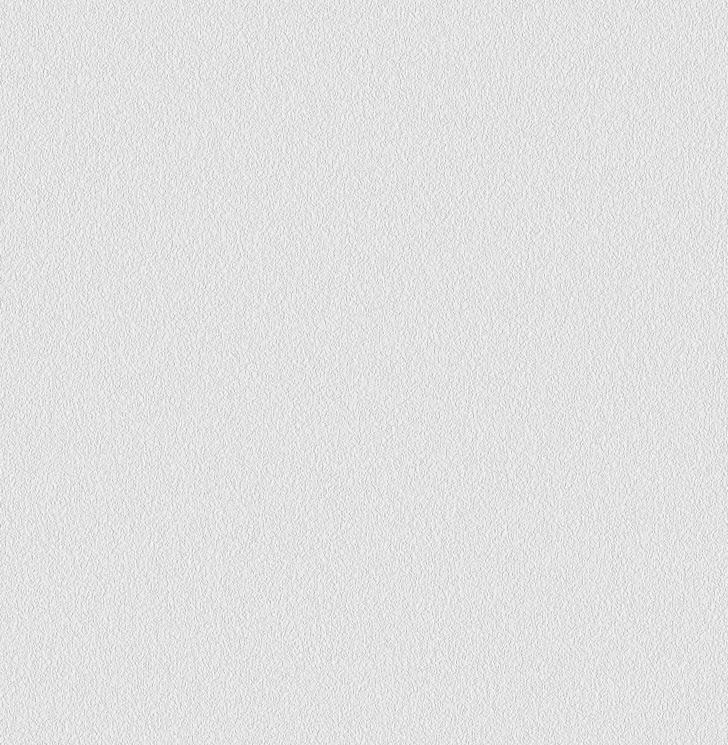}};
                    \begin{scope}[x={(image.south east)},y={(image.north west)}]
                    \draw[red,thick] (0.0,0.366) rectangle (0.207,0.580);
                    \draw[orange,thick] (0.803,0.805) rectangle (0.999,0.999);
                    \draw[green,thick] (0.344,0.134) rectangle (0.544,0.331);
                    \draw[blue,thick] (0.734,0.126) rectangle (0.938,0.328);
                    \end{scope}
                    \end{tikzpicture}
            \end{minipage}%
            \hspace{0mm}
            \begin{minipage}[t]{0.227\linewidth}
                \centering
                \includegraphics[width=1\linewidth]{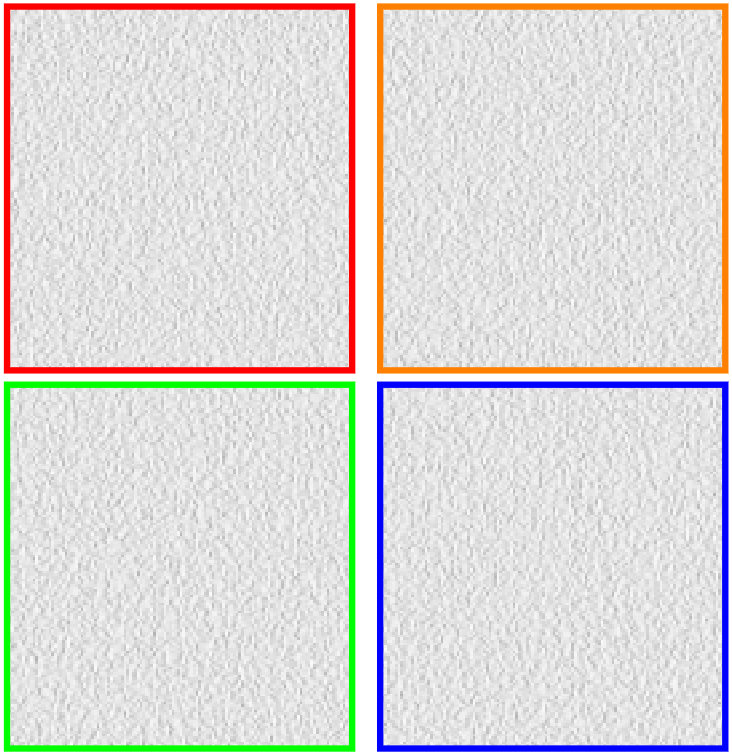}
            \end{minipage}%
            }
        \subfigure[\SpSNeRF~Lzh-A]{
            \begin{minipage}[t]{0.223\linewidth}
                \centering
                    \begin{tikzpicture}
                    \node[anchor=south west,inner sep=0] (image) at (0,0) {
                \includegraphics[width=1\linewidth]{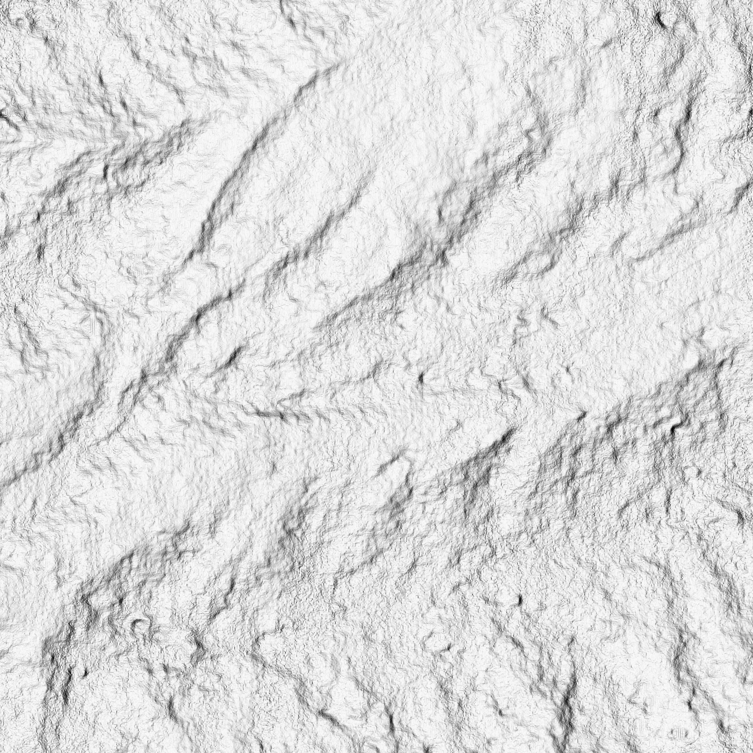}};
                    \begin{scope}[x={(image.south east)},y={(image.north west)}]
                    \draw[red,thick] (0.286,0.301) rectangle (0.486,0.494);
                    \draw[orange,thick] (0.567,0.737) rectangle (0.765,0.553);
                    \draw[green,thick] (0.307,0.000) rectangle (0.507,0.204);
                    \draw[blue,thick] (0.773,0.535) rectangle (0.975,0.728);
                    \end{scope}
                    \end{tikzpicture}
            \end{minipage}%
            \hspace{0mm}
            \begin{minipage}[t]{0.227\linewidth}
                \centering
                \includegraphics[width=1\linewidth]{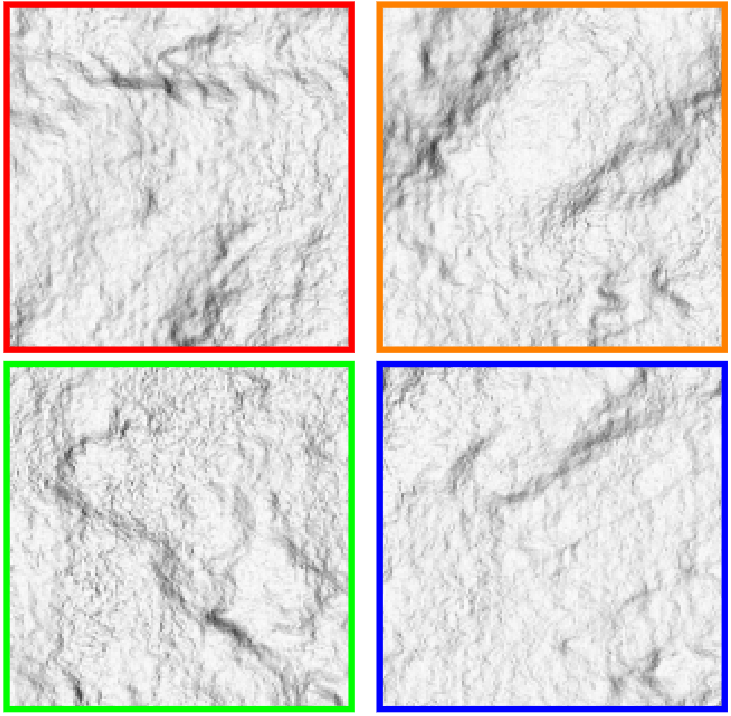}
            \end{minipage}%
            }
        \subfigure[\SpSNeRF~Lzh-B]{
            \begin{minipage}[t]{0.223\linewidth}
                \centering
                    \begin{tikzpicture}
                    \node[anchor=south west,inner sep=0] (image) at (0,0) {
                \includegraphics[width=1\linewidth]{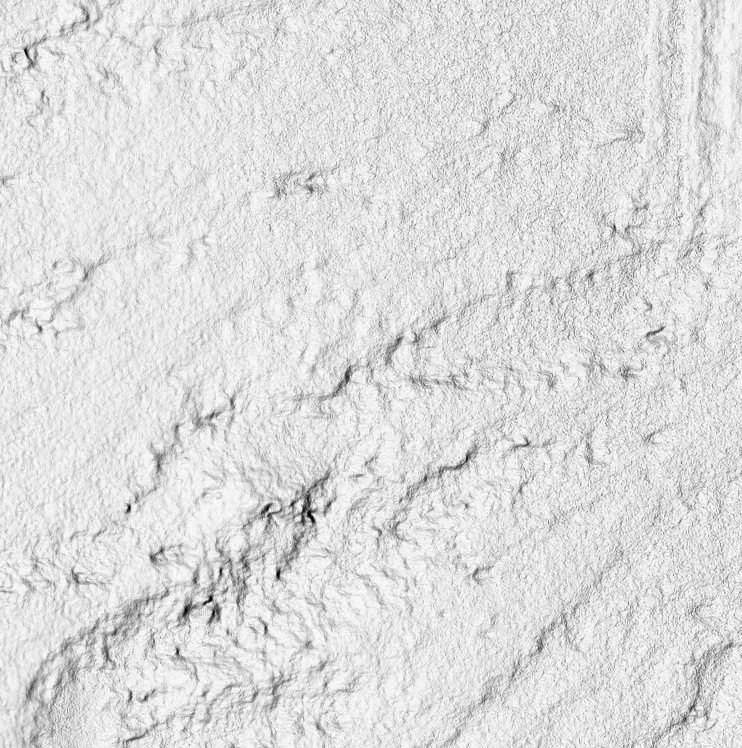}};
                    \begin{scope}[x={(image.south east)},y={(image.north west)}]
                    \draw[red,thick] (0.0,0.366) rectangle (0.207,0.580);
                    \draw[orange,thick] (0.803,0.805) rectangle (0.999,0.999);
                    \draw[green,thick] (0.344,0.134) rectangle (0.544,0.331);
                    \draw[blue,thick] (0.734,0.126) rectangle (0.938,0.328);
                    \end{scope}
                    \end{tikzpicture}
            \end{minipage}%
            \hspace{0mm}
            \begin{minipage}[t]{0.227\linewidth}
                \centering
                \includegraphics[width=1\linewidth]{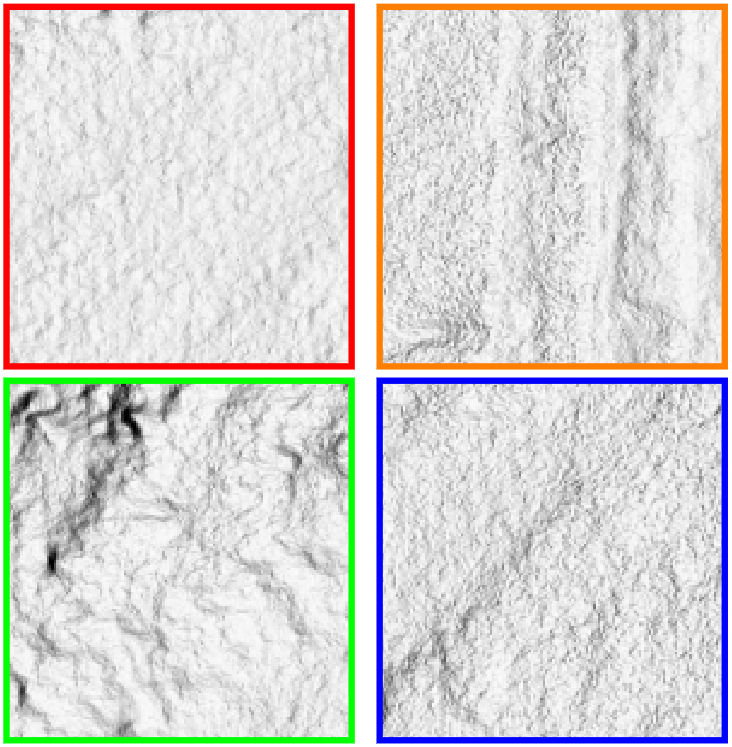}
            \end{minipage}%
            }
        \subfigure[\OurNeRFShort~Lzh-A]{
            \begin{minipage}[t]{0.223\linewidth}
                \centering
                    \begin{tikzpicture}
                    \node[anchor=south west,inner sep=0] (image) at (0,0) {
                \includegraphics[width=1\linewidth]{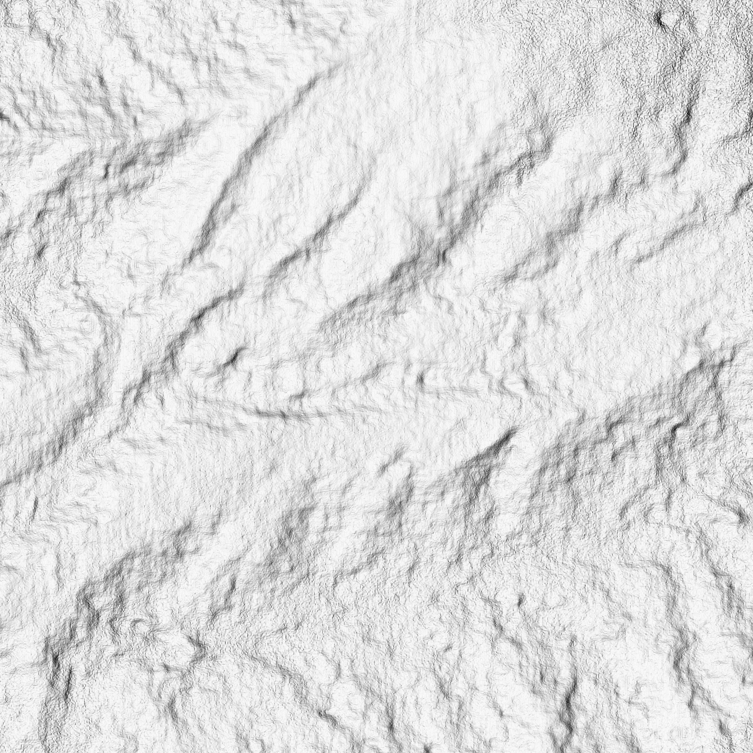}};
                    \begin{scope}[x={(image.south east)},y={(image.north west)}]
                    \draw[red,thick] (0.286,0.301) rectangle (0.486,0.494);
                    \draw[orange,thick] (0.567,0.737) rectangle (0.765,0.553);
                    \draw[green,thick] (0.307,0.000) rectangle (0.507,0.204);
                    \draw[blue,thick] (0.773,0.535) rectangle (0.975,0.728);
                    \end{scope}
                    \end{tikzpicture}
            \end{minipage}%
            \hspace{0mm}
            \begin{minipage}[t]{0.227\linewidth}
                \centering
                \includegraphics[width=1\linewidth]{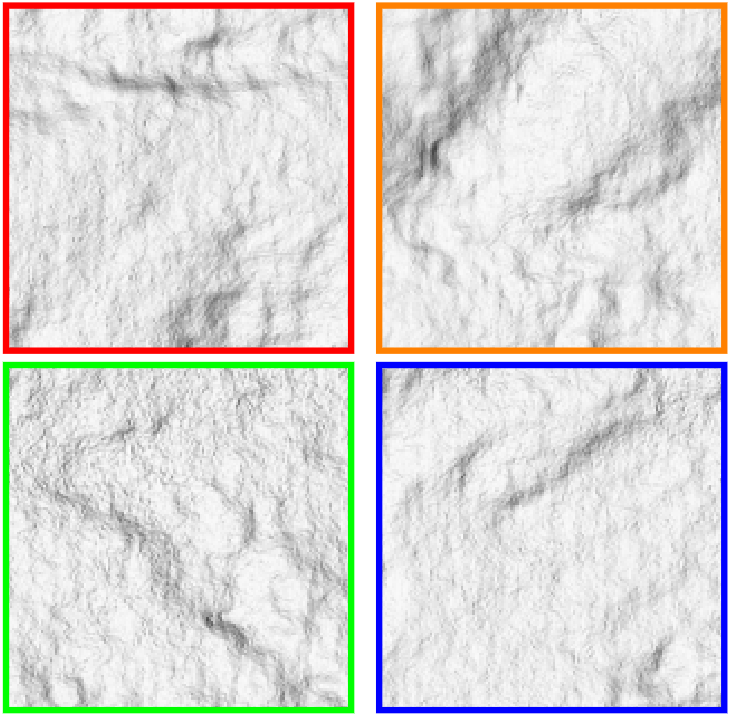}
            \end{minipage}%
            }
        \subfigure[\OurNeRFShort~Lzh-B]{
            \begin{minipage}[t]{0.223\linewidth}
                \centering
                    \begin{tikzpicture}
                    \node[anchor=south west,inner sep=0] (image) at (0,0) {
                \includegraphics[width=1\linewidth]{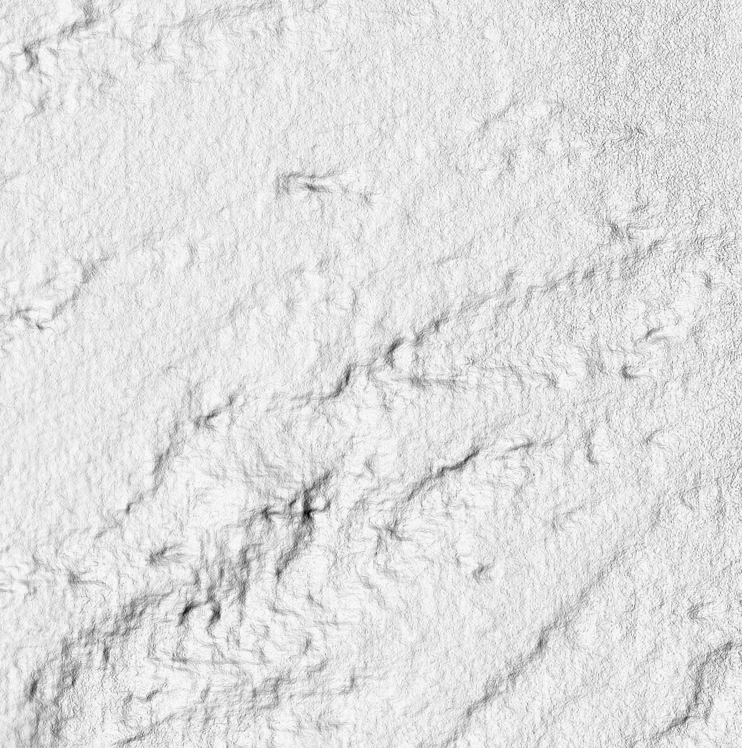}};
                    \begin{scope}[x={(image.south east)},y={(image.north west)}]
                    \draw[red,thick] (0.0,0.366) rectangle (0.207,0.580);
                    \draw[orange,thick] (0.803,0.805) rectangle (0.999,0.999);
                    \draw[green,thick] (0.344,0.134) rectangle (0.544,0.331);
                    \draw[blue,thick] (0.734,0.126) rectangle (0.938,0.328);
                    \end{scope}
                    \end{tikzpicture}
            \end{minipage}%
            \hspace{0mm}
            \begin{minipage}[t]{0.227\linewidth}
                \centering
                \includegraphics[width=1\linewidth]{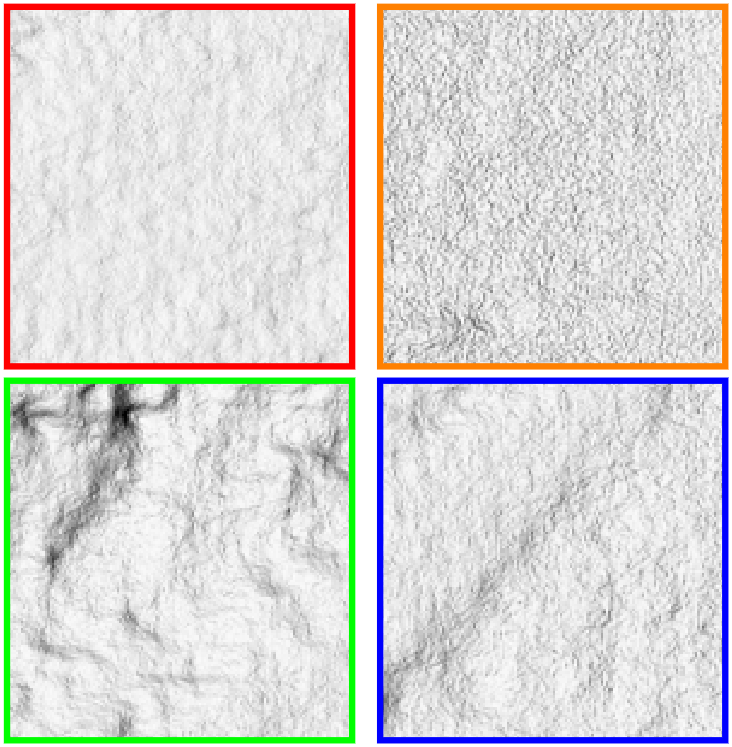}
            \end{minipage}%
            }    
        \subfigure[SGM$_{Z1}$ Lzh-A]{
            \begin{minipage}[t]{0.223\linewidth}
                \centering
                    \begin{tikzpicture}
                    \node[anchor=south west,inner sep=0] (image) at (0,0) {
                \includegraphics[width=1\linewidth]{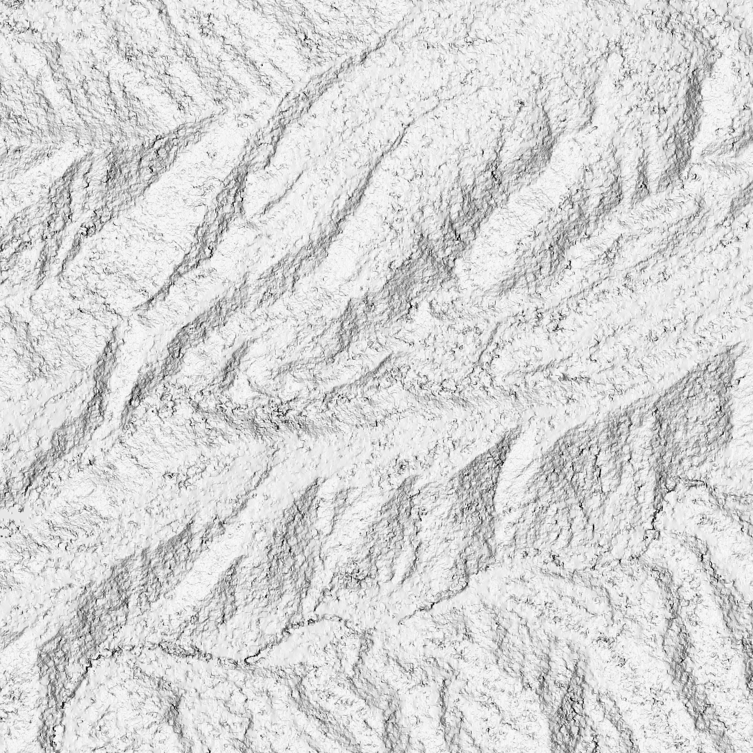}};
                    \begin{scope}[x={(image.south east)},y={(image.north west)}]
                    \draw[red,thick] (0.286,0.301) rectangle (0.486,0.494);
                    \draw[orange,thick] (0.567,0.737) rectangle (0.765,0.553);
                    \draw[green,thick] (0.307,0.000) rectangle (0.507,0.204);
                    \draw[blue,thick] (0.773,0.535) rectangle (0.975,0.728);
                    \end{scope}
                    \end{tikzpicture}
            \end{minipage}%
            \hspace{0mm}
            \begin{minipage}[t]{0.227\linewidth}
                \centering
                \includegraphics[width=1\linewidth]{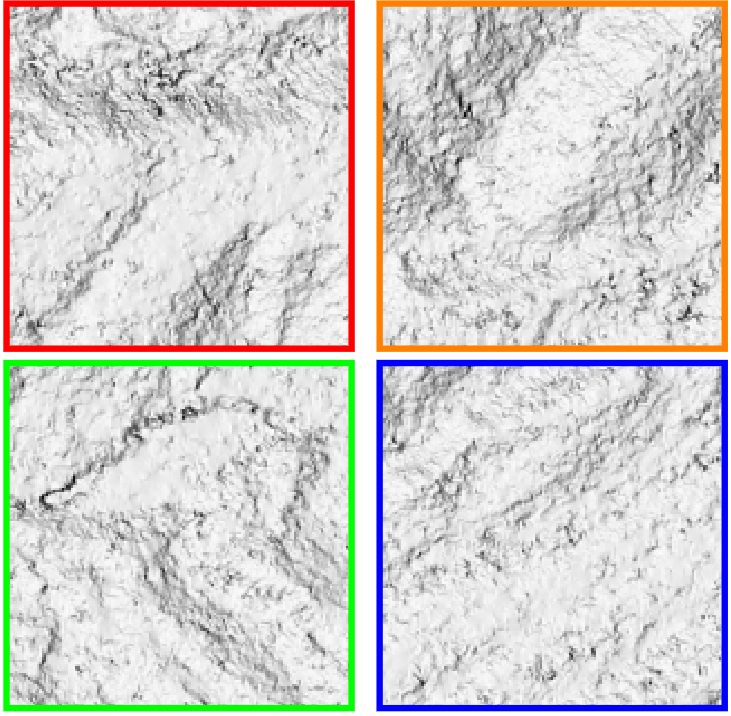}
            \end{minipage}%
            }
        \subfigure[SGM$_{Z1}$ Lzh-B]{
            \begin{minipage}[t]{0.223\linewidth}
                \centering
                    \begin{tikzpicture}
                    \node[anchor=south west,inner sep=0] (image) at (0,0) {
                \includegraphics[width=1\linewidth]{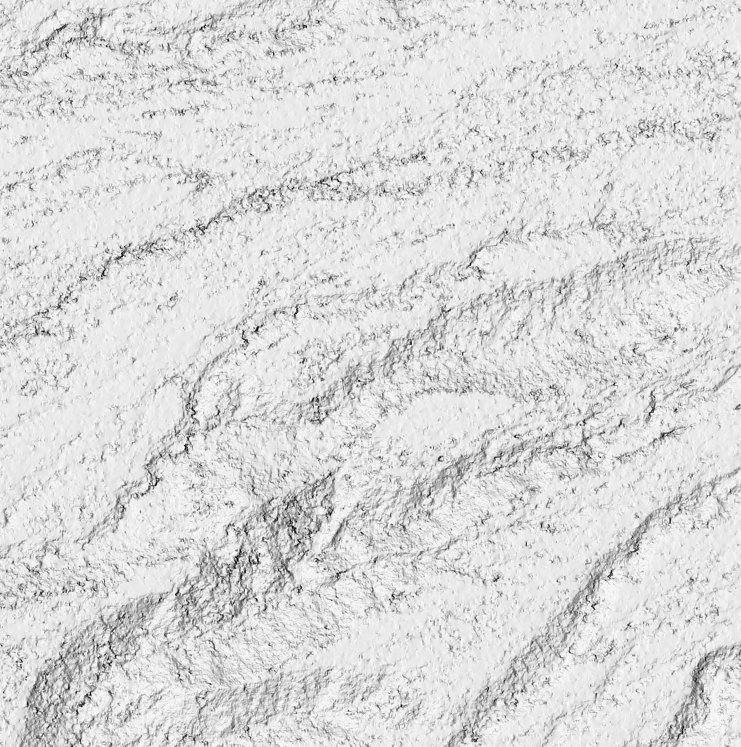}};
                    \begin{scope}[x={(image.south east)},y={(image.north west)}]
                    \draw[red,thick] (0.0,0.366) rectangle (0.207,0.580);
                    \draw[orange,thick] (0.803,0.805) rectangle (0.999,0.999);
                    \draw[green,thick] (0.344,0.134) rectangle (0.544,0.331);
                    \draw[blue,thick] (0.734,0.126) rectangle (0.938,0.328);
                    \end{scope}
                    \end{tikzpicture}
            \end{minipage}%
            \hspace{0mm}
            \begin{minipage}[t]{0.227\linewidth}
                \centering
                \includegraphics[width=1\linewidth]{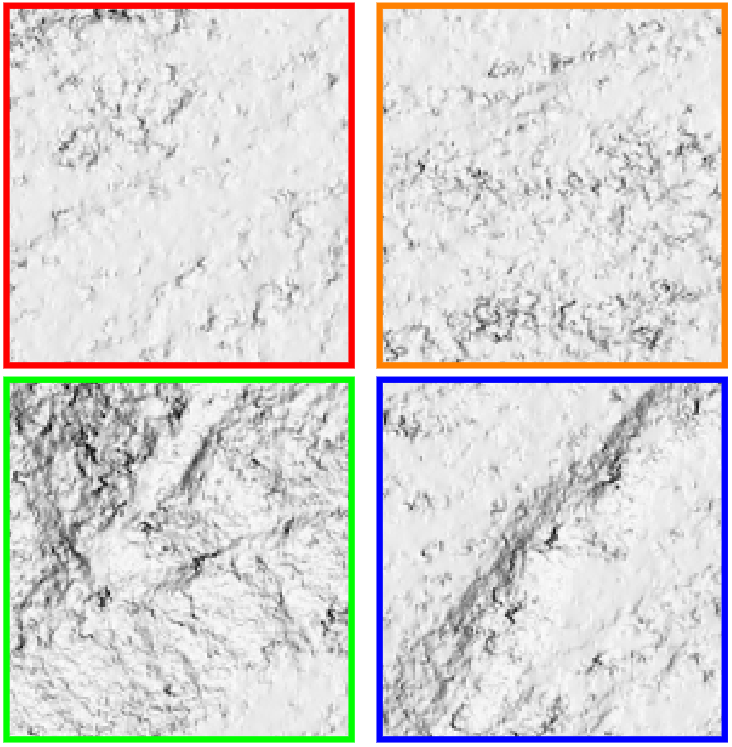}
            \end{minipage}%
            }
        \subfigure[GT Lzh-A]{
            \begin{minipage}[t]{0.223\linewidth}
                \centering
                    \begin{tikzpicture}
                    \node[anchor=south west,inner sep=0] (image) at (0,0) {
                \includegraphics[width=1\linewidth]{Lzh_013_GT_dsm.png}};
                    \begin{scope}[x={(image.south east)},y={(image.north west)}]
                    \draw[red,thick] (0.286,0.301) rectangle (0.486,0.494);
                    \draw[orange,thick] (0.567,0.737) rectangle (0.765,0.553);
                    \draw[green,thick] (0.307,0.000) rectangle (0.507,0.204);
                    \draw[blue,thick] (0.773,0.535) rectangle (0.975,0.728);
                    \end{scope}
                    \end{tikzpicture}
            \end{minipage}%
            \hspace{0mm}
            \begin{minipage}[t]{0.227\linewidth}
                \centering
                \includegraphics[width=1\linewidth]{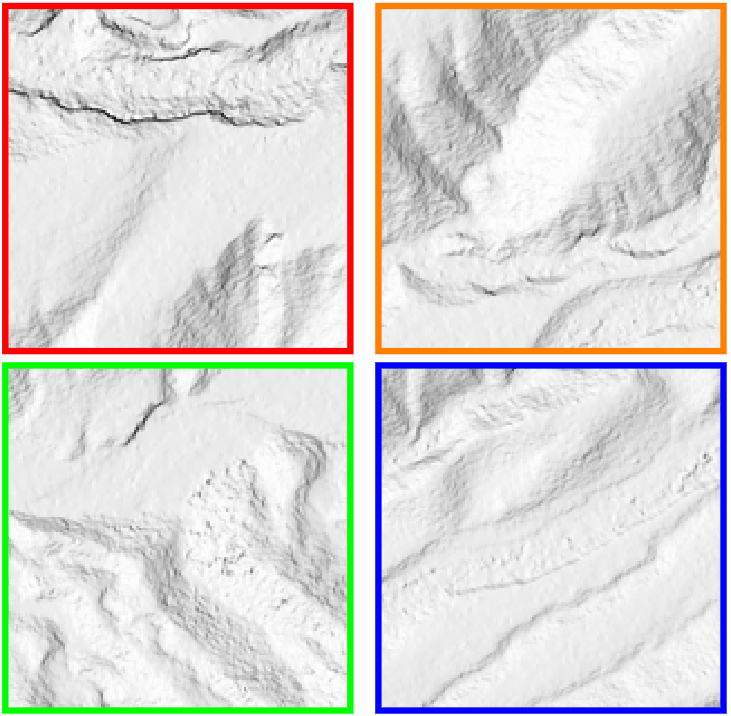}
            \end{minipage}%
            }
        \subfigure[GT Lzh-B]{
            \begin{minipage}[t]{0.223\linewidth}
                \centering
                    \begin{tikzpicture}
                    \node[anchor=south west,inner sep=0] (image) at (0,0) {
                \includegraphics[width=1\linewidth]{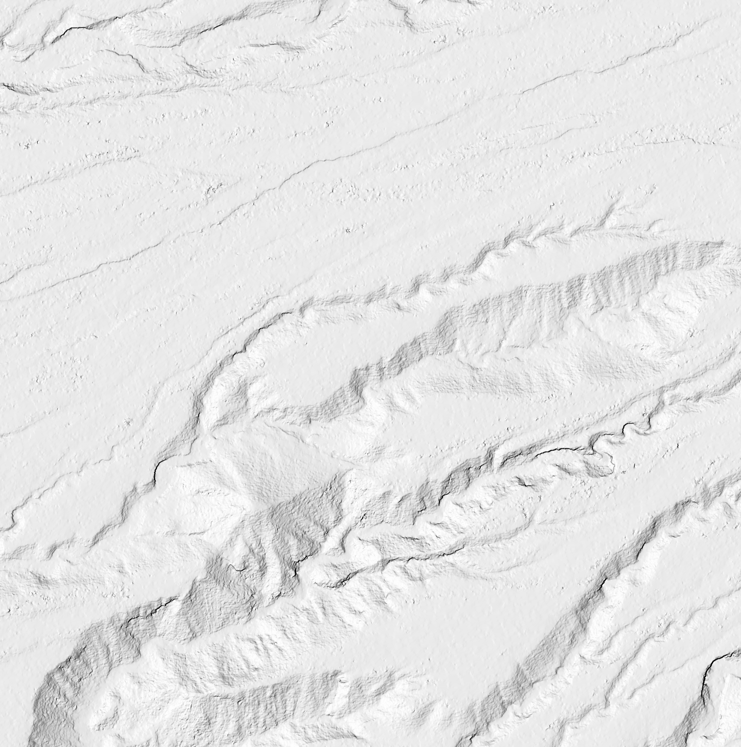}};
                    \begin{scope}[x={(image.south east)},y={(image.north west)}]
                    \draw[red,thick] (0.0,0.366) rectangle (0.207,0.580);
                    \draw[orange,thick] (0.803,0.805) rectangle (0.999,0.999);
                    \draw[green,thick] (0.344,0.134) rectangle (0.544,0.331);
                    \draw[blue,thick] (0.734,0.126) rectangle (0.938,0.328);
                    \end{scope}
                    \end{tikzpicture}
            \end{minipage}%
            \hspace{0mm}
            \begin{minipage}[t]{0.227\linewidth}
                \centering
                \includegraphics[width=1\linewidth]{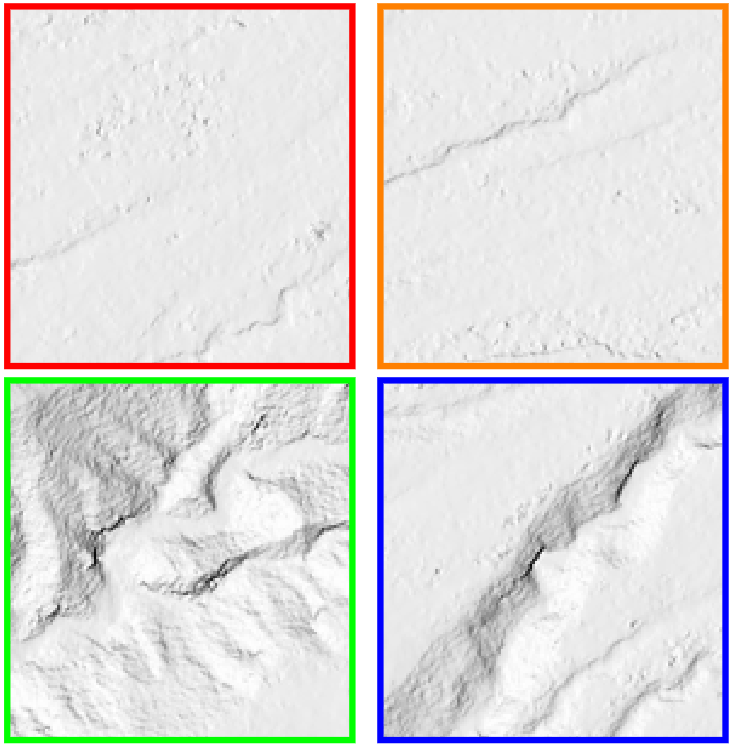}
            \end{minipage}%
            }
        \caption{\textbf{Altitude Estimation -- Lanzhou Dataset.} \SatNeRF~fails to recover the altitudes. Surfaces obtained by \SpSNeRF~and \OurNeRFShort~are similar, with \SpSNeRF~manifesting slightly noisier altitude predictions and a hallucination artefact. SGM$_{Z1}$ recovers surfaces with more detailed topography but more noise than \SpSNeRF~and \OurNeRFShort.}
        \label{maintestaltLzh}
    \end{center}
\end{figure*}

Quantitative metrics are presented in \Cref{maintestquantitativemetrics}, whereas qualitative visualisations are provided in \Cref{maintestaltDji,maintestaltLzh}. 
\SatNeRF, which was not designed for scenarios with few images, estimates surface altitudes that are tens of meters away from the ground truth surface. \SpSNeRF~performs better, thanks to the dense depth supervision, as opposed to supervision with sparse points in \SatNeRF. Visual assessment reveals that altitudes predicted by \SatNeRF~are either flat (\Cref{maintestaltDji} (a-b)), or contain a made up pattern (\Cref{maintestaltLzh} (a)). \SpSNeRF~altitudes are more faithful to ground truth but remain noisy. Our \OurNeRFShort~outperforms both versions of \NERF, producing less noisy surfaces while retaining detail.

\noindent Compared with surfaces obtained with the classical stereo matching \citep{mpd:06:sgm}, \OurNeRFShort~appears smoother and less detailed (compare \Cref{maintestaltLzh} (f) and (h)) in areas with good texture. However, on poorly textured areas where stereo matching is challenging, our \OurNeRFShort~predicts coherent altitudes (compare \Cref{maintestaltDji} (e) and (g)). Note also that our ground truth surface was generated with stereo matching algorithms, thus the comparison is possibly slightly biasing the MAE metric in favour of the stereo matching surface.
 


\subsection{Ablations}
\label{Findthebesttrainingstrategy}
\begin{figure}[!htbp]
	\begin{center}
			\begin{minipage}[t]{0.6\linewidth}
				\centering
				\includegraphics[width=1.0\linewidth]{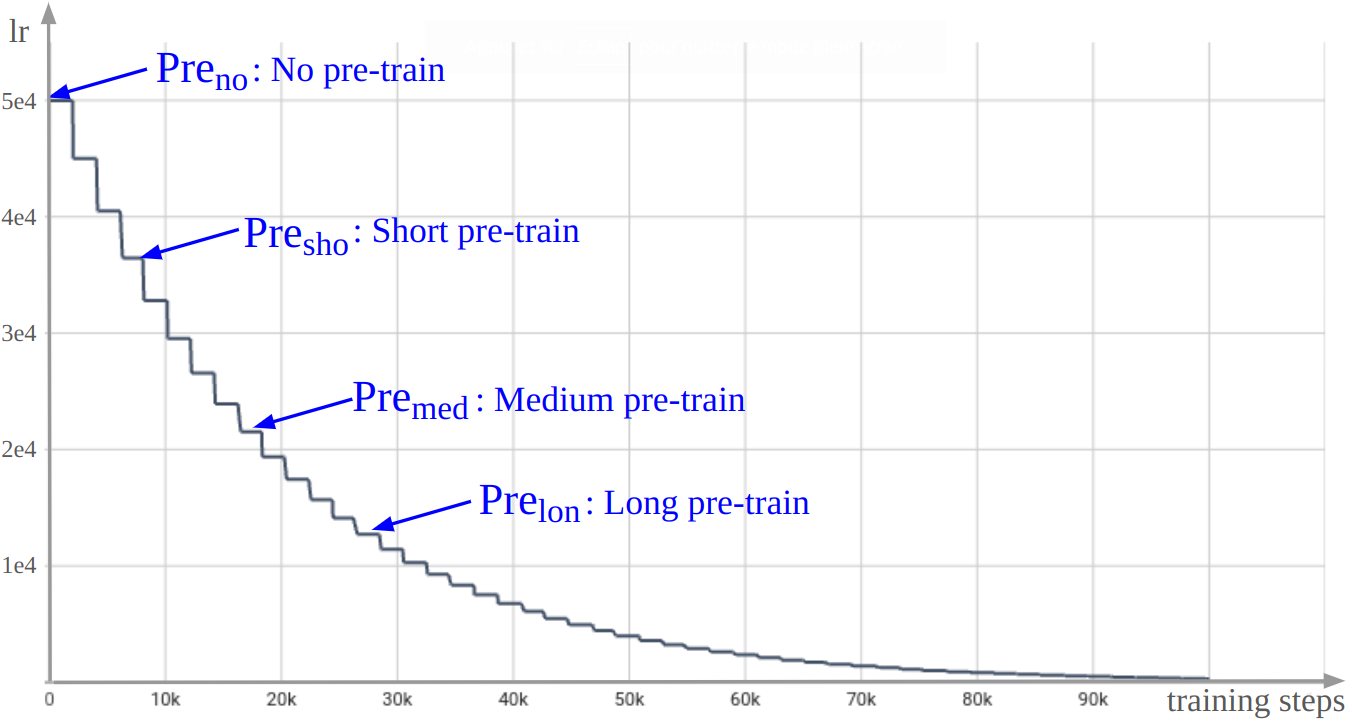}
			\end{minipage}%
		\caption{\textbf{Training Strategies.} Fours training strategies of \OurNeRFShort~from no pre-training to long pre-training.}
		\label{pre-trainstrategy}
	\end{center}
\end{figure}

\paragraph{Training strategy}
Our \OurNeRFShort~model is trained progressively to ensure proper initialisation of the geometry (i.e., density weights in \NERF) before learning the RPV parameters. We perform an ablation with different pre-training strategies to determine the optimal point at which the transition from Lambertian to RPV model should occur. In the Pre$_{no}$ approach, the entire network is trained without pre-training. In the Pre$_{sho}$, Pre$_{med}$ and Pre$_{lon}$ approaches, we start from the Lambertian assumption and switch to the RPV model at different  training steps, as shown in \Cref{pre-trainstrategy}. The initial learning rate is set to 5e4 and decreases to {3.65e5, 2.15e5 and 1.27e5, respectively.}

\begin{figure*}[!htbp]
    \begin{center}
        \subfigure[Pre$_{no}$, $\lambda=\frac{10}{3}$]{
            \begin{minipage}[t]{0.205\linewidth}
                \centering
                    \begin{tikzpicture}
                    \node[anchor=south west,inner sep=0] (image) at (0,0) {
                \includegraphics[width=1\linewidth]{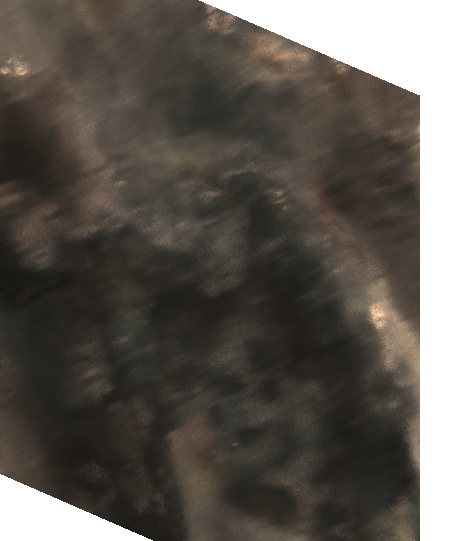}};
                    \begin{scope}[x={(image.south east)},y={(image.north west)}]
                    \draw[green,thick] (0.0,0.357) rectangle (0.165,0.214);
                    \draw[orange,thick] (0.168,0.315) rectangle (0.336,0.173);
                    \draw[red,thick] (0.240,0.635) rectangle (0.410,0.782);
                    \draw[blue,thick] (0.730,0.000) rectangle (0.930,0.160);
                    \end{scope}
                    \end{tikzpicture}
            \end{minipage}%
            \begin{minipage}[t]{0.245\linewidth}
                \centering
                \includegraphics[width=1\linewidth]{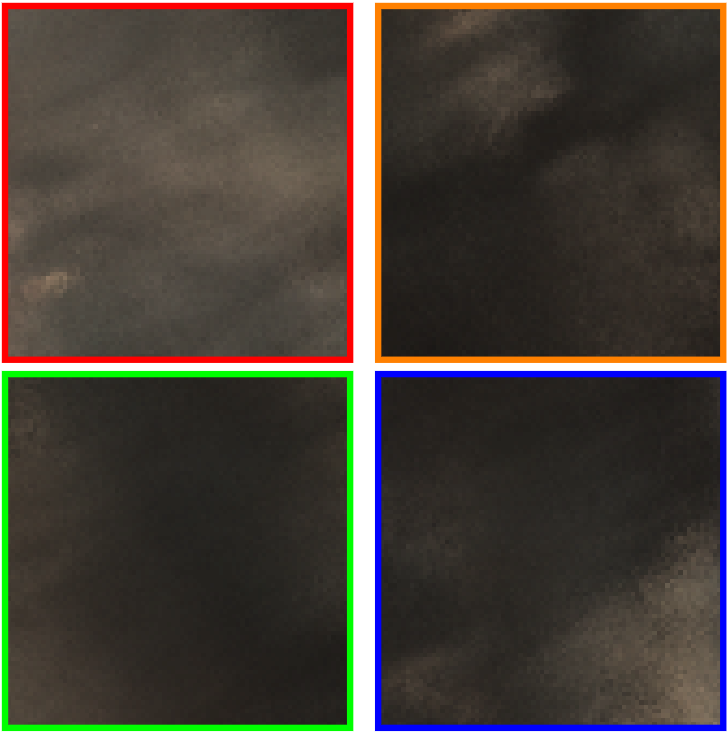}
            \end{minipage}%
        }        
        \subfigure[Pre$_{sho}$, $\lambda=\frac{10}{3}$]{
            \begin{minipage}[t]{0.205\linewidth}
                \centering
                    \begin{tikzpicture}
                    \node[anchor=south west,inner sep=0] (image) at (0,0) {
                \includegraphics[width=1\linewidth]{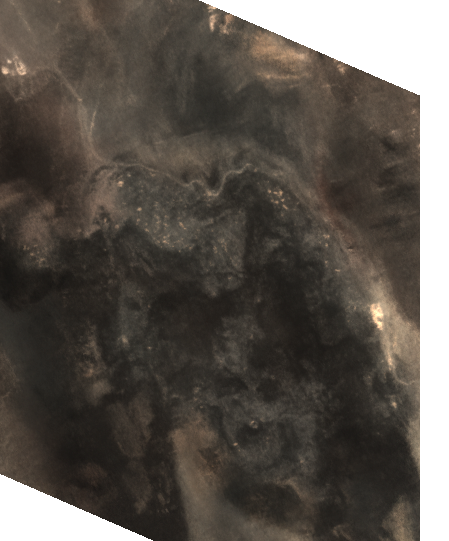}};
                    \begin{scope}[x={(image.south east)},y={(image.north west)}]
                    \draw[green,thick] (0.0,0.357) rectangle (0.165,0.214);
                    \draw[orange,thick] (0.168,0.315) rectangle (0.336,0.173);
                    \draw[red,thick] (0.240,0.635) rectangle (0.410,0.782);
                    \draw[blue,thick] (0.730,0.000) rectangle (0.930,0.160);
                    \end{scope}
                    \end{tikzpicture}
            \end{minipage}%
            \begin{minipage}[t]{0.245\linewidth}
                \centering
                \includegraphics[width=1\linewidth]{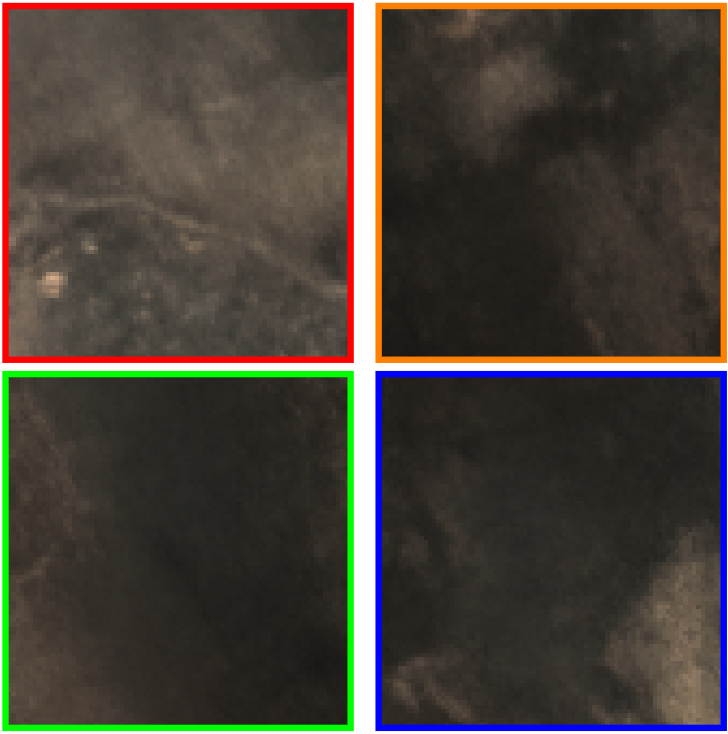}
            \end{minipage}%
        }
        \subfigure[Pre$_{med}$, $\lambda=\frac{10}{3}$]{
            \begin{minipage}[t]{0.205\linewidth}
                \centering
                    \begin{tikzpicture}
                    \node[anchor=south west,inner sep=0] (image) at (0,0) {
                \includegraphics[width=1\linewidth]{Nji_012_SpS_RPV-ir0-lr2152_val_3.png}};
                    \begin{scope}[x={(image.south east)},y={(image.north west)}]
                    \draw[green,thick] (0.0,0.357) rectangle (0.165,0.214);
                    \draw[orange,thick] (0.168,0.315) rectangle (0.336,0.173);
                    \draw[red,thick] (0.240,0.635) rectangle (0.410,0.782);
                    \draw[blue,thick] (0.730,0.000) rectangle (0.930,0.160);
                    \end{scope}
                    \end{tikzpicture}
            \end{minipage}%
            \begin{minipage}[t]{0.245\linewidth}
                \centering
                \includegraphics[width=1\linewidth]{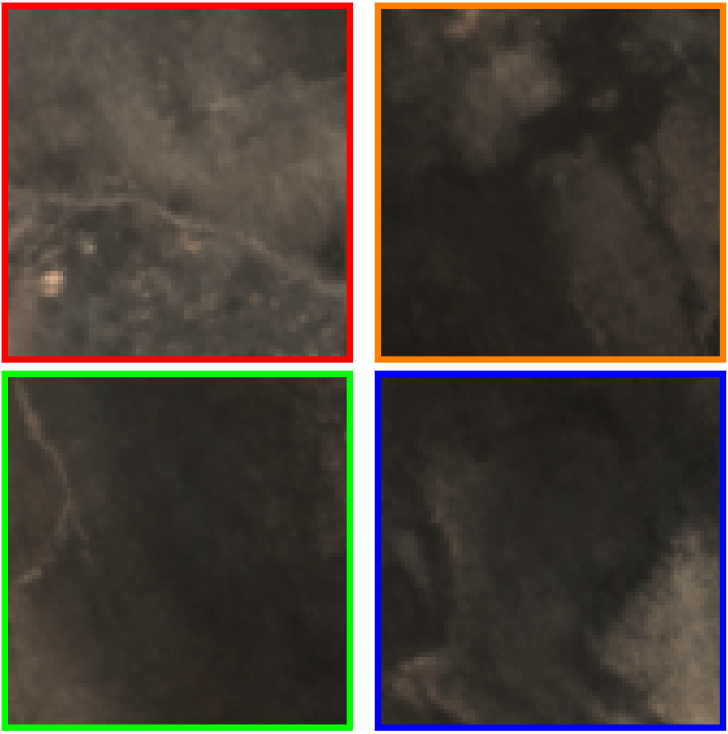}
            \end{minipage}%
        }        
        \subfigure[Pre$_{Lon}$, $\lambda=\frac{10}{3}$]{
            \begin{minipage}[t]{0.205\linewidth}
                \centering
                    \begin{tikzpicture}
                    \node[anchor=south west,inner sep=0] (image) at (0,0) {
                \includegraphics[width=1\linewidth]{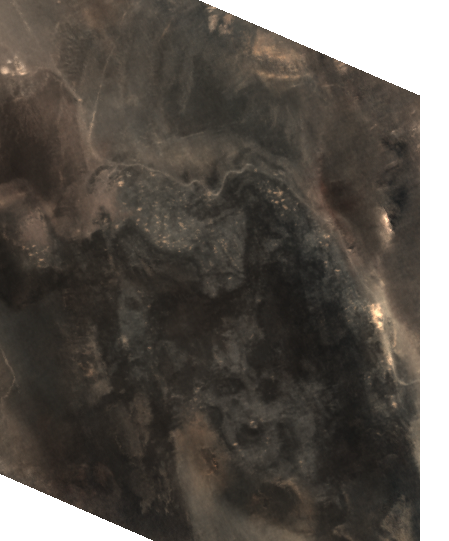}};
                    \begin{scope}[x={(image.south east)},y={(image.north west)}]
                    \draw[green,thick] (0.0,0.357) rectangle (0.165,0.214);
                    \draw[orange,thick] (0.168,0.315) rectangle (0.336,0.173);
                    \draw[red,thick] (0.240,0.635) rectangle (0.410,0.782);
                    \draw[blue,thick] (0.730,0.000) rectangle (0.930,0.160);
                    \end{scope}
                    \end{tikzpicture}
            \end{minipage}%
            \begin{minipage}[t]{0.245\linewidth}
                \centering
                \includegraphics[width=1\linewidth]{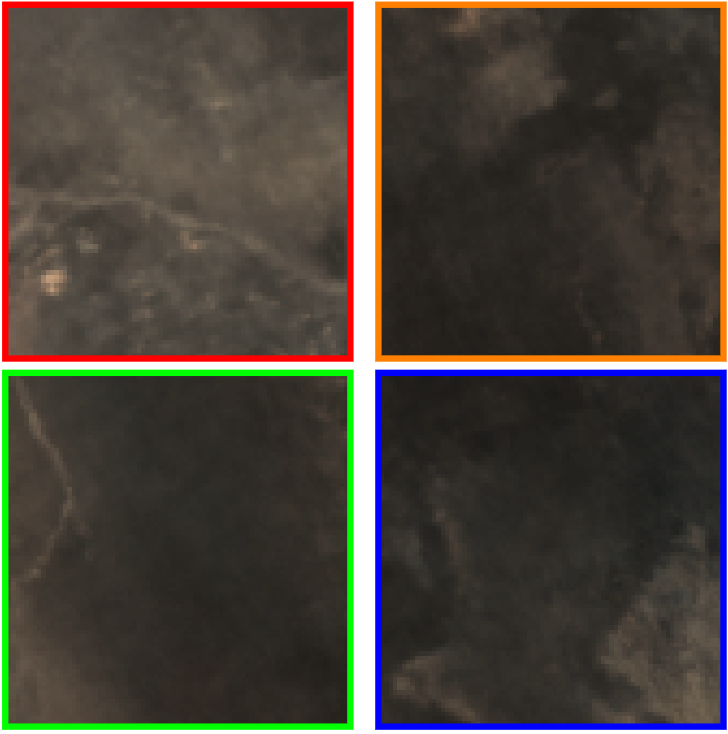}
            \end{minipage}%
        }        
        \subfigure[Pre$_{med}$, $\lambda=0$]{
            \begin{minipage}[t]{0.205\linewidth}
                \centering
                    \begin{tikzpicture}
                    \node[anchor=south west,inner sep=0] (image) at (0,0) {
                \includegraphics[width=1\linewidth]{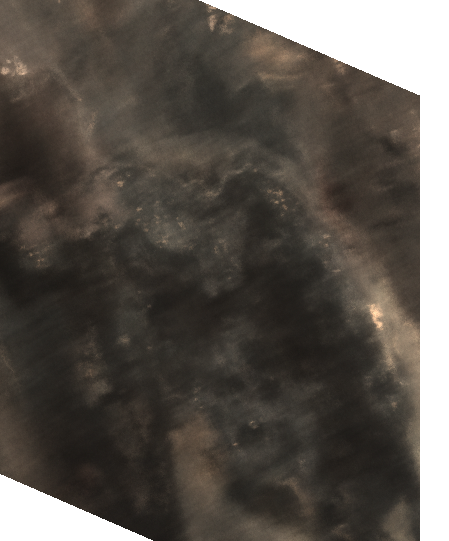}};
                    \begin{scope}[x={(image.south east)},y={(image.north west)}]
                    \draw[green,thick] (0.0,0.357) rectangle (0.165,0.214);
                    \draw[orange,thick] (0.168,0.315) rectangle (0.336,0.173);
                    \draw[red,thick] (0.240,0.635) rectangle (0.410,0.782);
                    \draw[blue,thick] (0.730,0.000) rectangle (0.930,0.160);
                    \end{scope}
                    \end{tikzpicture}
            \end{minipage}%
            \begin{minipage}[t]{0.245\linewidth}
                \centering
                \includegraphics[width=1\linewidth]{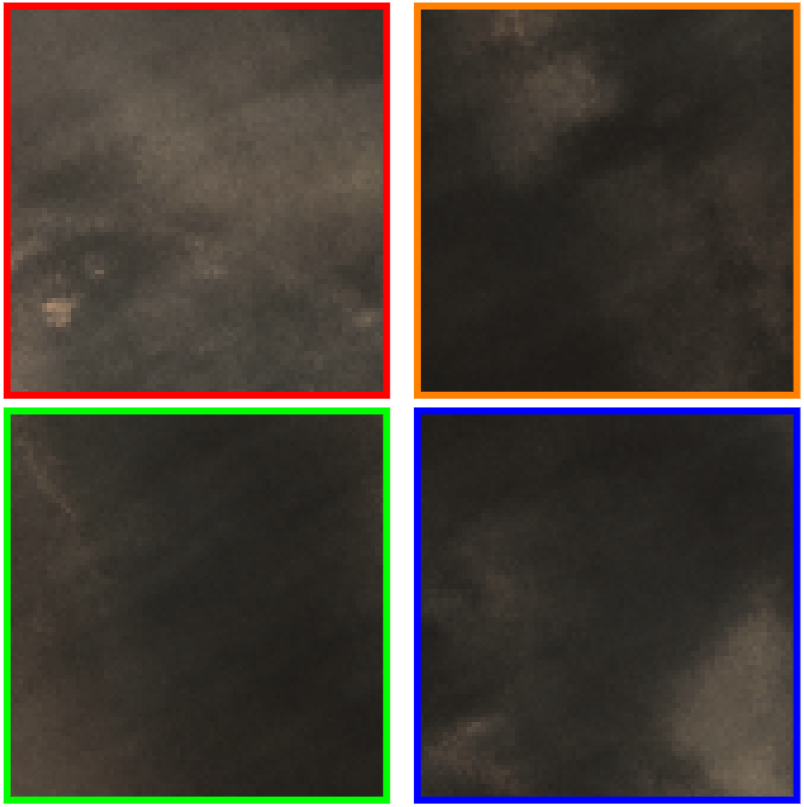}
            \end{minipage}%
        }
        \subfigure[Pre$_{med}$, $\lambda=\frac{1}{3}$]{
            \begin{minipage}[t]{0.205\linewidth}
                \centering
                    \begin{tikzpicture}
                    \node[anchor=south west,inner sep=0] (image) at (0,0) {
                \includegraphics[width=1\linewidth]{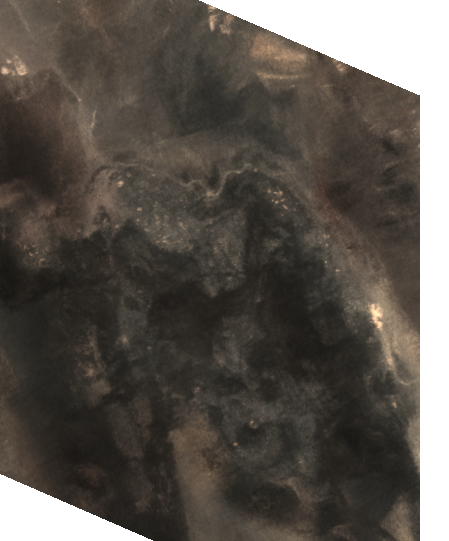}};
                    \begin{scope}[x={(image.south east)},y={(image.north west)}]
                    \draw[green,thick] (0.0,0.357) rectangle (0.165,0.214);
                    \draw[orange,thick] (0.168,0.315) rectangle (0.336,0.173);
                    \draw[red,thick] (0.240,0.635) rectangle (0.410,0.782);
                    \draw[blue,thick] (0.730,0.000) rectangle (0.930,0.160);
                    \end{scope}
                    \end{tikzpicture}
            \end{minipage}%
            \begin{minipage}[t]{0.245\linewidth}
                \centering
                \includegraphics[width=1\linewidth]{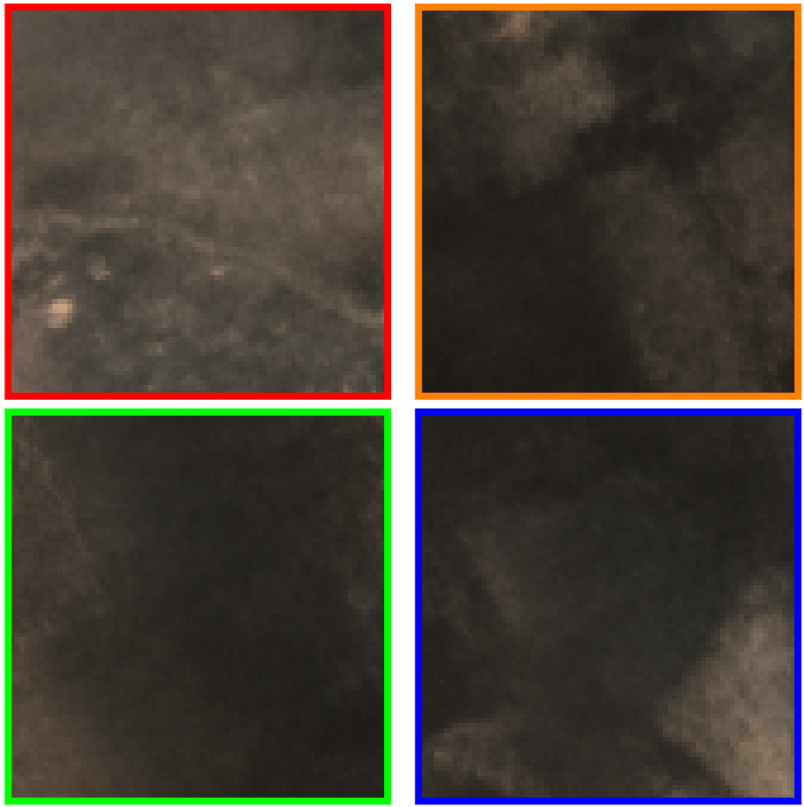}
            \end{minipage}%
        }        
        \subfigure[Pre$_{med}$, $\lambda=\frac{50}{3}$]{
            \begin{minipage}[t]{0.205\linewidth}
                \centering
                    \begin{tikzpicture}
                    \node[anchor=south west,inner sep=0] (image) at (0,0) {
                \includegraphics[width=1\linewidth]{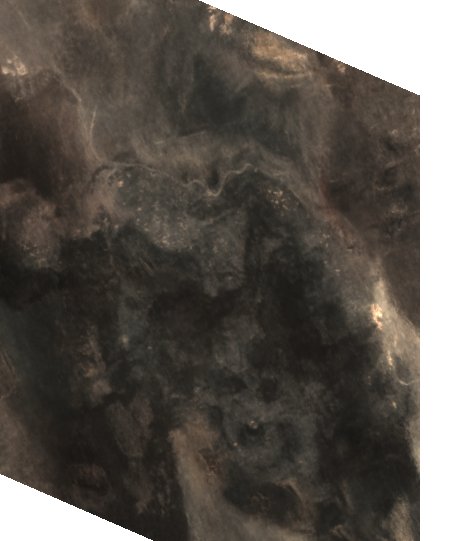}};
                    \begin{scope}[x={(image.south east)},y={(image.north west)}]
                    \draw[green,thick] (0.0,0.357) rectangle (0.165,0.214);
                    \draw[orange,thick] (0.168,0.315) rectangle (0.336,0.173);
                    \draw[red,thick] (0.240,0.635) rectangle (0.410,0.782);
                    \draw[blue,thick] (0.730,0.000) rectangle (0.930,0.160);
                    \end{scope}
                    \end{tikzpicture}
            \end{minipage}%
            \begin{minipage}[t]{0.245\linewidth}
                \centering
                \includegraphics[width=1\linewidth]{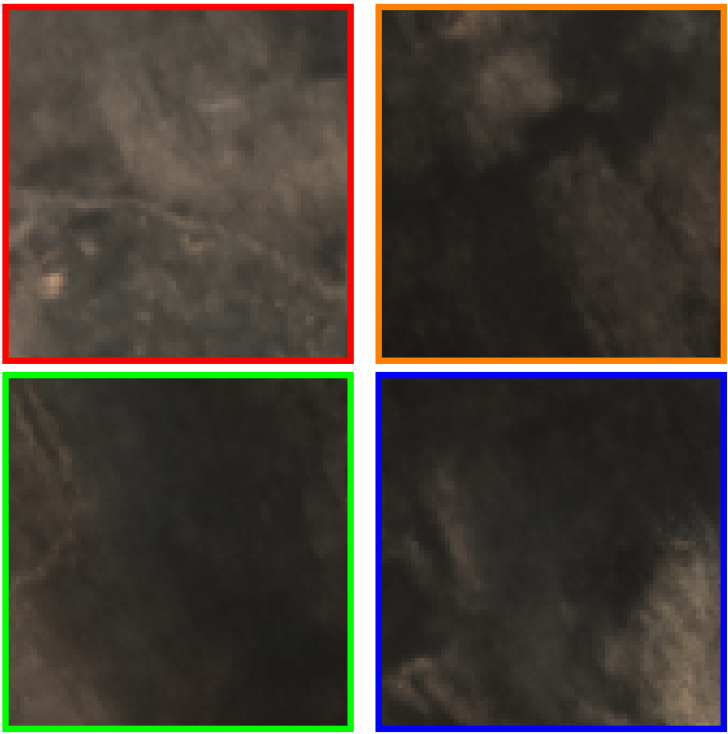}
            \end{minipage}%
        }
        \subfigure[GT]{
            \begin{minipage}[t]{0.205\linewidth}
                \centering
                    \begin{tikzpicture}
                    \node[anchor=south west,inner sep=0] (image) at (0,0) {
                \includegraphics[width=1\linewidth]{Nji_012_GT_val_3.png}};
                    \begin{scope}[x={(image.south east)},y={(image.north west)}]
                    \draw[green,thick] (0.0,0.357) rectangle (0.165,0.214);
                    \draw[orange,thick] (0.168,0.315) rectangle (0.336,0.173);
                    \draw[red,thick] (0.240,0.635) rectangle (0.410,0.782);
                    \draw[blue,thick] (0.730,0.000) rectangle (0.930,0.160);
                    \end{scope}
                    \end{tikzpicture}
            \end{minipage}%
            \begin{minipage}[t]{0.245\linewidth}
                \centering
                \includegraphics[width=1\linewidth]{Nji_012_detail_GT_val_3.png}
            \end{minipage}%
        }                
        \caption{\textbf{Training Strategies -- View Synthesis Visualisation.} Different pre-training strategies are tested on the \textit{very hard} scenario of Dji-A. Pre$_{no}$ generates a blurry image; Pre$_{sho}$, Pre$_{med}$ and Pre$_{lon}$ synthesised images that are close to GT, while Pre$_{med}$ shows advantages in restoring better details, particularly in the zoom-in view labeled in the orange rectangle. $\lambda=0$ recovers blurry image, $\lambda=\frac{1}{3}$ and $\lambda=\frac{50}{3}$ synthesised novel views with unified image tone, but zoom-in view shows blurrier detail than $\lambda=\frac{10}{3}$.}
        \label{ablationnovel}
    \end{center}
\end{figure*}

\begin{figure*}[!htbp]
    \begin{center}   
        \subfigure[Pre$_{no}$, $\lambda=\frac{10}{3}$]{
            \begin{minipage}[t]{0.223\linewidth}
                \centering
                    \begin{tikzpicture}
                    \node[anchor=south west,inner sep=0] (image) at (0,0) {
                \includegraphics[width=1\linewidth]{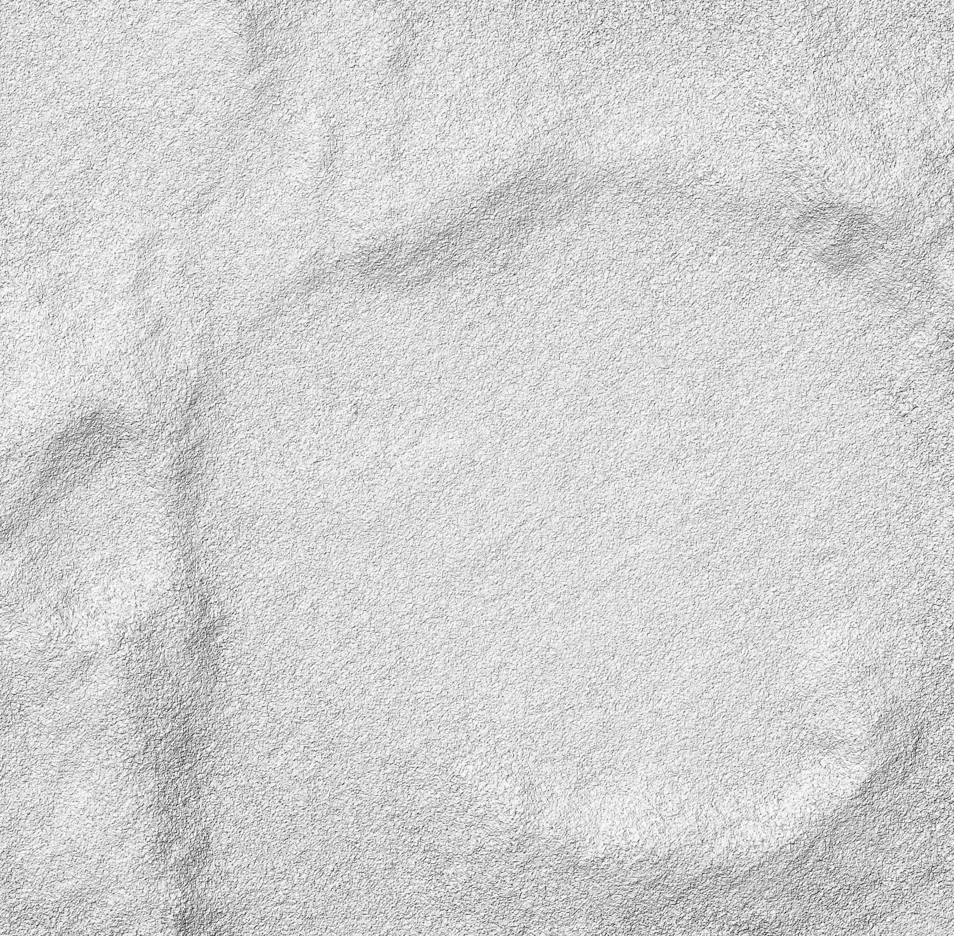}};
                    \begin{scope}[x={(image.south east)},y={(image.north west)}]
                    \draw[green,thick] (0.023,0.154) rectangle (0.186,0.323);
                    \draw[blue,thick] (0.840,0.140) rectangle (0.999,0.306);
                    \draw[orange,thick] (0.830,0.491) rectangle (0.992,0.655);
                    \draw[red,thick] (0.645,0.795) rectangle (0.802,0.628);
                    \end{scope}
                    \end{tikzpicture}
            \end{minipage}%
            \hspace{0mm}
            \begin{minipage}[t]{0.227\linewidth}
                \centering
                \includegraphics[width=1\linewidth]{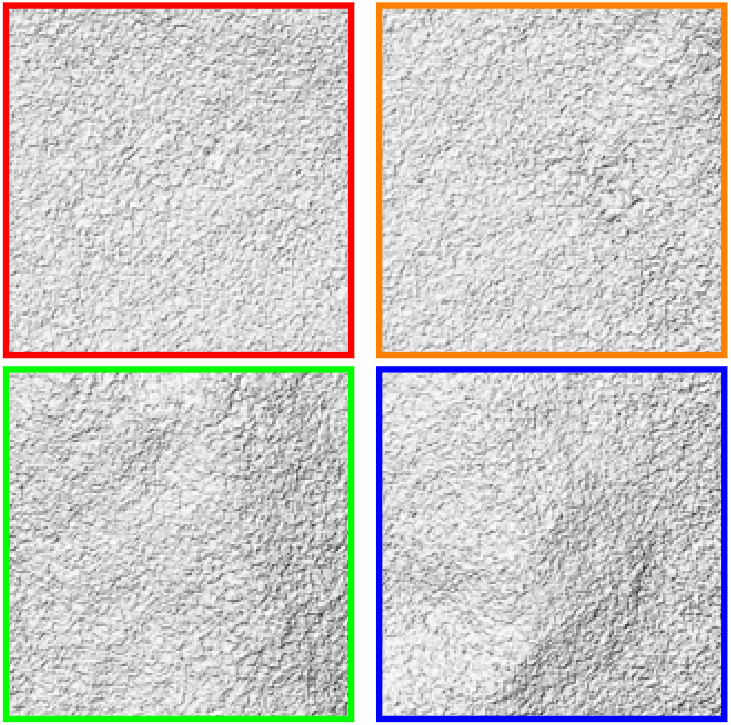}
            \end{minipage}%
            }
        \subfigure[Pre$_{sho}$, $\lambda=\frac{10}{3}$]{
            \begin{minipage}[t]{0.223\linewidth}
                \centering
                    \begin{tikzpicture}
                    \node[anchor=south west,inner sep=0] (image) at (0,0) {
                \includegraphics[width=1\linewidth]{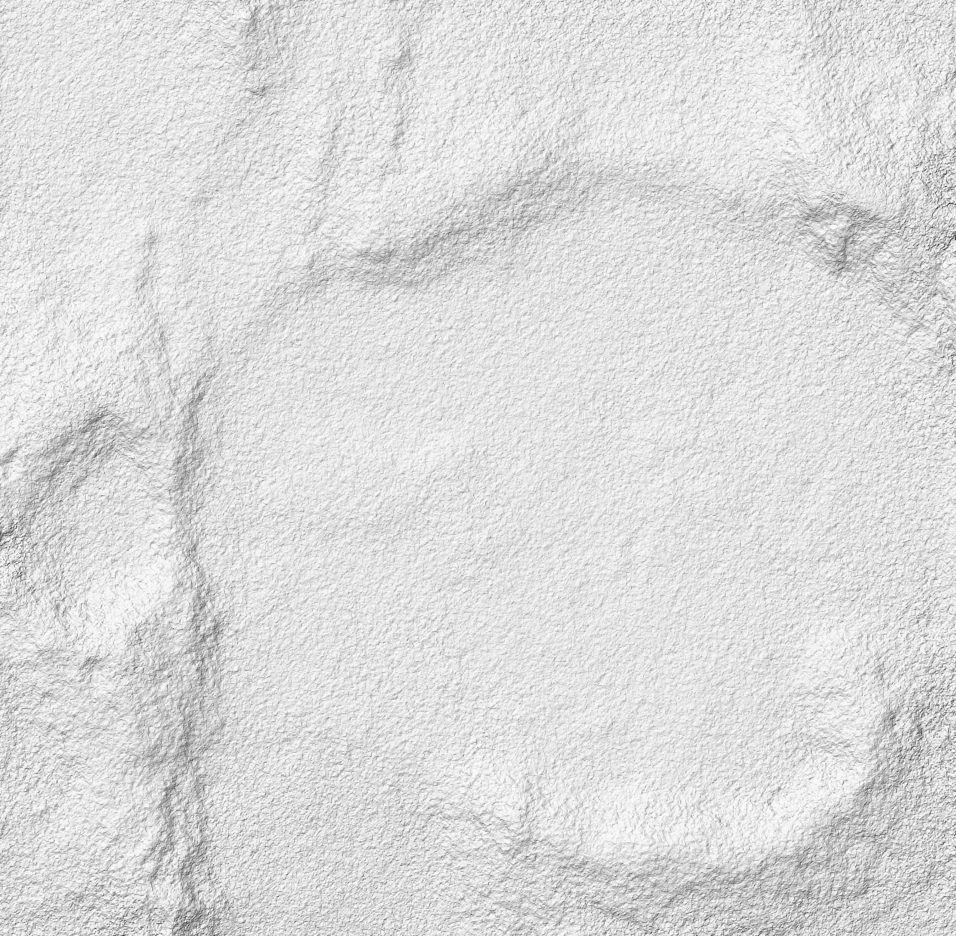}};
                    \begin{scope}[x={(image.south east)},y={(image.north west)}]
                    \draw[green,thick] (0.023,0.154) rectangle (0.186,0.323);
                    \draw[blue,thick] (0.840,0.140) rectangle (0.999,0.306);
                    \draw[orange,thick] (0.830,0.491) rectangle (0.992,0.655);
                    \draw[red,thick] (0.645,0.795) rectangle (0.802,0.628);
                    \end{scope}
                    \end{tikzpicture}
            \end{minipage}%
            \hspace{0mm}
            \begin{minipage}[t]{0.227\linewidth}
                \centering
                \includegraphics[width=1\linewidth]{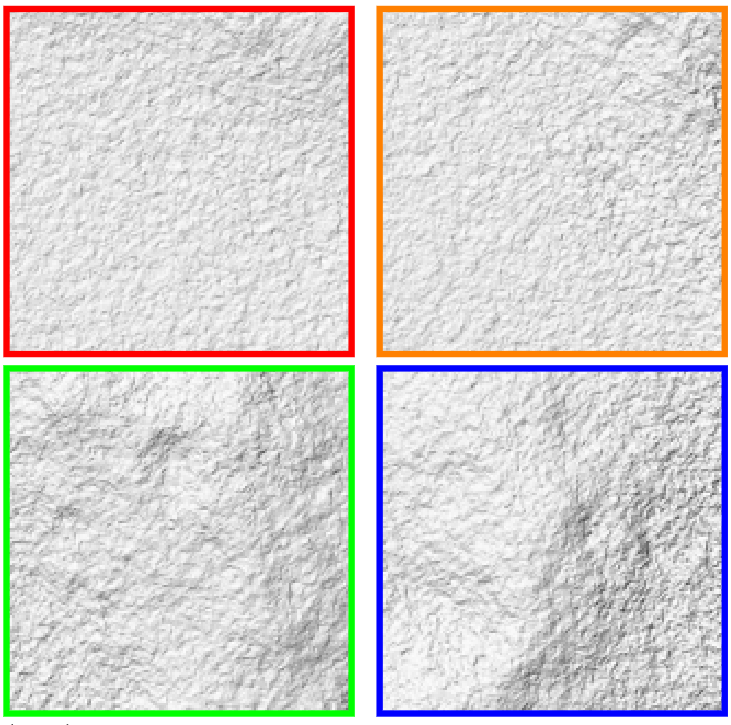}
            \end{minipage}%
            }    
        \subfigure[Pre$_{med}$, $\lambda=\frac{10}{3}$]{
            \begin{minipage}[t]{0.223\linewidth}
                \centering
                    \begin{tikzpicture}
                    \node[anchor=south west,inner sep=0] (image) at (0,0) {
                \includegraphics[width=1\linewidth]{Nji_012_SpS_RPV-ir0-lr2152_dsm.png}};
                    \begin{scope}[x={(image.south east)},y={(image.north west)}]
                    \draw[green,thick] (0.023,0.154) rectangle (0.186,0.323);
                    \draw[blue,thick] (0.840,0.140) rectangle (0.999,0.306);
                    \draw[orange,thick] (0.830,0.491) rectangle (0.992,0.655);
                    \draw[red,thick] (0.645,0.795) rectangle (0.802,0.628);
                    \end{scope}
                    \end{tikzpicture}
            \end{minipage}%
            \hspace{0mm}
            \begin{minipage}[t]{0.227\linewidth}
                \centering
                \includegraphics[width=1\linewidth]{Nji_012_detail_SpS_RPV-ir0-lr2152_dsm.png}
            \end{minipage}%
            }    
        \subfigure[Pre$_{lon}$, $\lambda=\frac{10}{3}$]{
            \begin{minipage}[t]{0.223\linewidth}
                \centering
                    \begin{tikzpicture}
                    \node[anchor=south west,inner sep=0] (image) at (0,0) {
                \includegraphics[width=1\linewidth]{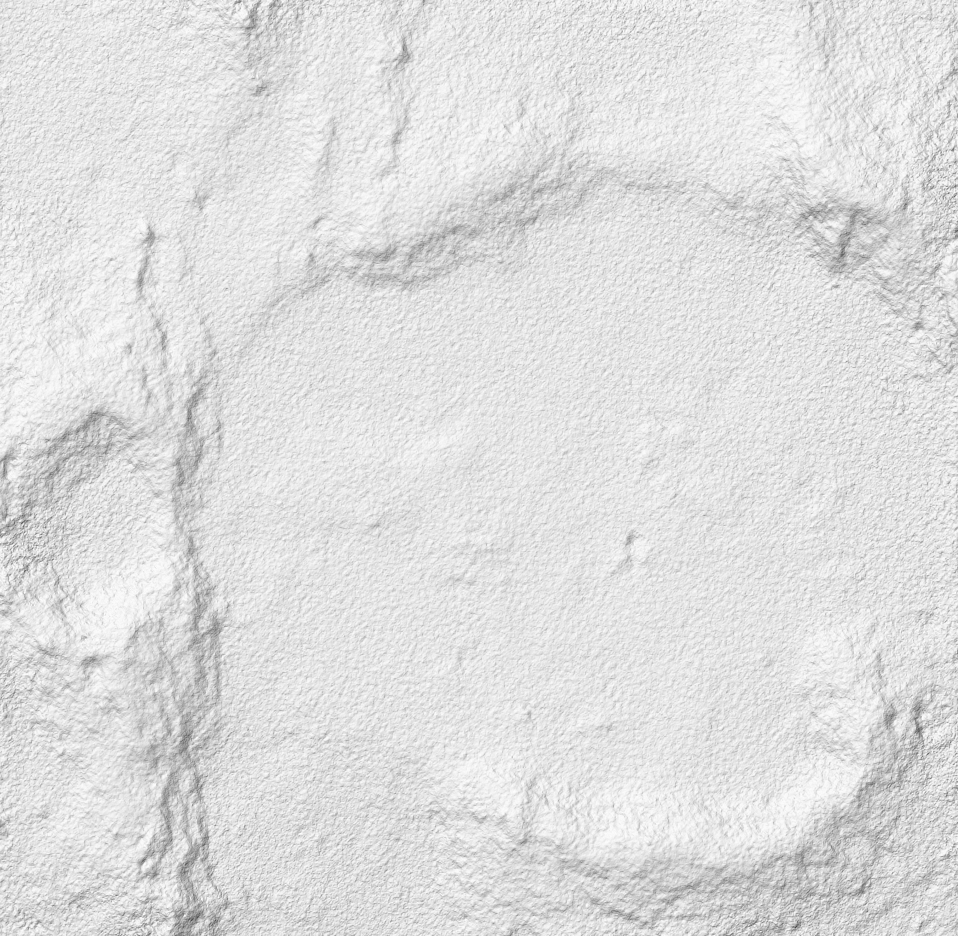}};
                    \begin{scope}[x={(image.south east)},y={(image.north west)}]
                    \draw[green,thick] (0.023,0.154) rectangle (0.186,0.323);
                    \draw[blue,thick] (0.840,0.140) rectangle (0.999,0.306);
                    \draw[orange,thick] (0.830,0.491) rectangle (0.992,0.655);
                    \draw[red,thick] (0.645,0.795) rectangle (0.802,0.628);
                    \end{scope}
                    \end{tikzpicture}
            \end{minipage}%
            \hspace{0mm}
            \begin{minipage}[t]{0.227\linewidth}
                \centering
                \includegraphics[width=1\linewidth]{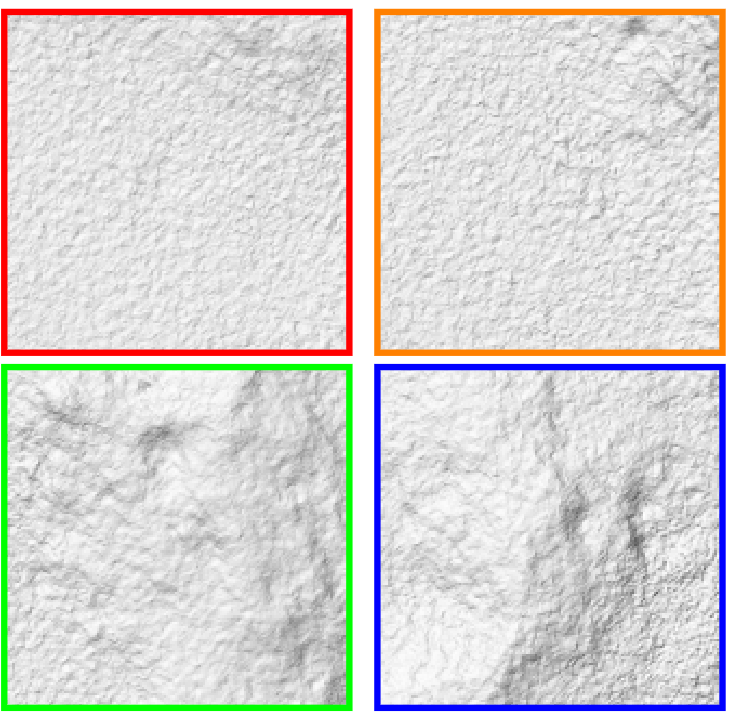}
            \end{minipage}%
            }    
        \subfigure[Pre$_{med}$, $\lambda=0$]{
            \begin{minipage}[t]{0.223\linewidth}
                \centering
                    \begin{tikzpicture}
                    \node[anchor=south west,inner sep=0] (image) at (0,0) {
                \includegraphics[width=1\linewidth]{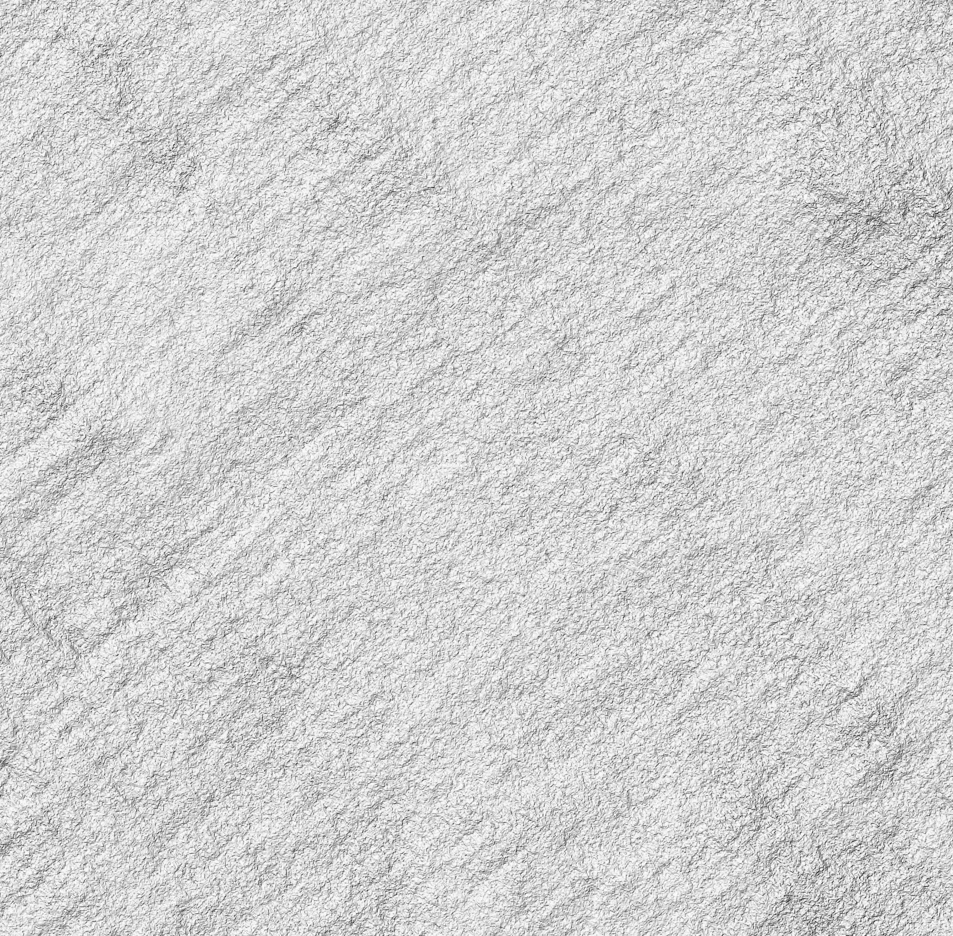}};
                    \begin{scope}[x={(image.south east)},y={(image.north west)}]
                    \draw[green,thick] (0.023,0.154) rectangle (0.186,0.323);
                    \draw[blue,thick] (0.840,0.140) rectangle (0.999,0.306);
                    \draw[orange,thick] (0.830,0.491) rectangle (0.992,0.655);
                    \draw[red,thick] (0.645,0.795) rectangle (0.802,0.628);
                    \end{scope}
                    \end{tikzpicture}
            \end{minipage}%
            \hspace{0mm}
            \begin{minipage}[t]{0.227\linewidth}
                \centering
                \includegraphics[width=1\linewidth]{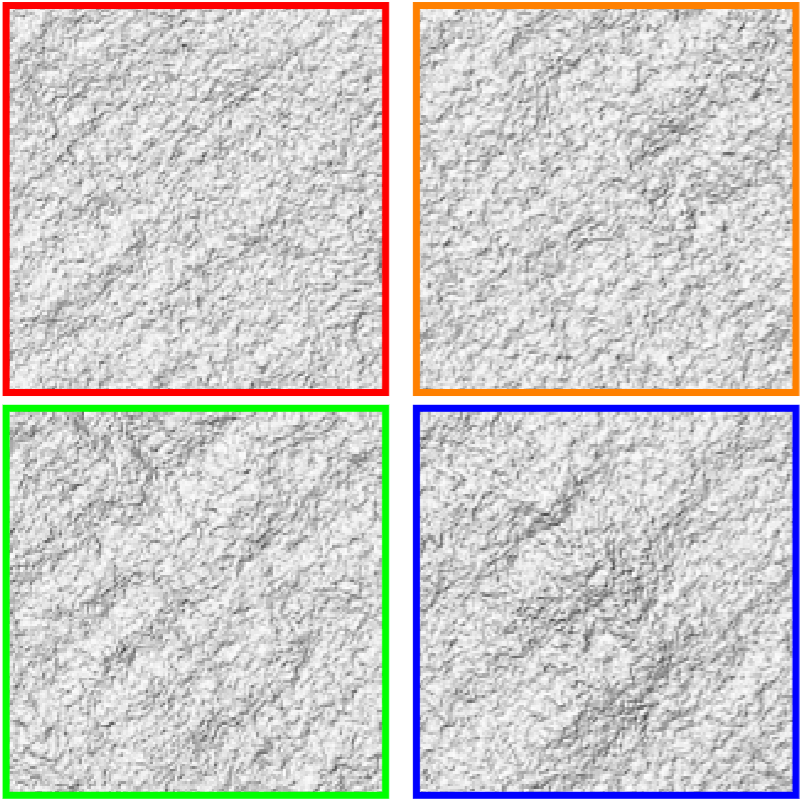}
            \end{minipage}%
            }
        \subfigure[Pre$_{med}$, $\lambda=\frac{1}{3}$]{
            \begin{minipage}[t]{0.223\linewidth}
                \centering
                    \begin{tikzpicture}
                    \node[anchor=south west,inner sep=0] (image) at (0,0) {
                \includegraphics[width=1\linewidth]{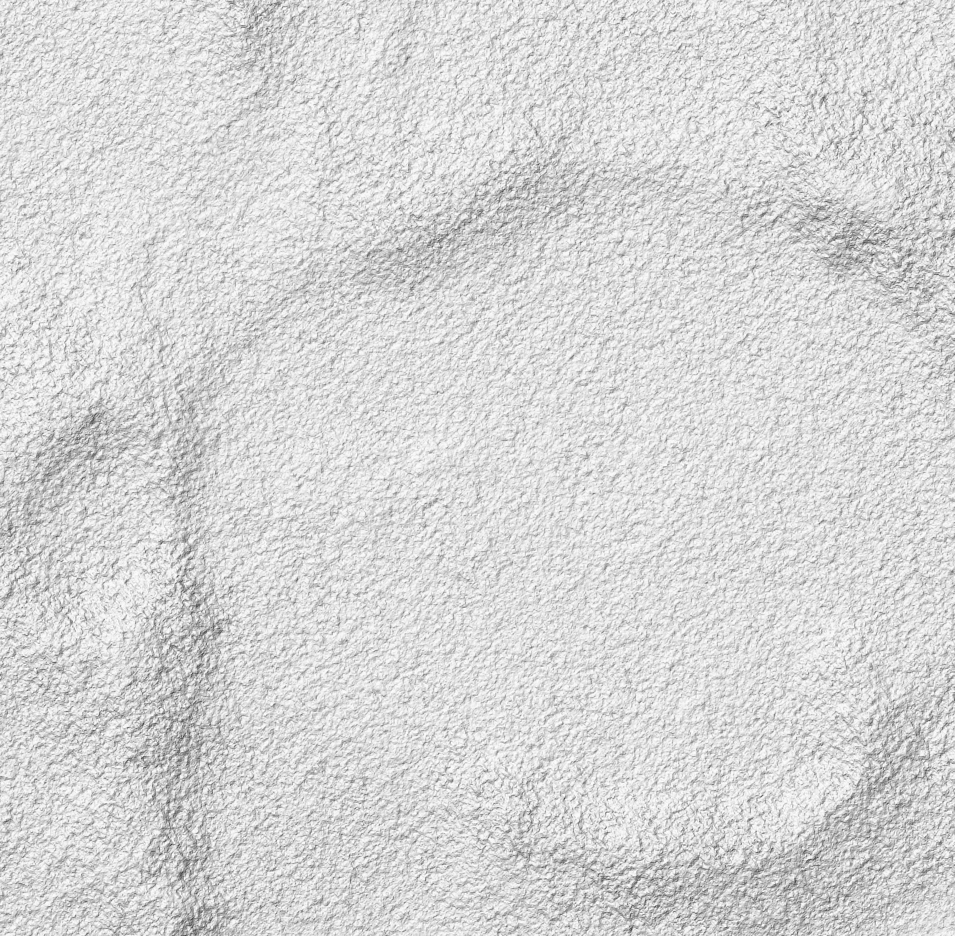}};
                    \begin{scope}[x={(image.south east)},y={(image.north west)}]
                    \draw[green,thick] (0.023,0.154) rectangle (0.186,0.323);
                    \draw[blue,thick] (0.840,0.140) rectangle (0.999,0.306);
                    \draw[orange,thick] (0.830,0.491) rectangle (0.992,0.655);
                    \draw[red,thick] (0.645,0.795) rectangle (0.802,0.628);
                    \end{scope}
                    \end{tikzpicture}
            \end{minipage}%
            \hspace{0mm}
            \begin{minipage}[t]{0.227\linewidth}
                \centering
                \includegraphics[width=1\linewidth]{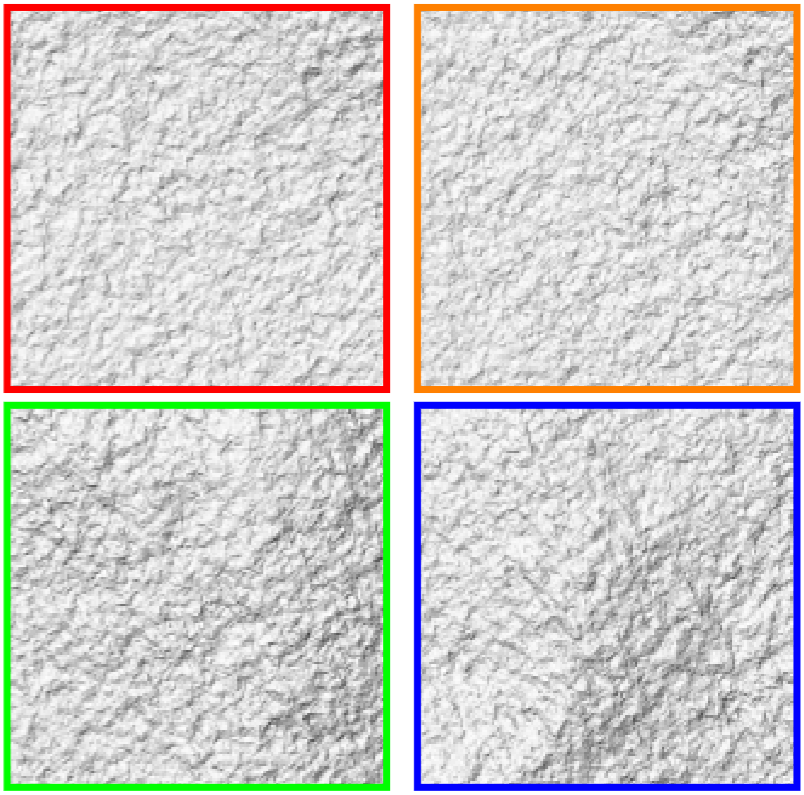}
            \end{minipage}%
            }
            \subfigure[Pre$_{med}$, $\lambda=\frac{50}{3}$]{
            \begin{minipage}[t]{0.223\linewidth}
                \centering
                    \begin{tikzpicture}
                    \node[anchor=south west,inner sep=0] (image) at (0,0) {
                \includegraphics[width=1\linewidth]{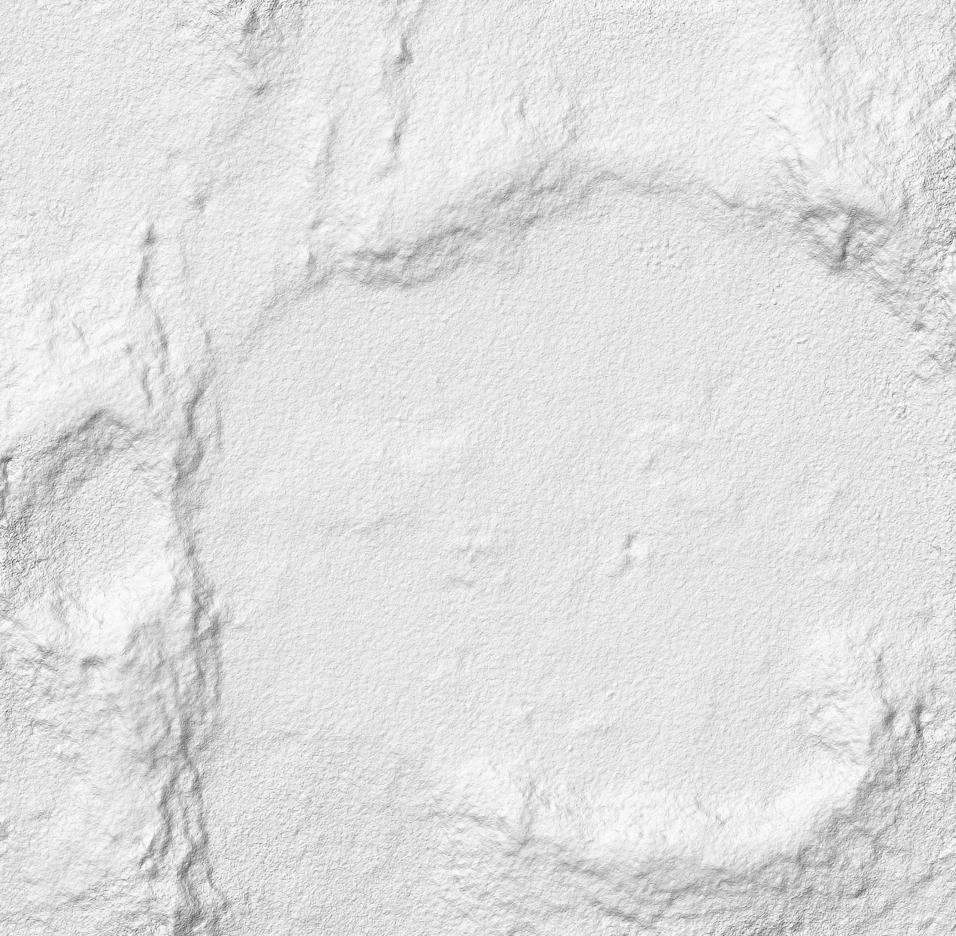}};
                    \begin{scope}[x={(image.south east)},y={(image.north west)}]
                    \draw[green,thick] (0.023,0.154) rectangle (0.186,0.323);
                    \draw[blue,thick] (0.840,0.140) rectangle (0.999,0.306);
                    \draw[orange,thick] (0.830,0.491) rectangle (0.992,0.655);
                    \draw[red,thick] (0.645,0.795) rectangle (0.802,0.628);
                    \end{scope}
                    \end{tikzpicture}
            \end{minipage}%
            \hspace{0mm}
            \begin{minipage}[t]{0.227\linewidth}
                \centering
                \includegraphics[width=1\linewidth]{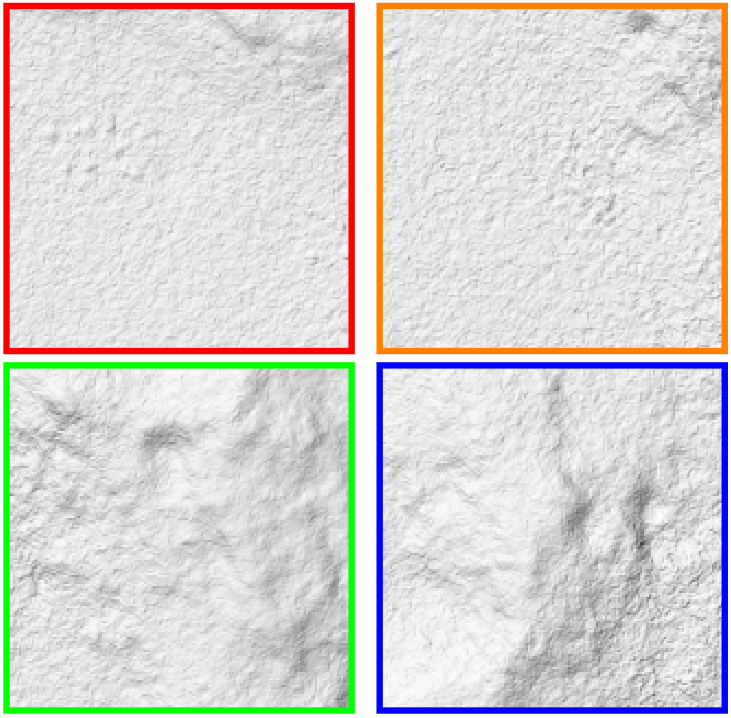}
            \end{minipage}%
            }
        \subfigure[GT]{
            \begin{minipage}[t]{0.223\linewidth}
                \centering
                    \begin{tikzpicture}
                    \node[anchor=south west,inner sep=0] (image) at (0,0) {
                \includegraphics[width=1\linewidth]{Nji_012_GT_dsm.png}};
                    \begin{scope}[x={(image.south east)},y={(image.north west)}]
                    \draw[green,thick] (0.023,0.154) rectangle (0.186,0.323);
                    \draw[blue,thick] (0.840,0.140) rectangle (0.999,0.306);
                    \draw[orange,thick] (0.830,0.491) rectangle (0.992,0.655);
                    \draw[red,thick] (0.645,0.795) rectangle (0.802,0.628);
                    \end{scope}
                    \end{tikzpicture}
            \end{minipage}%
            \hspace{0mm}
            \begin{minipage}[t]{0.227\linewidth}
                \centering
                \includegraphics[width=1\linewidth]{Nji_012_detail_GT_dsm.png}
            \end{minipage}%
            }
        \caption{\textbf{Training Strategies -- Altitude Estimation Visualisation} on Dji-A sites, using different training strategy for \OurNeRFShort. Pre$_{no}$'s surface is very noisy; Pre$_{sho}$, Pre$_{med}$ and Pre$_{lon}$ generate similar surfaces, among which Pre$_{sho}$ is the noisiest. $\lambda=0$ fails to recover topographic reliefs due to a lack of depth supervision. Performance is improved in $\lambda=\frac{1}{3}$, with noisy details. $\lambda=\frac{50}{3}$ estimates surface smoother than $\lambda=\frac{10}{3}$, thanks to the good quality of input depth.}
        \label{ablationalt}
    \end{center}
\end{figure*}

Qualitative results are presented in \Cref{ablationnovel,ablationalt}, while quantitative metrics are provided in \Cref{ablationmetrics}. The absence of pre-training leads to blurry synthetic images and noisy altitude estimations. Performance differences between Pre$_{sho}$, Pre$_{med}$, and Pre$_{lon}$ are minor, with Pre$_{med}$ emerging as the most optimal choice.




\paragraph{Depth Loss Weighting}
We perform an ablation experiment to evaluate the contribution of the depth loss term. Removing the term entirely results in very poor altitude predictions, confirming that a simple \NERF~architecture is unable to learn from just three views. By increasing the weight in the range $[\frac{1}{3},\frac{50}{3}]$, altitude metrics improve consistently (see \Cref{ablationalt}). The PSNR and SSIM metrics corresponding to the synthetic image quality reach a maximum at $\lambda=\frac{10}{3}$, suggesting that assigning greater importance to depths compromises the rendering quality. We set the $\lambda=\frac{10}{3}$ in all our experiments.



\begin{figure}[!htbp] 
    \begin{center}
        \subfigure[$Ren_{vol}$ novel view]{
            \begin{minipage}[t]{0.205\linewidth}
                \centering
                    \begin{tikzpicture}
                    \node[anchor=south west,inner sep=0] (image) at (0,0) {
                \includegraphics[width=1\linewidth]{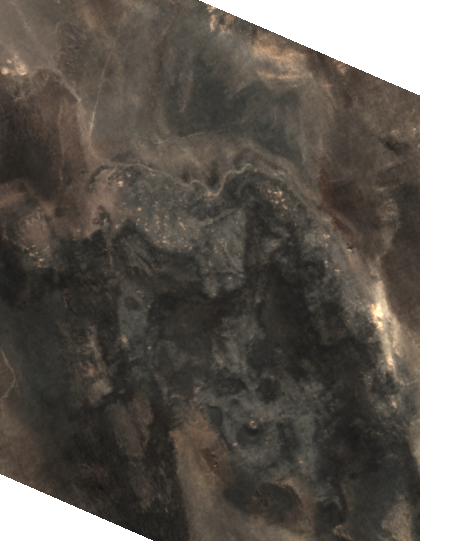}};
                    \begin{scope}[x={(image.south east)},y={(image.north west)}]
                    \draw[orange,thick] (0.521,0.953) rectangle (0.689,0.799);
                    \draw[green,thick] (0.168,0.315) rectangle (0.336,0.173);
                    \draw[red,thick] (0.0,0.94) rectangle (0.169,0.803);
                    \draw[blue,thick] (0.730,0.000) rectangle (0.930,0.160);
                    \end{scope}
                    \end{tikzpicture}
            \end{minipage}%
            \begin{minipage}[t]{0.245\linewidth}
                \centering
                \includegraphics[width=1\linewidth]{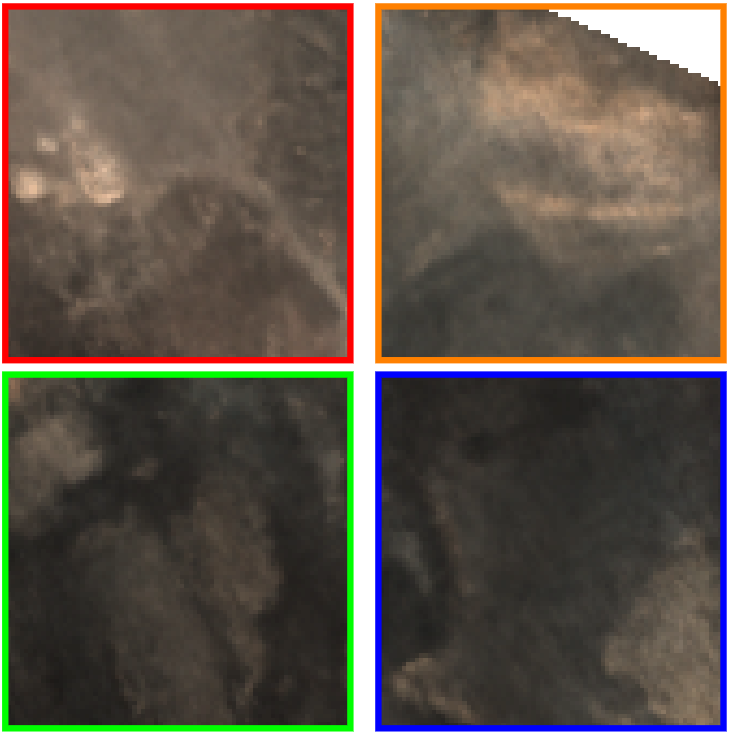}
            \end{minipage}%
        }
        \subfigure[$Ren_{vol}$ surface]{
            \begin{minipage}[t]{0.223\linewidth}
                \centering
                    \begin{tikzpicture}
                    \node[anchor=south west,inner sep=0] (image) at (0,0) {
                \includegraphics[width=1\linewidth]{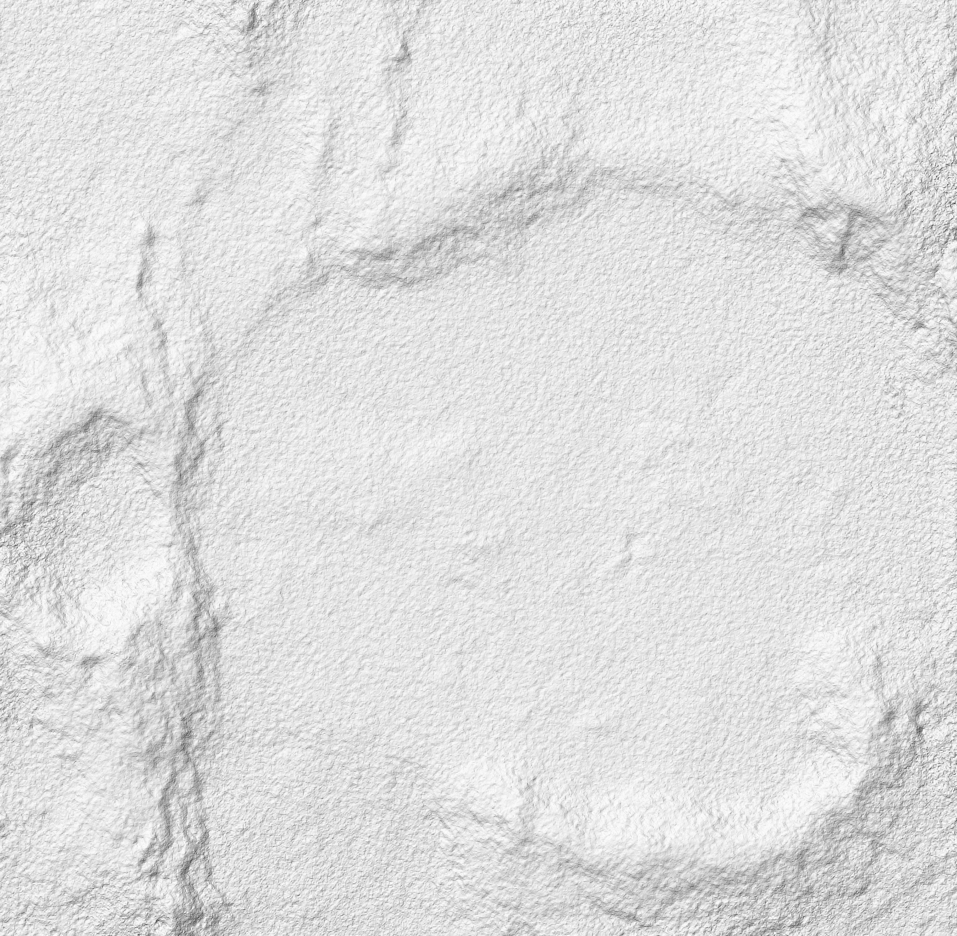}};
                    \begin{scope}[x={(image.south east)},y={(image.north west)}]
                    \draw[green,thick] (0.023,0.154) rectangle (0.186,0.323);
                    \draw[blue,thick] (0.840,0.140) rectangle (0.999,0.306);
                    \draw[orange,thick] (0.830,0.491) rectangle (0.992,0.655);
                    \draw[red,thick] (0.645,0.795) rectangle (0.802,0.628);
                    \end{scope}
                    \end{tikzpicture}
            \end{minipage}%
            \hspace{0mm}
            \begin{minipage}[t]{0.227\linewidth}
                \centering
                \includegraphics[width=1\linewidth]{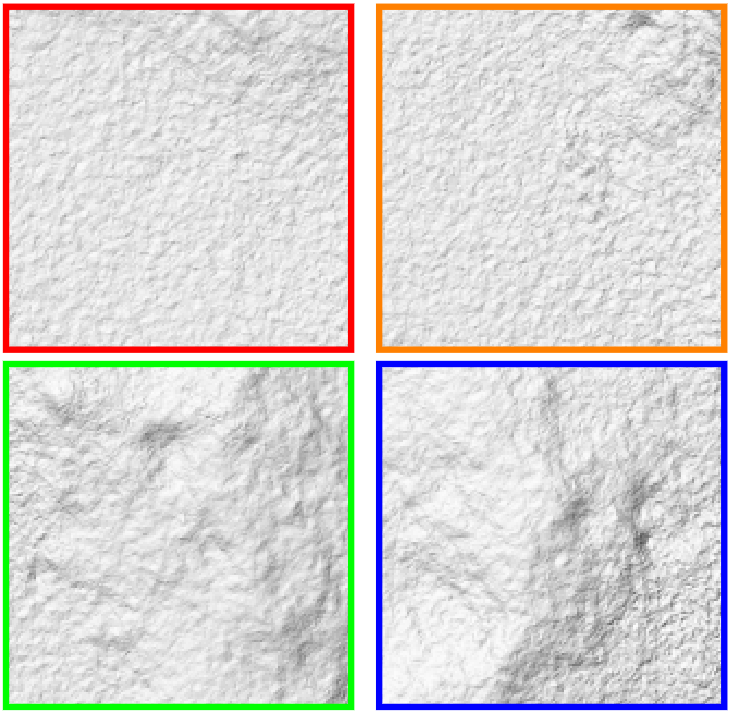}
            \end{minipage}%
        }                    
        \subfigure[$Ren_{sur}$ novel view]{
            \begin{minipage}[t]{0.205\linewidth}
                \centering
                    \begin{tikzpicture}
                    \node[anchor=south west,inner sep=0] (image) at (0,0) {
                \includegraphics[width=1\linewidth]{Nji_012_SpS_RPV-ir0-lr2152_val_3.png}};
                    \begin{scope}[x={(image.south east)},y={(image.north west)}]
                    \draw[orange,thick] (0.521,0.953) rectangle (0.689,0.799);
                    \draw[green,thick] (0.168,0.315) rectangle (0.336,0.173);
                    \draw[red,thick] (0.0,0.94) rectangle (0.169,0.803);
                    \draw[blue,thick] (0.730,0.000) rectangle (0.930,0.160);
                    \end{scope}
                    \end{tikzpicture}
            \end{minipage}%
            \begin{minipage}[t]{0.245\linewidth}
                \centering
                \includegraphics[width=1\linewidth]{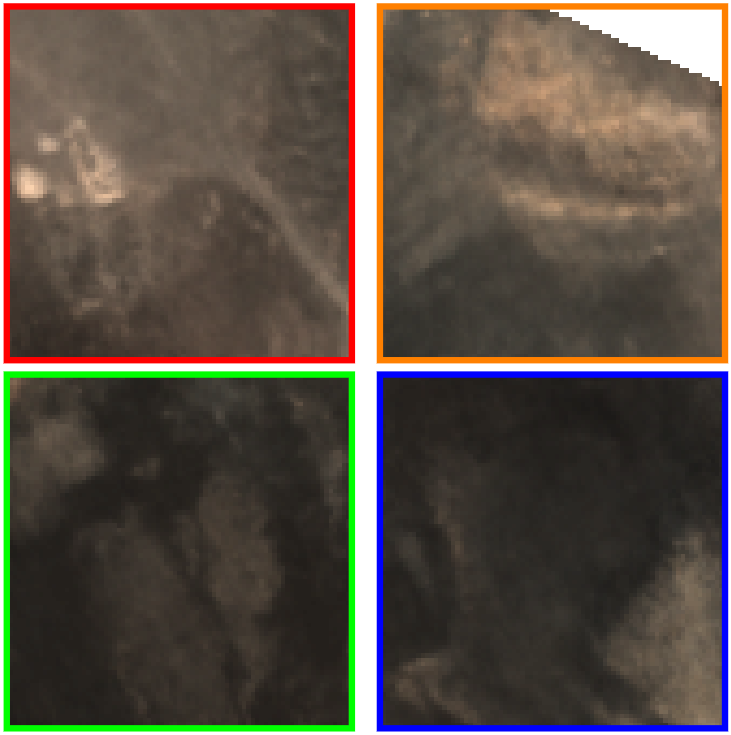}
            \end{minipage}%
        }
        \subfigure[$Ren_{sur}$ surface]{
            \begin{minipage}[t]{0.223\linewidth}
                \centering
                    \begin{tikzpicture}
                    \node[anchor=south west,inner sep=0] (image) at (0,0) {
                \includegraphics[width=1\linewidth]{Nji_012_SpS_RPV-ir0-lr2152_dsm.png}};
                    \begin{scope}[x={(image.south east)},y={(image.north west)}]
                    \draw[green,thick] (0.023,0.154) rectangle (0.186,0.323);
                    \draw[blue,thick] (0.840,0.140) rectangle (0.999,0.306);
                    \draw[orange,thick] (0.830,0.491) rectangle (0.992,0.655);
                    \draw[red,thick] (0.645,0.795) rectangle (0.802,0.628);
                    \end{scope}
                    \end{tikzpicture}
            \end{minipage}%
            \hspace{0mm}
            \begin{minipage}[t]{0.227\linewidth}
                \centering
                \includegraphics[width=1\linewidth]{Nji_012_detail_SpS_RPV-ir0-lr2152_dsm.png}
            \end{minipage}%
            }                    
        \subfigure[GT novel view]{
            \begin{minipage}[t]{0.205\linewidth}
                \centering
                    \begin{tikzpicture}
                    \node[anchor=south west,inner sep=0] (image) at (0,0) {
                \includegraphics[width=1\linewidth]{Nji_012_GT_val_3.png}};
                    \begin{scope}[x={(image.south east)},y={(image.north west)}]
                    \draw[orange,thick] (0.521,0.953) rectangle (0.689,0.799);
                    \draw[green,thick] (0.168,0.315) rectangle (0.336,0.173);
                    \draw[red,thick] (0.0,0.94) rectangle (0.169,0.803);
                    \draw[blue,thick] (0.730,0.000) rectangle (0.930,0.160);
                    \end{scope}
                    \end{tikzpicture}
            \end{minipage}%
            \begin{minipage}[t]{0.245\linewidth}
                \centering
                \includegraphics[width=1\linewidth]{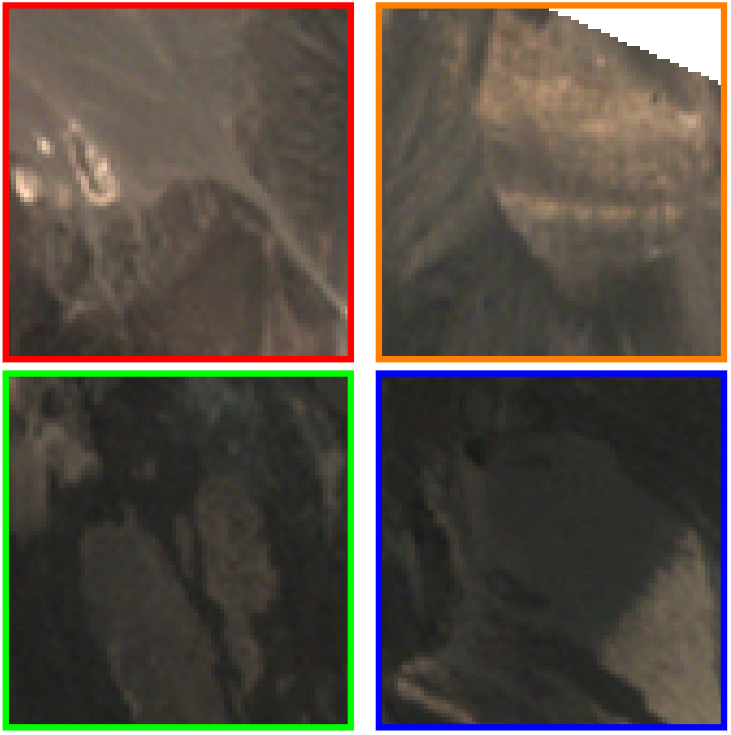}
            \end{minipage}%
        }        
        \subfigure[GT surface]{
            \begin{minipage}[t]{0.223\linewidth}
                \centering
                    \begin{tikzpicture}
                    \node[anchor=south west,inner sep=0] (image) at (0,0) {
                \includegraphics[width=1\linewidth]{Nji_012_GT_dsm.png}};
                    \begin{scope}[x={(image.south east)},y={(image.north west)}]
                    \draw[green,thick] (0.023,0.154) rectangle (0.186,0.323);
                    \draw[blue,thick] (0.840,0.140) rectangle (0.999,0.306);
                    \draw[orange,thick] (0.830,0.491) rectangle (0.992,0.655);
                    \draw[red,thick] (0.645,0.795) rectangle (0.802,0.628);
                    \end{scope}
                    \end{tikzpicture}
            \end{minipage}%
            \hspace{0mm}
            \begin{minipage}[t]{0.227\linewidth}
                \centering
                \includegraphics[width=1\linewidth]{Nji_012_detail_GT_dsm.png}
            \end{minipage}%
            }                    
        \caption{\textbf{Rendering -- Visualisations.} $Ren_{vol}$ rendered novel view with similar quality to $Ren_{sur}$, while zoom-in shows blurrier details. The surface qualities of $Ren_{vol}$ and $Ren_{sur}$ are similar with no obvious differences.}
        \label{survsvolfig}
    \end{center}
\end{figure}

\paragraph{Surface or volume rendering}\label{sub:rendering}
The rendering equation in \Cref{renderingequation} can be applied as \textit{surface rendering} ($Ren_{sur}$) or \textit{volume rendering} ($Ren_{vol}$). In surface rendering, the RPV parameters \textbf{n}, $\bm{\rho_0}$, $\textbf{k}$, $\bm{\Theta}$ and $\bm{\rho_c}$ are estimated at the surface by accumulating $N$ points along the ray and applying the rendering once for each ray. In {volume rendering}, rendering is applied to each sample individually, associating it with a color \textbf{c}, and accumulation is done on the sample colours instead of RPV parameters. 
{$Ren_{sur}$ is more rigorous than $Ren_{vol}$, as the latter assumes every point along the ray follow the RPV reflectance, while $Ren_{sur}$ assumes the same only for points on the surface which is concordant wit the RPV model definition.} 
We demonstrate in \Cref{survsvoltab} and \Cref{survsvolfig} that $Ren_{sur}$ outperforms $Ren_{vol}$ and adopt this rendering method in our experiments.


\newcommand{\gtspsnr}[1]{\gradientcelld{#1}{24}{29}{34}{gLow}{gMid}{gHigh}{70}}
\newcommand{\gtsssim}[1]{\gradientcelld{#1}{0.8}{0.89}{0.98}{gLow}{gMid}{gHigh}{70}}
\newcommand{\gtsgeom}[1]{\gradientcelld{#1}{1.0}{3.0}{5.0}{gHigh}{gMid}{gLow}{70}}

\begin{table}[htbp]
\centering
\begin{tabular}{|l|c|c|c|c|c|c|c|c|}
\hline
\multirow{2}*{\textit{Pre}} & \multirow{2}*{$\lambda$} & \multirow{2}*{MAE $\downarrow$} & \multicolumn{3}{|c}{PSNR $\uparrow$} & \multicolumn{3}{|c|}{SSIM $\uparrow$} \\ \cline{4-9}
& & & Easy & Hard & VHard & Easy & Hard & VHard \\ \hline \hline
\textit{no}&\multirow{4}*{$\frac{10}{3}$}&\gtsgeom{1.632}&\gtspsnr{38.057}&\gtspsnr{33.446}&\gtspsnr{31.186}&\gtsssim{0.966}&\gtsssim{0.93}&\gtsssim{0.884}\\\cline{1-1}\cline{3-9}
\textit{sho}&&\gtsgeom{1.449}&\gtspsnr{41.166}&\gtspsnr{35.36}&\gtspsnr{32.999}&\gtsssim{0.983}&\textcolor{magenta}{\gtsssim{0.951}}&\gtsssim{0.913}\\\cline{1-1}\cline{3-9}
\textit{med}&&\textcolor{magenta}{\gtsgeom{1.378}}&\textcolor{blue}{\gtspsnr{41.844}}&\textcolor{blue}{\gtspsnr{36.232}}&\textcolor{blue}{\gtspsnr{33.35}}&\textcolor{blue}{\gtsssim{0.985}}&\textcolor{blue}{\gtsssim{0.957}}&\textcolor{blue}{\gtsssim{0.918}}\\\cline{1-1}\cline{3-9}

\textit{lon}&&\gtsgeom{1.39}&\textcolor{magenta}{\gtspsnr{41.415}}&\gtspsnr{35.645}&\gtspsnr{31.663}&\textcolor{magenta}{\gtsssim{0.984}}&\gtsssim{0.947}&\gtsssim{0.88}\\\cline{1-9}
\multirow{3}*{\textit{med}}&$\lambda=0$&\gtsgeom{9.432}&\gtspsnr{39.408}&\gtspsnr{34.755}&\gtspsnr{31.870}&\gtsssim{0.973}&\gtsssim{0.941}&\gtsssim{0.900}\\\cline{2-9}
&$\lambda=\frac{1}{3}$&\gtsgeom{1.877}&\gtspsnr{40.894}&\textcolor{magenta}{\gtspsnr{35.822}}&\textcolor{magenta}{\gtspsnr{33.318}}&\gtsssim{0.982}&\textcolor{magenta}{\gtsssim{0.951}}&\textcolor{magenta}{\gtsssim{0.914}}\\\cline{2-9}
&$\lambda=\frac{50}{3}$&\textcolor{blue}{\gtsgeom{1.353}}&\gtspsnr{41.109}&\gtspsnr{34.915}&\gtspsnr{32.107}&\gtsssim{0.983}&\gtsssim{0.949}&\gtsssim{0.908}\\\cline{1-9}
\end{tabular}
\caption{\textbf{Training Strategies -- Quantitative Evaluation.} \textit{Pre} refers to the various training settings shown in \Cref{pre-trainstrategy}, while $\lambda$ is the parameter that balances the contribution of colour and depth losses in \Cref{loss}. The best and second best performing metrics are in \textcolor{blue}{blue} and \textcolor{magenta}{magenta}. Pre$_{med}$ achieved the best PSNR and SSIM and the second best MAE. $\lambda=\frac{50}{3}$ ranks the best for MAE, but has worse PSNR and SSIM than Pre$_{med}$. Tests correspond to Dji-A dataset.}
\label{ablationmetrics}
\end{table}

\newcommand{\grendpsnr}[1]{\gradientcelld{#1}{24}{29}{34}{gLow}{gMid}{gHigh}{70}}
\newcommand{\grendssim}[1]{\gradientcelld{#1}{0.8}{0.89}{0.98}{gLow}{gMid}{gHigh}{70}}
\newcommand{\grendgeom}[1]{\gradientcelld{#1}{1.370}{1.380}{1.500}{gHigh}{gMid}{gLow}{70}}

\begin{table}[htbp!]
\centering
\begin{tabular}{|l|c|c|c|c|c|c|c|}
\hline
\multirow{2}*{}&\multirow{2}*{MAE$\downarrow$}&\multicolumn{3}{|c}{PSNR$\uparrow$}&\multicolumn{3}{|c|}{SSIM$\uparrow$}\\\cline{3-8}
&&Easy&Hard&VHard&Easy&Hard&VHard\\\hline\hline
$Ren_{vol}$&\grendgeom{1.399}&\grendpsnr{41.743}&\grendpsnr{35.867}&\grendpsnr{31.770}&\textbf{\grendssim{0.985}}&\grendssim{0.955}&\grendssim{0.903}\\\hline
$Ren_{sur}$&\textbf{\grendgeom{1.378}}&\textbf{\grendpsnr{41.844}}&\textbf{\grendpsnr{36.232}}&\textbf{\grendpsnr{33.350}}&\textbf{\grendssim{0.985}}&\textbf{\grendssim{0.957}}&\textbf{\grendssim{0.918}}\\\hline
\end{tabular}
\caption{\textbf{Rendering -- Quantitative Evaluation of $Ren_{vol}$ and $Ren_{sur}$}. $Ren_{vol}$ produces new views and surfaces close to $Ren_{sur}$ with slightly poorer metrics overall. Tests correspond to Dji-A dataset.}
\label{survsvoltab}
\end{table}





\newpage
\section{Conclusion}
We presented \OurNeRFShort, an extension of \NERF~adapted to sparse satellite imagery, capable of estimating realistic BRDFs for natural surfaces. By incorporating the semi-empirical Rahman-Pinty-Verstraete (RPV) model, \OurNeRFShort~enhances the rendering of anisotropic surface reflectance, leading to improved quality of both synthetic images and recovered surface altitudes. Although our experiments show promising results, certain limitations remain. At present, our method does not explicitly model shading effects, and although our surface reconstructions outperform other \NERF-based approaches, they still lack the regularity achieved by \acc{SGM}. Future work will address these challenges.

\section{Acknowledgements}
This research was funded by CNES (Centre national d’études spatiales) through a PostDoc scholarship as well as CAROLInA/SURFACEs projects. The Djibouti dataset was obtained through the CNES ISIS framework. Numerical computations were carried out on the SCAPAD cluster at the Institut de Physique du Globe de Paris. We thank Arthur Delorme for the Lanzhou dataset. We also thank Manchun Lei and Antoine Lucas for fruitful discussions.






\bibliographystyle{elsarticle-harv}
\footnotesize
\bibliography{BRDFNeRF}

\begin{thebibliography}{64}
\expandafter\ifx\csname natexlab\endcsname\relax\def\natexlab#1{#1}\fi
\providecommand{\url}[1]{\texttt{#1}}
\providecommand{\href}[2]{#2}
\providecommand{\path}[1]{#1}
\providecommand{\DOIprefix}{doi:}
\providecommand{\ArXivprefix}{arXiv:}
\providecommand{\URLprefix}{URL: }
\providecommand{\Pubmedprefix}{pmid:}
\providecommand{\doi}[1]{\href{http://dx.doi.org/#1}{\path{#1}}}
\providecommand{\Pubmed}[1]{\href{pmid:#1}{\path{#1}}}
\providecommand{\bibinfo}[2]{#2}
\ifx\xfnm\relax \def\xfnm[#1]{\unskip,\space#1}\fi
\bibitem[{Behari et~al.(2024)Behari, Dave, Tiwary, Yang and
  Raskar}]{behari2024sundial}
\bibinfo{author}{Behari, N.}, \bibinfo{author}{Dave, A.},
  \bibinfo{author}{Tiwary, K.}, \bibinfo{author}{Yang, W.},
  \bibinfo{author}{Raskar, R.}, \bibinfo{year}{2024}.
\newblock \bibinfo{title}{Sundial: 3d satellite understanding through direct
  ambient and complex lighting decomposition}, in: \bibinfo{booktitle}{CVPR},
  pp. \bibinfo{pages}{522--532}.
\bibitem[{Bi et~al.(2020)Bi, Xu, Srinivasan, Mildenhall, Sunkavalli, Hasan,
  Hold-Geoffroy, Kriegman and Ramamoorthi}]{bi2020neural}
\bibinfo{author}{Bi, S.}, \bibinfo{author}{Xu, Z.},
  \bibinfo{author}{Srinivasan, P.}, \bibinfo{author}{Mildenhall, B.},
  \bibinfo{author}{Sunkavalli, K.}, \bibinfo{author}{Hasan, M.},
  \bibinfo{author}{Hold-Geoffroy, Y.}, \bibinfo{author}{Kriegman, D.},
  \bibinfo{author}{Ramamoorthi, R.}, \bibinfo{year}{2020}.
\newblock \bibinfo{title}{Neural reflectance fields for appearance
  acquisition}.
\bibitem[{Biliouris et~al.(2009)Biliouris, Van~der Zande, Verstraeten,
  Stuckens, Muys, Dutr{\'e} and Coppin}]{biliouris2009rpv}
\bibinfo{author}{Biliouris, D.}, \bibinfo{author}{Van~der Zande, D.},
  \bibinfo{author}{Verstraeten, W.W.}, \bibinfo{author}{Stuckens, J.},
  \bibinfo{author}{Muys, B.}, \bibinfo{author}{Dutr{\'e}, P.},
  \bibinfo{author}{Coppin, P.}, \bibinfo{year}{2009}.
\newblock \bibinfo{title}{Rpv model parameters based on hyperspectral
  bidirectional reflectance measurements of fagus sylvatica l. leaves}.
\newblock \bibinfo{journal}{Remote Sensing} \bibinfo{volume}{1},
  \bibinfo{pages}{92--106}.
\bibitem[{Billouard et~al.(2024)Billouard, Derksen, Sarrazin and
  Vallet}]{billouard2024sat}
\bibinfo{author}{Billouard, C.}, \bibinfo{author}{Derksen, D.},
  \bibinfo{author}{Sarrazin, E.}, \bibinfo{author}{Vallet, B.},
  \bibinfo{year}{2024}.
\newblock \bibinfo{title}{Sat-ngp: Unleashing neural graphics primitives for
  fast relightable transient-free 3d reconstruction from satellite imagery}.
\newblock \bibinfo{journal}{arXiv preprint arXiv:2403.18711} .
\bibitem[{Bosch et~al.(2019)Bosch, Foster, Christie, Wang, Hager and
  Brown}]{bosch2019semantic}
\bibinfo{author}{Bosch, M.}, \bibinfo{author}{Foster, K.},
  \bibinfo{author}{Christie, G.}, \bibinfo{author}{Wang, S.},
  \bibinfo{author}{Hager, G.D.}, \bibinfo{author}{Brown, M.},
  \bibinfo{year}{2019}.
\newblock \bibinfo{title}{Semantic stereo for incidental satellite images}, in:
  \bibinfo{booktitle}{2019 IEEE Winter Conference on Applications of Computer
  Vision (WACV)}, \bibinfo{organization}{IEEE}. pp.
  \bibinfo{pages}{1524--1532}.
\bibitem[{Boss et~al.(2021)Boss, Braun, Jampani, Barron, Liu and
  Lensch}]{boss2021nerd}
\bibinfo{author}{Boss, M.}, \bibinfo{author}{Braun, R.},
  \bibinfo{author}{Jampani, V.}, \bibinfo{author}{Barron, J.T.},
  \bibinfo{author}{Liu, C.}, \bibinfo{author}{Lensch, H.P.A.},
  \bibinfo{year}{2021}.
\newblock \bibinfo{title}{Nerd: Neural reflectance decomposition from image
  collections}.
\bibitem[{Chang and Chen(2018)}]{PSMNet}
\bibinfo{author}{Chang, J.R.}, \bibinfo{author}{Chen, Y.S.},
  \bibinfo{year}{2018}.
\newblock \bibinfo{title}{Pyramid stereo matching network}, in:
  \bibinfo{booktitle}{CVPR}.
\bibitem[{Chebbi et~al.(2023)Chebbi, Rupnik, Pierrot-Deseilligny and
  Lopes}]{chebbi2023deepsim}
\bibinfo{author}{Chebbi, M.A.}, \bibinfo{author}{Rupnik, E.},
  \bibinfo{author}{Pierrot-Deseilligny, M.}, \bibinfo{author}{Lopes, P.},
  \bibinfo{year}{2023}.
\newblock \bibinfo{title}{Deepsim-nets: Deep similarity networks for stereo
  image matching}, in: \bibinfo{booktitle}{CVPR}, pp.
  \bibinfo{pages}{2097--2105}.
\bibitem[{Combes et~al.(2007)Combes, Bousquet, Jacquemoud, Sinoquet,
  Varlet-Grancher and Moya}]{combes2007new}
\bibinfo{author}{Combes, D.}, \bibinfo{author}{Bousquet, L.},
  \bibinfo{author}{Jacquemoud, S.}, \bibinfo{author}{Sinoquet, H.},
  \bibinfo{author}{Varlet-Grancher, C.}, \bibinfo{author}{Moya, I.},
  \bibinfo{year}{2007}.
\newblock \bibinfo{title}{A new spectrogoniophotometer to measure leaf spectral
  and directional optical properties}.
\newblock \bibinfo{journal}{Remote Sensing of Environment}
  \bibinfo{volume}{109}, \bibinfo{pages}{107--117}.
\bibitem[{Deng et~al.(2022)Deng, Liu, Zhu and Ramanan}]{deng2022depth}
\bibinfo{author}{Deng, K.}, \bibinfo{author}{Liu, A.}, \bibinfo{author}{Zhu,
  J.Y.}, \bibinfo{author}{Ramanan, D.}, \bibinfo{year}{2022}.
\newblock \bibinfo{title}{Depth-supervised nerf: Fewer views and faster
  training for free}, in: \bibinfo{booktitle}{CVPR}, pp.
  \bibinfo{pages}{12882--12891}.
\bibitem[{Derksen and Izzo(2021)}]{derksen2021shadow}
\bibinfo{author}{Derksen, D.}, \bibinfo{author}{Izzo, D.},
  \bibinfo{year}{2021}.
\newblock \bibinfo{title}{Shadow neural radiance fields for multi-view
  satellite photogrammetry}, in: \bibinfo{booktitle}{CVPRW}, pp.
  \bibinfo{pages}{1152--1161}.
\bibitem[{Dumont et~al.(2010)Dumont, Brissaud, Picard, Schmitt, Gallet and
  Arnaud}]{dumont2010high}
\bibinfo{author}{Dumont, M.}, \bibinfo{author}{Brissaud, O.},
  \bibinfo{author}{Picard, G.}, \bibinfo{author}{Schmitt, B.},
  \bibinfo{author}{Gallet, J.C.}, \bibinfo{author}{Arnaud, Y.},
  \bibinfo{year}{2010}.
\newblock \bibinfo{title}{High-accuracy measurements of snow bidirectional
  reflectance distribution function at visible and nir wavelengths--comparison
  with modelling results}.
\newblock \bibinfo{journal}{Atmospheric Chemistry and Physics}
  \bibinfo{volume}{10}, \bibinfo{pages}{2507--2520}.
\bibitem[{Ehret et~al.(2023)Ehret, Marí, Derksen, Gasnier and
  Facciolo}]{ehret2023radar}
\bibinfo{author}{Ehret, T.}, \bibinfo{author}{Marí, R.},
  \bibinfo{author}{Derksen, D.}, \bibinfo{author}{Gasnier, N.},
  \bibinfo{author}{Facciolo, G.}, \bibinfo{year}{2023}.
\newblock \bibinfo{title}{Radar fields: An extension of radiance fields to
  sar}.
\bibitem[{Gableman and Kak(2024)}]{gableman2024incorporating}
\bibinfo{author}{Gableman, M.}, \bibinfo{author}{Kak, A.},
  \bibinfo{year}{2024}.
\newblock \bibinfo{title}{Incorporating season and solar specificity into
  renderings made by a nerf architecture using satellite images}.
\newblock \bibinfo{journal}{IEEE Transactions on Pattern Analysis and Machine
  Intelligence} .
\bibitem[{Gao et~al.(2003)Gao, Schaaf, Strahler, Jin and Li}]{gao2003detecting}
\bibinfo{author}{Gao, F.}, \bibinfo{author}{Schaaf, C.},
  \bibinfo{author}{Strahler, A.}, \bibinfo{author}{Jin, Y.},
  \bibinfo{author}{Li, X.}, \bibinfo{year}{2003}.
\newblock \bibinfo{title}{Detecting vegetation structure using a kernel-based
  brdf model}.
\newblock \bibinfo{journal}{Remote Sensing of Environment}
  \bibinfo{volume}{86}, \bibinfo{pages}{198--205}.
\bibitem[{Guo et~al.(2024)Guo, Wang, Gao, Xie and Song}]{guo2024depthguided}
\bibinfo{author}{Guo, S.}, \bibinfo{author}{Wang, Q.}, \bibinfo{author}{Gao,
  Y.}, \bibinfo{author}{Xie, R.}, \bibinfo{author}{Song, L.},
  \bibinfo{year}{2024}.
\newblock \bibinfo{title}{Depth-guided robust and fast point cloud fusion nerf
  for sparse input views}.
\bibitem[{Hapke(1981)}]{hapke1981bidirectional}
\bibinfo{author}{Hapke, B.}, \bibinfo{year}{1981}.
\newblock \bibinfo{title}{Bidirectional reflectance spectroscopy: 1. theory}.
\newblock \bibinfo{journal}{Journal of Geophysical Research: Solid Earth}
  \bibinfo{volume}{86}, \bibinfo{pages}{3039--3054}.
\bibitem[{Hirschm{\"u}ller(2008)}]{hirschmuller:08:sgm}
\bibinfo{author}{Hirschm{\"u}ller, H.}, \bibinfo{year}{2008}.
\newblock \bibinfo{title}{Stereo processing by semiglobal matching and mutual
  information}.
\newblock \bibinfo{journal}{Pattern Analysis and Machine Intelligence, IEEE
  Transactions on} \bibinfo{volume}{30}, \bibinfo{pages}{328--341}.
\bibitem[{Koukal et~al.(2014)Koukal, Atzberger and
  Schneider}]{koukal2014evaluation}
\bibinfo{author}{Koukal, T.}, \bibinfo{author}{Atzberger, C.},
  \bibinfo{author}{Schneider, W.}, \bibinfo{year}{2014}.
\newblock \bibinfo{title}{Evaluation of semi-empirical brdf models inverted
  against multi-angle data from a digital airborne frame camera for enhancing
  forest type classification}.
\newblock \bibinfo{journal}{Remote Sensing of Environment}
  \bibinfo{volume}{151}, \bibinfo{pages}{27--43}.
\bibitem[{Labarre et~al.(2019)Labarre, Jacquemoud, Ferrari, Delorme, Derrien,
  Grandin, Jalludin, Lemaitre, M{\'e}tois, Pierrot-Deseilligny
  et~al.}]{labarre2019retrieving}
\bibinfo{author}{Labarre, S.}, \bibinfo{author}{Jacquemoud, S.},
  \bibinfo{author}{Ferrari, C.}, \bibinfo{author}{Delorme, A.},
  \bibinfo{author}{Derrien, A.}, \bibinfo{author}{Grandin, R.},
  \bibinfo{author}{Jalludin, M.}, \bibinfo{author}{Lemaitre, F.},
  \bibinfo{author}{M{\'e}tois, M.}, \bibinfo{author}{Pierrot-Deseilligny, M.},
  et~al., \bibinfo{year}{2019}.
\newblock \bibinfo{title}{Retrieving soil surface roughness with the hapke
  photometric model: Confrontation with the ground truth}.
\newblock \bibinfo{journal}{Remote sensing of environment}
  \bibinfo{volume}{225}, \bibinfo{pages}{1--15}.
\bibitem[{Lattanzio et~al.(2007)Lattanzio, Govaerts and
  Pinty}]{lattanzio2007consistency}
\bibinfo{author}{Lattanzio, A.}, \bibinfo{author}{Govaerts, Y.},
  \bibinfo{author}{Pinty, B.}, \bibinfo{year}{2007}.
\newblock \bibinfo{title}{Consistency of surface anisotropy characterization
  with meteosat observations}.
\newblock \bibinfo{journal}{Advances in Space Research} \bibinfo{volume}{39},
  \bibinfo{pages}{131--135}.
\bibitem[{Li and Li(2022)}]{li2022neural}
\bibinfo{author}{Li, J.}, \bibinfo{author}{Li, H.}, \bibinfo{year}{2022}.
\newblock \bibinfo{title}{Neural reflectance for shape recovery with shadow
  handling}, in: \bibinfo{booktitle}{CVPR}, pp. \bibinfo{pages}{16221--16230}.
\bibitem[{Lucht et~al.(2000)Lucht, Schaaf and Strahler}]{lucht2000algorithm}
\bibinfo{author}{Lucht, W.}, \bibinfo{author}{Schaaf, C.B.},
  \bibinfo{author}{Strahler, A.H.}, \bibinfo{year}{2000}.
\newblock \bibinfo{title}{An algorithm for the retrieval of albedo from space
  using semiempirical brdf models}.
\newblock \bibinfo{journal}{IEEE Transactions on Geoscience and Remote sensing}
  \bibinfo{volume}{38}, \bibinfo{pages}{977--998}.
\bibitem[{Lv and Sun(2016)}]{lv2016multi}
\bibinfo{author}{Lv, Y.}, \bibinfo{author}{Sun, Z.}, \bibinfo{year}{2016}.
\newblock \bibinfo{title}{Multi-angular spectral reflectance to characterize
  the particle size of surfaces of desert and cultivated soil}.
\newblock \bibinfo{journal}{European Journal of Soil Science}
  \bibinfo{volume}{67}, \bibinfo{pages}{253--265}.
\bibitem[{Mai et~al.(2023)Mai, Verbin, Kuester and
  Fridovich-Keil}]{mai2023neural}
\bibinfo{author}{Mai, A.}, \bibinfo{author}{Verbin, D.},
  \bibinfo{author}{Kuester, F.}, \bibinfo{author}{Fridovich-Keil, S.},
  \bibinfo{year}{2023}.
\newblock \bibinfo{title}{Neural microfacet fields for inverse rendering}.
\bibitem[{Mar{\'\i} et~al.(2022)Mar{\'\i}, Facciolo and Ehret}]{mari2022sat}
\bibinfo{author}{Mar{\'\i}, R.}, \bibinfo{author}{Facciolo, G.},
  \bibinfo{author}{Ehret, T.}, \bibinfo{year}{2022}.
\newblock \bibinfo{title}{{Sat-NeRF: Learning multi-view satellite
  photogrammetry with transient objects and shadow modeling using RPC
  cameras}}, in: \bibinfo{booktitle}{CVPRW}, pp. \bibinfo{pages}{1311--1321}.
\bibitem[{Mar{\'\i} et~al.(2023)Mar{\'\i}, Facciolo and Ehret}]{Mari_2023_CVPR}
\bibinfo{author}{Mar{\'\i}, R.}, \bibinfo{author}{Facciolo, G.},
  \bibinfo{author}{Ehret, T.}, \bibinfo{year}{2023}.
\newblock \bibinfo{title}{Multi-date earth observation nerf: The detail is in
  the shadows}, in: \bibinfo{booktitle}{CVPRW}, pp.
  \bibinfo{pages}{2034--2044}.
\bibitem[{Martonchik et~al.(1998a)Martonchik, Diner, Kahn, Ackerman,
  Verstraete, Pinty and Gordon}]{martonchik1998techniques}
\bibinfo{author}{Martonchik, J.V.}, \bibinfo{author}{Diner, D.J.},
  \bibinfo{author}{Kahn, R.A.}, \bibinfo{author}{Ackerman, T.P.},
  \bibinfo{author}{Verstraete, M.M.}, \bibinfo{author}{Pinty, B.},
  \bibinfo{author}{Gordon, H.R.}, \bibinfo{year}{1998}a.
\newblock \bibinfo{title}{Techniques for the retrieval of aerosol properties
  over land and ocean using multiangle imaging}.
\newblock \bibinfo{journal}{IEEE Transactions on Geoscience and Remote Sensing}
  \bibinfo{volume}{36}, \bibinfo{pages}{1212--1227}.
\bibitem[{Martonchik et~al.(1998b)Martonchik, Diner, Pinty, Verstraete, Myneni,
  Knyazikhin and Gordon}]{martonchik1998determination}
\bibinfo{author}{Martonchik, J.V.}, \bibinfo{author}{Diner, D.J.},
  \bibinfo{author}{Pinty, B.}, \bibinfo{author}{Verstraete, M.M.},
  \bibinfo{author}{Myneni, R.B.}, \bibinfo{author}{Knyazikhin, Y.},
  \bibinfo{author}{Gordon, H.R.}, \bibinfo{year}{1998}b.
\newblock \bibinfo{title}{Determination of land and ocean reflective,
  radiative, and biophysical properties using multiangle imaging}.
\newblock \bibinfo{journal}{IEEE Transactions on Geoscience and Remote Sensing}
  \bibinfo{volume}{36}, \bibinfo{pages}{1266--1281}.
\bibitem[{Mildenhall et~al.(2020)Mildenhall, Srinivasan, Tancik, Barron,
  Ramamoorthi and Ng}]{Mildenhall20eccv_nerf}
\bibinfo{author}{Mildenhall, B.}, \bibinfo{author}{Srinivasan, P.P.},
  \bibinfo{author}{Tancik, M.}, \bibinfo{author}{Barron, J.T.},
  \bibinfo{author}{Ramamoorthi, R.}, \bibinfo{author}{Ng, R.},
  \bibinfo{year}{2020}.
\newblock \bibinfo{title}{{NeRF}: Representing scenes as neural radiance fields
  for view synthesis}, in: \bibinfo{booktitle}{ECCV}, pp.
  \bibinfo{pages}{99--106}.
\bibitem[{Minnaert(1941)}]{minnaert1941reciprocity}
\bibinfo{author}{Minnaert, M.}, \bibinfo{year}{1941}.
\newblock \bibinfo{title}{The reciprocity principle in lunar photometry}.
\newblock \bibinfo{journal}{Astrophysical Journal, vol. 93, p. 403-410 (1941).}
  \bibinfo{volume}{93}, \bibinfo{pages}{403--410}.
\bibitem[{Pierrot-Deseilligny and Paparoditis(2006)}]{mpd:06:sgm}
\bibinfo{author}{Pierrot-Deseilligny, M.}, \bibinfo{author}{Paparoditis, N.},
  \bibinfo{year}{2006}.
\newblock \bibinfo{title}{A multiresolution and optimization-based image
  matching approach: An application to surface reconstruction from spot5-hrs
  stereo imagery.}, in: \bibinfo{booktitle}{ISPRS Workshop On Topographic
  Mapping From Space}.
\bibitem[{Pinty and Verstraete(1991)}]{pinty1991extracting}
\bibinfo{author}{Pinty, B.}, \bibinfo{author}{Verstraete, M.M.},
  \bibinfo{year}{1991}.
\newblock \bibinfo{title}{Extracting information on surface properties from
  bidirectional reflectance measurements}.
\newblock \bibinfo{journal}{Journal of Geophysical Research: Atmospheres}
  \bibinfo{volume}{96}, \bibinfo{pages}{2865--2874}.
\bibitem[{Privette and Roy(2002)}]{privette2002first}
\bibinfo{author}{Privette, J.}, \bibinfo{author}{Roy, D.},
  \bibinfo{year}{2002}.
\newblock \bibinfo{title}{First operational brdf, albedo and nadir reflectance
  products from modis}.
\newblock \bibinfo{journal}{Remote Sensing of Environment} ,
  \bibinfo{pages}{83--135}.
\bibitem[{Qu and Deng(2023)}]{qu2023sat}
\bibinfo{author}{Qu, Y.}, \bibinfo{author}{Deng, F.}, \bibinfo{year}{2023}.
\newblock \bibinfo{title}{Sat-mesh: Learning neural implicit surfaces for
  multi-view satellite reconstruction}.
\newblock \bibinfo{journal}{Remote Sensing} \bibinfo{volume}{15},
  \bibinfo{pages}{4297}.
\bibitem[{Rahman et~al.(1993)Rahman, Pinty and Verstraete}]{rahman1993coupled}
\bibinfo{author}{Rahman, H.}, \bibinfo{author}{Pinty, B.},
  \bibinfo{author}{Verstraete, M.M.}, \bibinfo{year}{1993}.
\newblock \bibinfo{title}{Coupled surface-atmosphere reflectance (csar) model:
  2. semiempirical surface model usable with noaa advanced very high resolution
  radiometer data}.
\newblock \bibinfo{journal}{Journal of Geophysical Research: Atmospheres}
  \bibinfo{volume}{98}, \bibinfo{pages}{20791--20801}.
\bibitem[{Roessle et~al.(2022)Roessle, Barron, Mildenhall, Srinivasan and
  Nie{\ss}ner}]{roessle2022dense}
\bibinfo{author}{Roessle, B.}, \bibinfo{author}{Barron, J.T.},
  \bibinfo{author}{Mildenhall, B.}, \bibinfo{author}{Srinivasan, P.P.},
  \bibinfo{author}{Nie{\ss}ner, M.}, \bibinfo{year}{2022}.
\newblock \bibinfo{title}{Dense depth priors for neural radiance fields from
  sparse input views}, in: \bibinfo{booktitle}{CVPR}, pp.
  \bibinfo{pages}{12892--12901}.
\bibitem[{Rosu et~al.(2015)Rosu, Pierrot-Deseilligny, Delorme, Binet and
  Klinger}]{rosu2015measurement}
\bibinfo{author}{Rosu, A.}, \bibinfo{author}{Pierrot-Deseilligny, M.},
  \bibinfo{author}{Delorme, A.}, \bibinfo{author}{Binet, R.},
  \bibinfo{author}{Klinger, Y.}, \bibinfo{year}{2015}.
\newblock \bibinfo{title}{Measurement of ground displacement from optical
  satellite image correlation using the free open-source software micmac}.
\newblock \bibinfo{journal}{ISPRS Journal of Photogrammetry and Remote Sensing}
  \bibinfo{volume}{100}, \bibinfo{pages}{48--59}.
\bibitem[{Roujean et~al.(1992)Roujean, Leroy and
  Deschamps}]{roujean1992bidirectional}
\bibinfo{author}{Roujean, J.L.}, \bibinfo{author}{Leroy, M.},
  \bibinfo{author}{Deschamps, P.Y.}, \bibinfo{year}{1992}.
\newblock \bibinfo{title}{A bidirectional reflectance model of the earth's
  surface for the correction of remote sensing data}.
\newblock \bibinfo{journal}{Journal of Geophysical Research: Atmospheres}
  \bibinfo{volume}{97}, \bibinfo{pages}{20455--20468}.
\bibitem[{Sandmeier and Strahler(2000)}]{sandmeier2000brdf}
\bibinfo{author}{Sandmeier, S.}, \bibinfo{author}{Strahler, A.},
  \bibinfo{year}{2000}.
\newblock \bibinfo{title}{Brdf laboratory measurements}.
\newblock \bibinfo{journal}{Remote Sensing Reviews} \bibinfo{volume}{18},
  \bibinfo{pages}{481--502}.
\bibitem[{Shibayama and Wiegand(1985)}]{shibayama1985view}
\bibinfo{author}{Shibayama, M.}, \bibinfo{author}{Wiegand, C.},
  \bibinfo{year}{1985}.
\newblock \bibinfo{title}{View azimuth and zenith, and solar angle effects on
  wheat canopy reflectance}.
\newblock \bibinfo{journal}{Remote Sensing of Environment}
  \bibinfo{volume}{18}, \bibinfo{pages}{91--103}.
\bibitem[{Somraj and Soundararajan(2023)}]{Somraj_2023}
\bibinfo{author}{Somraj, N.}, \bibinfo{author}{Soundararajan, R.},
  \bibinfo{year}{2023}.
\newblock \bibinfo{title}{Vip-nerf: Visibility prior for sparse input neural
  radiance fields}, in: \bibinfo{booktitle}{ACM SIGGRAPH 2023 Conference
  Proceedings}, pp. \bibinfo{pages}{1--11}.
\bibitem[{national~d'études spatiales(2002)}]{OTB2002}
\bibinfo{author}{national~d'études spatiales, C.}, \bibinfo{year}{2002}.
\newblock \bibinfo{title}{Orfeo toolbox}.
\newblock \bibinfo{note}{\url{https://www.orfeo-toolbox.org/}}.
\bibitem[{Srinivasan et~al.(2021)Srinivasan, Deng, Zhang, Tancik, Mildenhall
  and Barron}]{srinivasan2020nerv}
\bibinfo{author}{Srinivasan, P.P.}, \bibinfo{author}{Deng, B.},
  \bibinfo{author}{Zhang, X.}, \bibinfo{author}{Tancik, M.},
  \bibinfo{author}{Mildenhall, B.}, \bibinfo{author}{Barron, J.T.},
  \bibinfo{year}{2021}.
\newblock \bibinfo{title}{Nerv: Neural reflectance and visibility fields for
  relighting and view synthesis}, in: \bibinfo{booktitle}{CVPR}, pp.
  \bibinfo{pages}{7495--7504}.
\bibitem[{Verbin et~al.(2022)Verbin, Hedman, Mildenhall, Zickler, Barron and
  Srinivasan}]{verbin2022refnerf}
\bibinfo{author}{Verbin, D.}, \bibinfo{author}{Hedman, P.},
  \bibinfo{author}{Mildenhall, B.}, \bibinfo{author}{Zickler, T.},
  \bibinfo{author}{Barron, J.T.}, \bibinfo{author}{Srinivasan, P.P.},
  \bibinfo{year}{2022}.
\newblock \bibinfo{title}{{Ref-NeRF}: Structured view-dependent appearance for
  neural radiance fields}.
\newblock \bibinfo{journal}{CVPR} .
\bibitem[{Vermote et~al.(1997)Vermote, Tanr{\'e}, Deuze, Herman and
  Morcette}]{vermote1997second}
\bibinfo{author}{Vermote, E.F.}, \bibinfo{author}{Tanr{\'e}, D.},
  \bibinfo{author}{Deuze, J.L.}, \bibinfo{author}{Herman, M.},
  \bibinfo{author}{Morcette, J.J.}, \bibinfo{year}{1997}.
\newblock \bibinfo{title}{Second simulation of the satellite signal in the
  solar spectrum, 6s: An overview}.
\newblock \bibinfo{journal}{IEEE transactions on geoscience and remote sensing}
  \bibinfo{volume}{35}, \bibinfo{pages}{675--686}.
\bibitem[{Walter et~al.(2007)Walter, Marschner, Li and
  Torrance}]{walter2007microfacet}
\bibinfo{author}{Walter, B.}, \bibinfo{author}{Marschner, S.R.},
  \bibinfo{author}{Li, H.}, \bibinfo{author}{Torrance, K.E.},
  \bibinfo{year}{2007}.
\newblock \bibinfo{title}{Microfacet models for refraction through rough
  surfaces}, in: \bibinfo{booktitle}{Proceedings of the 18th Eurographics
  conference on Rendering Techniques}, pp. \bibinfo{pages}{195--206}.
\bibitem[{Walthall et~al.(1985)Walthall, Norman, Welles, Campbell and
  Blad}]{walthall1985simple}
\bibinfo{author}{Walthall, C.}, \bibinfo{author}{Norman, J.},
  \bibinfo{author}{Welles, J.}, \bibinfo{author}{Campbell, G.},
  \bibinfo{author}{Blad, B.}, \bibinfo{year}{1985}.
\newblock \bibinfo{title}{Simple equation to approximate the bidirectional
  reflectance from vegetative canopies and bare soil surfaces}.
\newblock \bibinfo{journal}{Applied Optics} \bibinfo{volume}{24},
  \bibinfo{pages}{383--387}.
\bibitem[{Wan et~al.(2024)Wan, Guan, Zhao, Wen and She}]{wan2024constraining}
\bibinfo{author}{Wan, Q.}, \bibinfo{author}{Guan, Y.}, \bibinfo{author}{Zhao,
  Q.}, \bibinfo{author}{Wen, X.}, \bibinfo{author}{She, J.},
  \bibinfo{year}{2024}.
\newblock \bibinfo{title}{Constraining the geometry of nerfs for accurate dsm
  generation from multi-view satellite images}.
\newblock \bibinfo{journal}{ISPRS International Journal of Geo-Information}
  \bibinfo{volume}{13}, \bibinfo{pages}{243}.
\bibitem[{Wang et~al.(2023)Wang, Chen, Loy and Liu}]{wang2022sparsenerf}
\bibinfo{author}{Wang, G.}, \bibinfo{author}{Chen, Z.}, \bibinfo{author}{Loy,
  C.C.}, \bibinfo{author}{Liu, Z.}, \bibinfo{year}{2023}.
\newblock \bibinfo{title}{Sparsenerf: Distilling depth ranking for few-shot
  novel view synthesis}, in: \bibinfo{booktitle}{ICCV}.
\bibitem[{Wang et~al.(2004)Wang, Bovik, Sheikh and Simoncelli}]{wang2004image}
\bibinfo{author}{Wang, Z.}, \bibinfo{author}{Bovik, A.C.},
  \bibinfo{author}{Sheikh, H.R.}, \bibinfo{author}{Simoncelli, E.P.},
  \bibinfo{year}{2004}.
\newblock \bibinfo{title}{Image quality assessment: from error visibility to
  structural similarity}.
\newblock \bibinfo{journal}{IEEE transactions on image processing}
  \bibinfo{volume}{13}, \bibinfo{pages}{600--612}.
\bibitem[{Wanner et~al.(1995)Wanner, Li and Strahler}]{wanner1995derivation}
\bibinfo{author}{Wanner, W.}, \bibinfo{author}{Li, X.},
  \bibinfo{author}{Strahler, A.}, \bibinfo{year}{1995}.
\newblock \bibinfo{title}{On the derivation of kernels for kernel-driven models
  of bidirectional reflectance}.
\newblock \bibinfo{journal}{Journal of Geophysical Research: Atmospheres}
  \bibinfo{volume}{100}, \bibinfo{pages}{21077--21089}.
\bibitem[{Wei et~al.(2021)Wei, Liu, Rao, Zhao, Lu and Zhou}]{wei2021nerfingmvs}
\bibinfo{author}{Wei, Y.}, \bibinfo{author}{Liu, S.}, \bibinfo{author}{Rao,
  Y.}, \bibinfo{author}{Zhao, W.}, \bibinfo{author}{Lu, J.},
  \bibinfo{author}{Zhou, J.}, \bibinfo{year}{2021}.
\newblock \bibinfo{title}{Nerfingmvs: Guided optimization of neural radiance
  fields for indoor multi-view stereo}.
\bibitem[{Widlowski et~al.(2001)Widlowski, Pinty, Gobron, Verstraete and
  Davis}]{widlowski2001characterization}
\bibinfo{author}{Widlowski, J.L.}, \bibinfo{author}{Pinty, B.},
  \bibinfo{author}{Gobron, N.}, \bibinfo{author}{Verstraete, M.M.},
  \bibinfo{author}{Davis, A.B.}, \bibinfo{year}{2001}.
\newblock \bibinfo{title}{Characterization of surface heterogeneity detected at
  the misr/terra subpixel scale}.
\newblock \bibinfo{journal}{Geophysical Research Letters} \bibinfo{volume}{28},
  \bibinfo{pages}{4639--4642}.
\bibitem[{Widlowski et~al.(2004)Widlowski, Pinty, Gobron, Verstraete, Diner and
  Davis}]{widlowski2004canopy}
\bibinfo{author}{Widlowski, J.L.}, \bibinfo{author}{Pinty, B.},
  \bibinfo{author}{Gobron, N.}, \bibinfo{author}{Verstraete, M.M.},
  \bibinfo{author}{Diner, D.}, \bibinfo{author}{Davis, A.},
  \bibinfo{year}{2004}.
\newblock \bibinfo{title}{Canopy structure parameters derived from
  multi-angular remote sensing data for terrestrial carbon studies}.
\newblock \bibinfo{journal}{Climatic Change} \bibinfo{volume}{67},
  \bibinfo{pages}{403--415}.
\bibitem[{Wu et~al.(2024)Wu, Vallet, Pierrot-Deseilligny and
  Rupnik}]{wu2024evaluation}
\bibinfo{author}{Wu, T.}, \bibinfo{author}{Vallet, B.},
  \bibinfo{author}{Pierrot-Deseilligny, M.}, \bibinfo{author}{Rupnik, E.},
  \bibinfo{year}{2024}.
\newblock \bibinfo{title}{An evaluation of deep learning based stereo dense
  matching dataset shift from aerial images and a large scale stereo dataset}.
\newblock \bibinfo{journal}{International Journal of Applied Earth Observation
  and Geoinformation} \bibinfo{volume}{128}, \bibinfo{pages}{103715}.
\bibitem[{Xie et~al.(2023)Xie, Zhang, Jeon and Yang}]{xie2023remote}
\bibinfo{author}{Xie, S.}, \bibinfo{author}{Zhang, L.}, \bibinfo{author}{Jeon,
  G.}, \bibinfo{author}{Yang, X.}, \bibinfo{year}{2023}.
\newblock \bibinfo{title}{Remote sensing neural radiance fields for multi-view
  satellite photogrammetry}.
\newblock \bibinfo{journal}{Remote Sensing} \bibinfo{volume}{15},
  \bibinfo{pages}{3808}.
\bibitem[{Xu et~al.(2022)Xu, Jiang, Wang, Fan, Shi and Wang}]{xu2022sinnerf}
\bibinfo{author}{Xu, D.}, \bibinfo{author}{Jiang, Y.}, \bibinfo{author}{Wang,
  P.}, \bibinfo{author}{Fan, Z.}, \bibinfo{author}{Shi, H.},
  \bibinfo{author}{Wang, Z.}, \bibinfo{year}{2022}.
\newblock \bibinfo{title}{Sinnerf: Training neural radiance fields on complex
  scenes from a single image}, in: \bibinfo{booktitle}{ECCV},
  \bibinfo{organization}{Springer}. pp. \bibinfo{pages}{736--753}.
\bibitem[{Yang et~al.(2022a)Yang, Chen, Chen, Chen and Wong}]{yang2022psnerf}
\bibinfo{author}{Yang, W.}, \bibinfo{author}{Chen, G.}, \bibinfo{author}{Chen,
  C.}, \bibinfo{author}{Chen, Z.}, \bibinfo{author}{Wong, K.Y.K.},
  \bibinfo{year}{2022}a.
\newblock \bibinfo{title}{Ps-nerf: Neural inverse rendering for multi-view
  photometric stereo}, in: \bibinfo{booktitle}{ECCV}.
\bibitem[{Yang et~al.(2022b)Yang, Chen, Chen, Chen and Wong}]{yang2022s}
\bibinfo{author}{Yang, W.}, \bibinfo{author}{Chen, G.}, \bibinfo{author}{Chen,
  C.}, \bibinfo{author}{Chen, Z.}, \bibinfo{author}{Wong, K.Y.K.},
  \bibinfo{year}{2022}b.
\newblock \bibinfo{title}{S $3$-nerf: Neural reflectance field from shading and
  shadow under a single viewpoint}.
\newblock \bibinfo{journal}{Advances in Neural Information Processing Systems}
  \bibinfo{volume}{35}, \bibinfo{pages}{1568--1582}.
\bibitem[{Yu et~al.(2021)Yu, Ye, Tancik and Kanazawa}]{yu2021pixelnerf}
\bibinfo{author}{Yu, A.}, \bibinfo{author}{Ye, V.}, \bibinfo{author}{Tancik,
  M.}, \bibinfo{author}{Kanazawa, A.}, \bibinfo{year}{2021}.
\newblock \bibinfo{title}{pixelnerf: Neural radiance fields from one or few
  images}, in: \bibinfo{booktitle}{CVPR}, pp. \bibinfo{pages}{4578--4587}.
\bibitem[{Zhang and Rupnik(2023)}]{zhang2023spsnerf1}
\bibinfo{author}{Zhang, L.}, \bibinfo{author}{Rupnik, E.},
  \bibinfo{year}{2023}.
\newblock \bibinfo{title}{{SparseSat-NeRF: Dense Depth Supervised Neural
  Radiance Fields for Sparse Satellite Images}}.
\newblock \bibinfo{journal}{ISPRS Annals} .
\bibitem[{Zhang et~al.(2024)Zhang, Zhou, Li and Wei}]{zhang2024satensorf}
\bibinfo{author}{Zhang, T.}, \bibinfo{author}{Zhou, Y.}, \bibinfo{author}{Li,
  Y.}, \bibinfo{author}{Wei, X.}, \bibinfo{year}{2024}.
\newblock \bibinfo{title}{Satensorf: Fast satellite tensorial radiance field
  for multi-date satellite imagery of large size}.
\newblock \bibinfo{journal}{IEEE Transactions on Geoscience and Remote Sensing}
  .
\bibitem[{Zhang et~al.(2021)Zhang, Srinivasan, Deng, Debevec, Freeman and
  Barron}]{Zhang_2021}
\bibinfo{author}{Zhang, X.}, \bibinfo{author}{Srinivasan, P.P.},
  \bibinfo{author}{Deng, B.}, \bibinfo{author}{Debevec, P.},
  \bibinfo{author}{Freeman, W.T.}, \bibinfo{author}{Barron, J.T.},
  \bibinfo{year}{2021}.
\newblock \bibinfo{title}{Nerfactor: Neural factorization of shape and
  reflectance under an unknown illumination}.
\newblock \bibinfo{journal}{ACM Transactions on Graphics (ToG)}
  \bibinfo{volume}{40}, \bibinfo{pages}{1--18}.

\end{thebibliography}




\end{document}